\documentclass{article} 
\usepackage{iclr2025_conference,times}

\usepackage{amsmath,amsfonts,bm}

\def\eqref#1{equation~\ref{#1}}

\def\1{\bm{1}}

\DeclareMathAlphabet{\mathsfit}{\encodingdefault}{\sfdefault}{m}{sl}
\SetMathAlphabet{\mathsfit}{bold}{\encodingdefault}{\sfdefault}{bx}{n}

\usepackage{graphicx}
\usepackage{hyperref}
\usepackage{url}
\usepackage[ruled]{algorithm2e}
\newcommand{\RN}[1]{%
    \textup{\lowercase\expandafter{\it \romannumeral#1}}%
}

\title{TLCM: Training-Efficient Latent Consistency Model for Image Generation with 2-8 Steps}

\footnotetext{$\dagger$ Corresponding authors.}
\author{Qingsong Xie\textsuperscript{1}$\dagger$, Zhenyi Liao\textsuperscript{2}, {Zhijie Deng}\textsuperscript{2}$\dagger$, {Chen Chen}\textsuperscript{1} \& {Haonan Lu}\textsuperscript{1} \\
\textsuperscript{1}AI Center, Guangdong OPPO Mobile Telecommunications Corp., Ltd \\
\textsuperscript{2}Qing Yuan Research Institute, SEIEE, Shanghai Jiao Tong University\\
\textcolor{blue}{\textbf{Project: \url{https://github.com/OPPO-Mente-Lab/TLCM}}}
}

\iclrfinalcopy 
\begin{document}

\maketitle

\begin{figure}[ht]
\vspace{-3ex}
  \centering
  \includegraphics[width=0.24\linewidth]{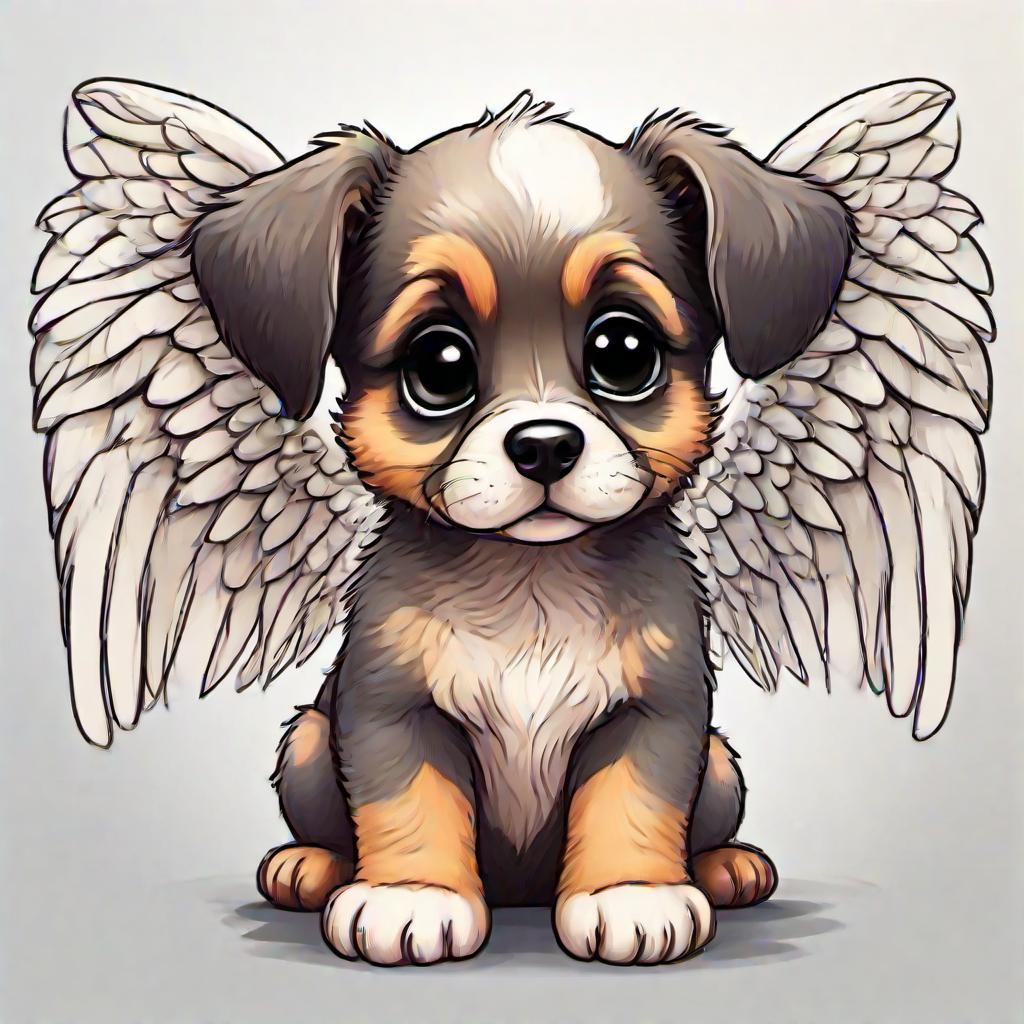}
  \hspace{-5pt}
  \includegraphics[width=0.24\linewidth]{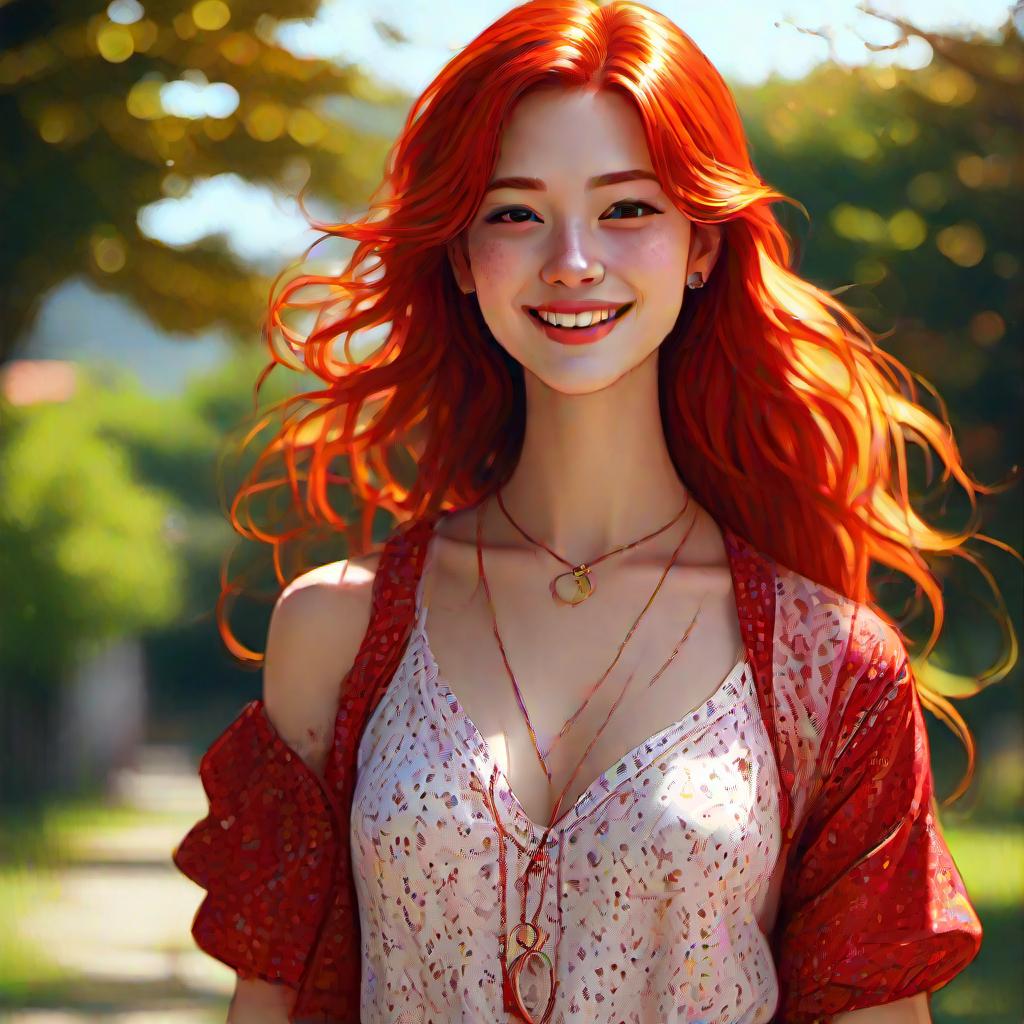}
   \hspace{-5pt}
  \includegraphics[width=0.24\linewidth]{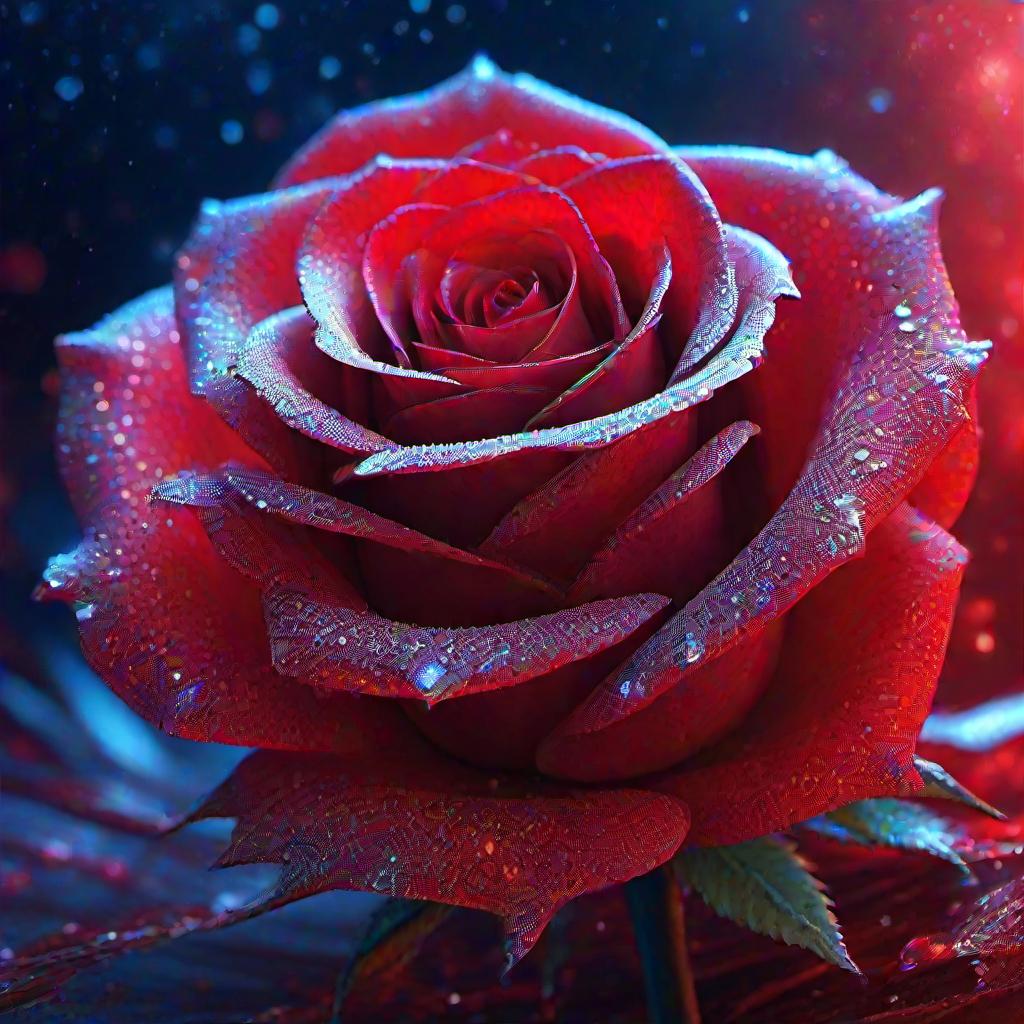}
   \hspace{-5pt}
  \includegraphics[width=0.24\linewidth]{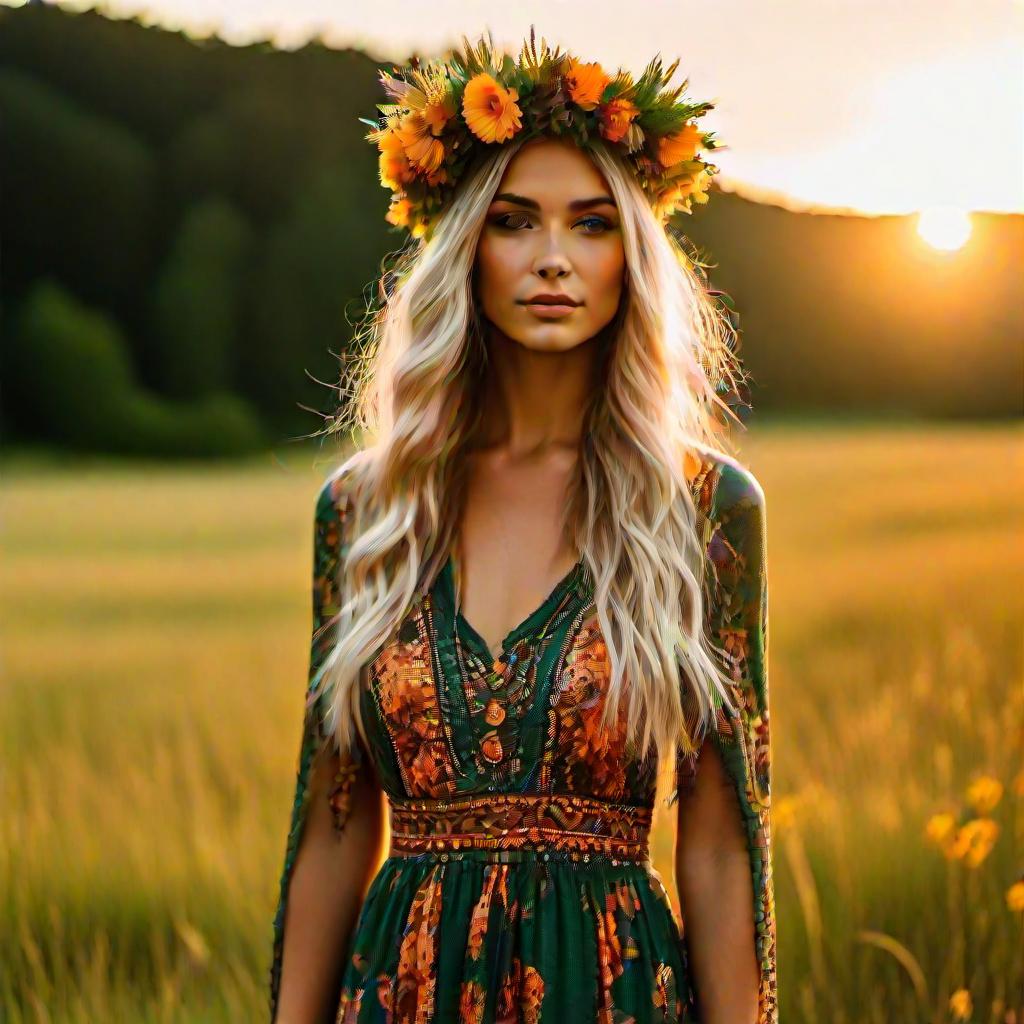}
    \vspace{-1pt}
    \includegraphics[width=0.24\linewidth]{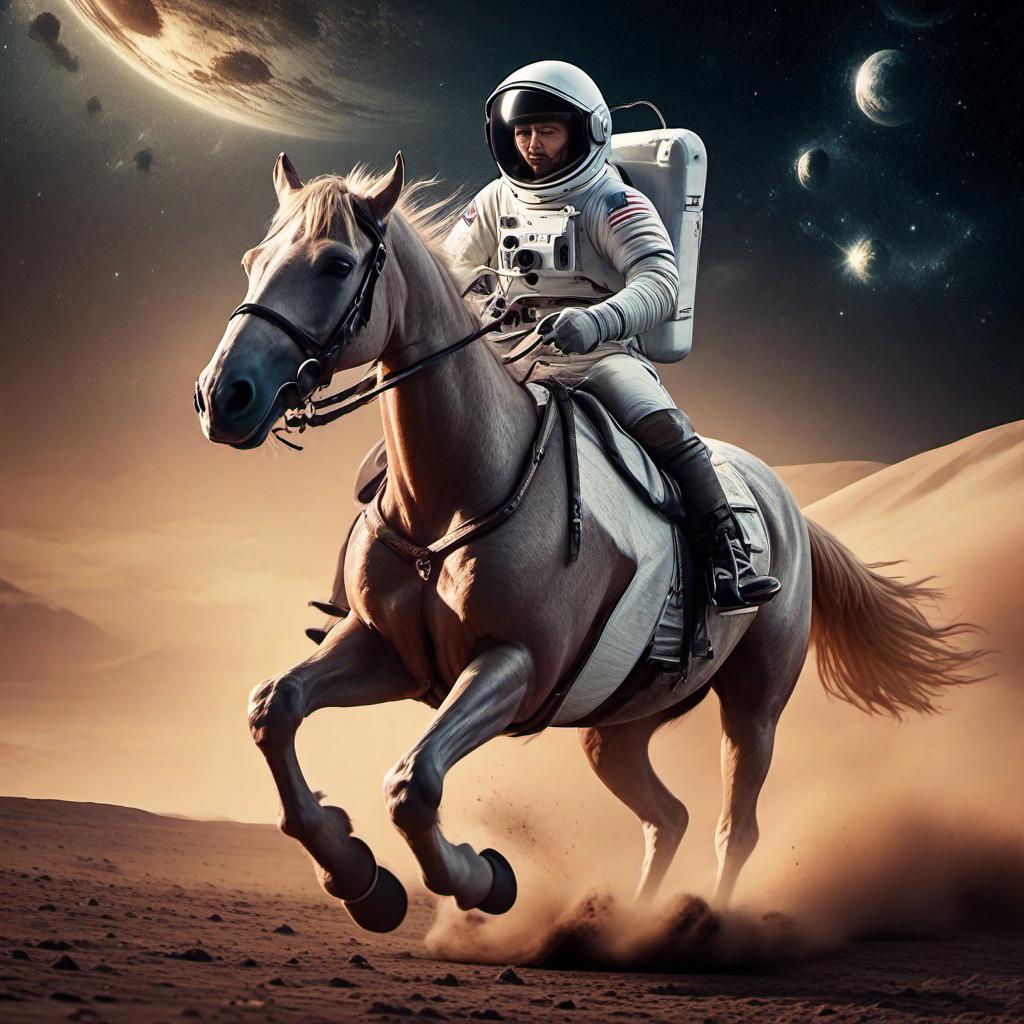}
    \hspace{-5pt}
    \includegraphics[width=0.24\linewidth]{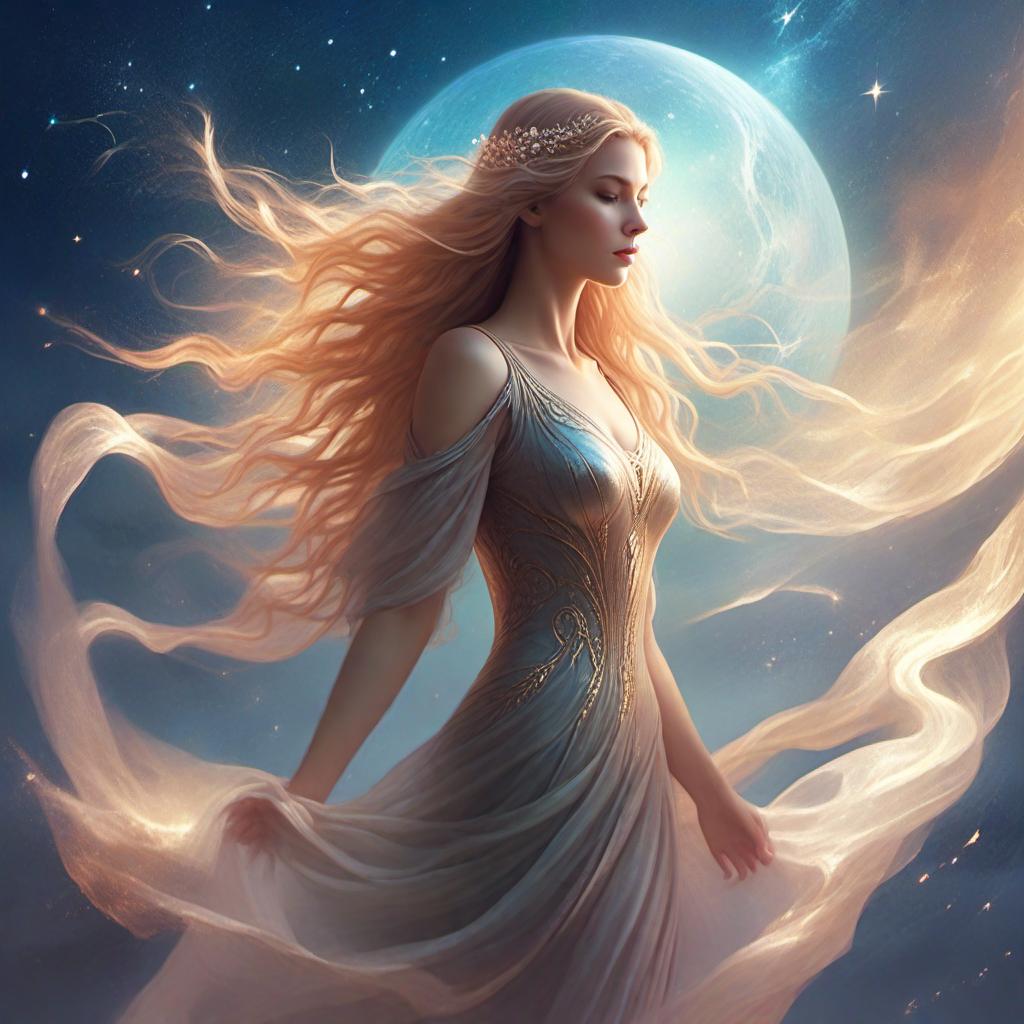}
    \hspace{-5pt}
  \includegraphics[width=0.24\linewidth]{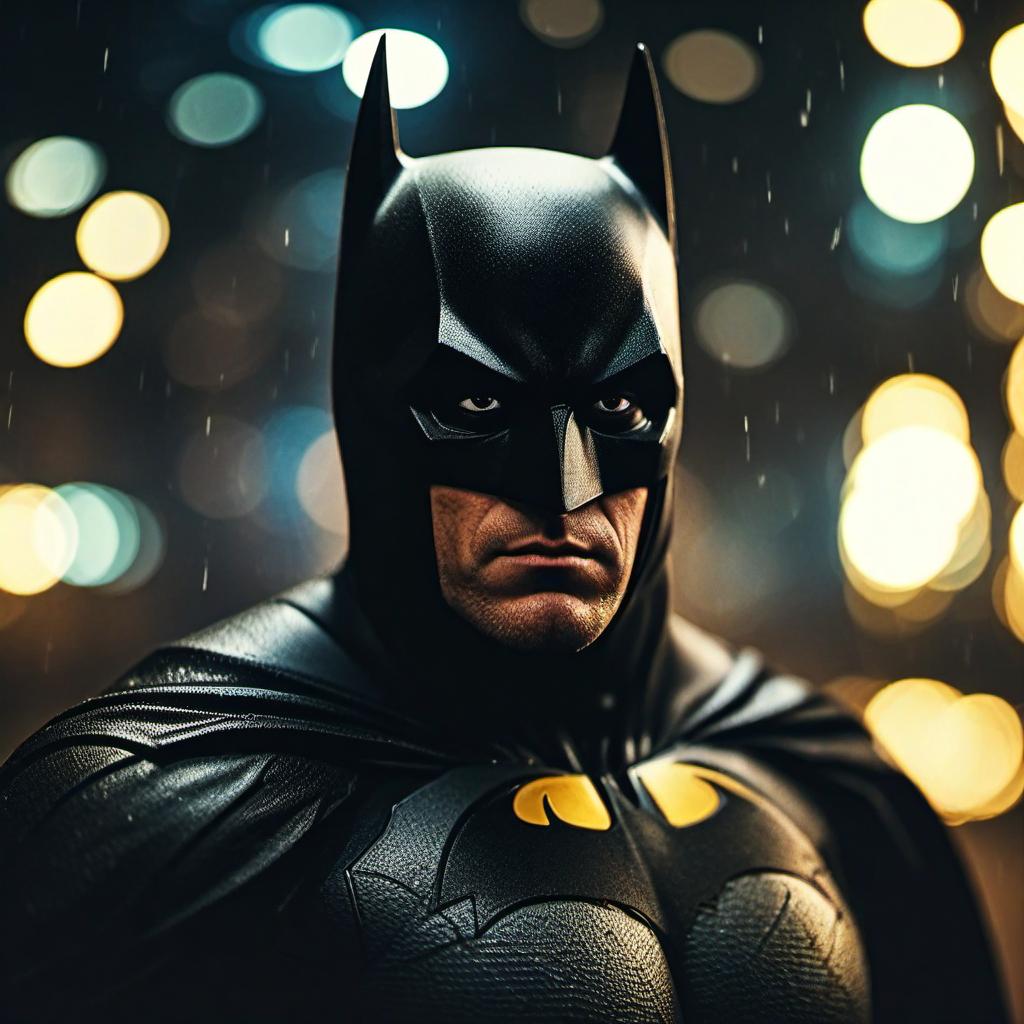}
    \hspace{-5pt}
     \includegraphics[width=0.24\linewidth]{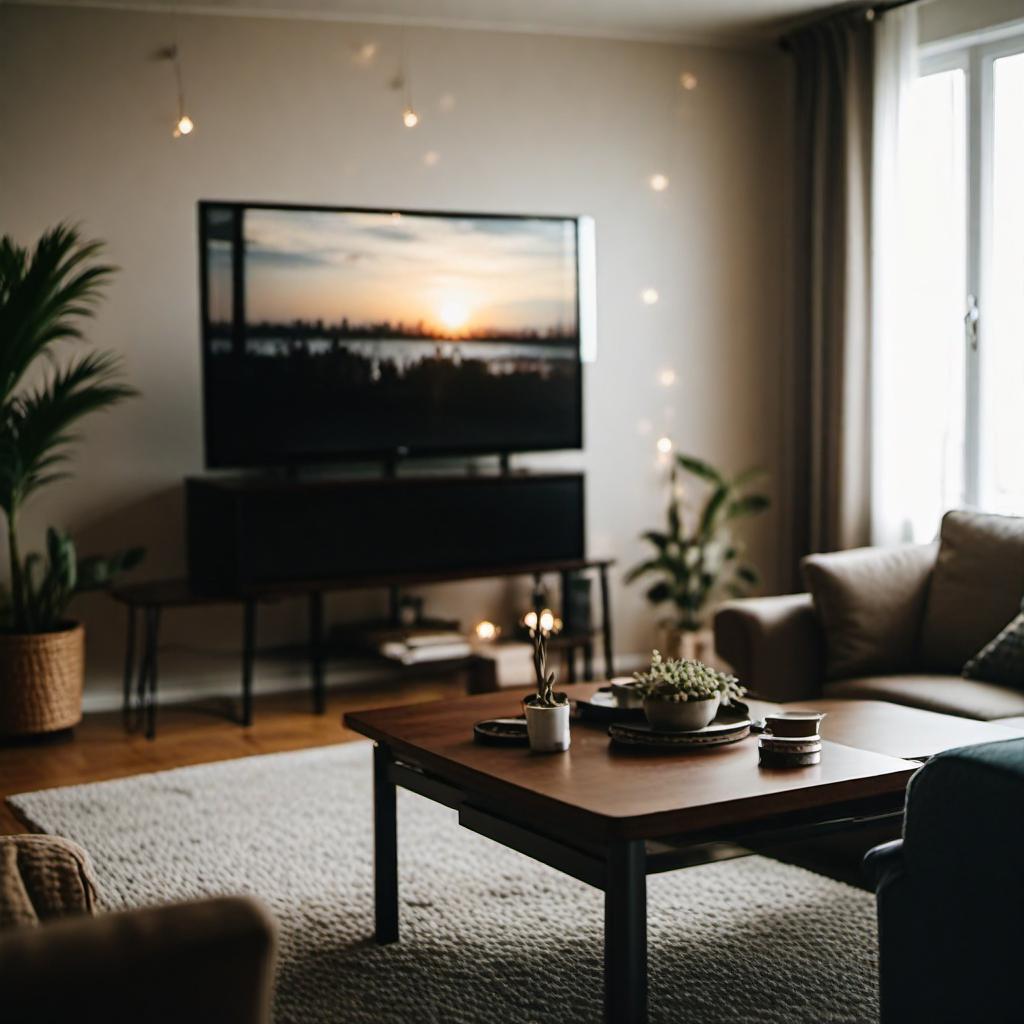}
     \vspace{-1pt}
  \includegraphics[width=0.24\linewidth]{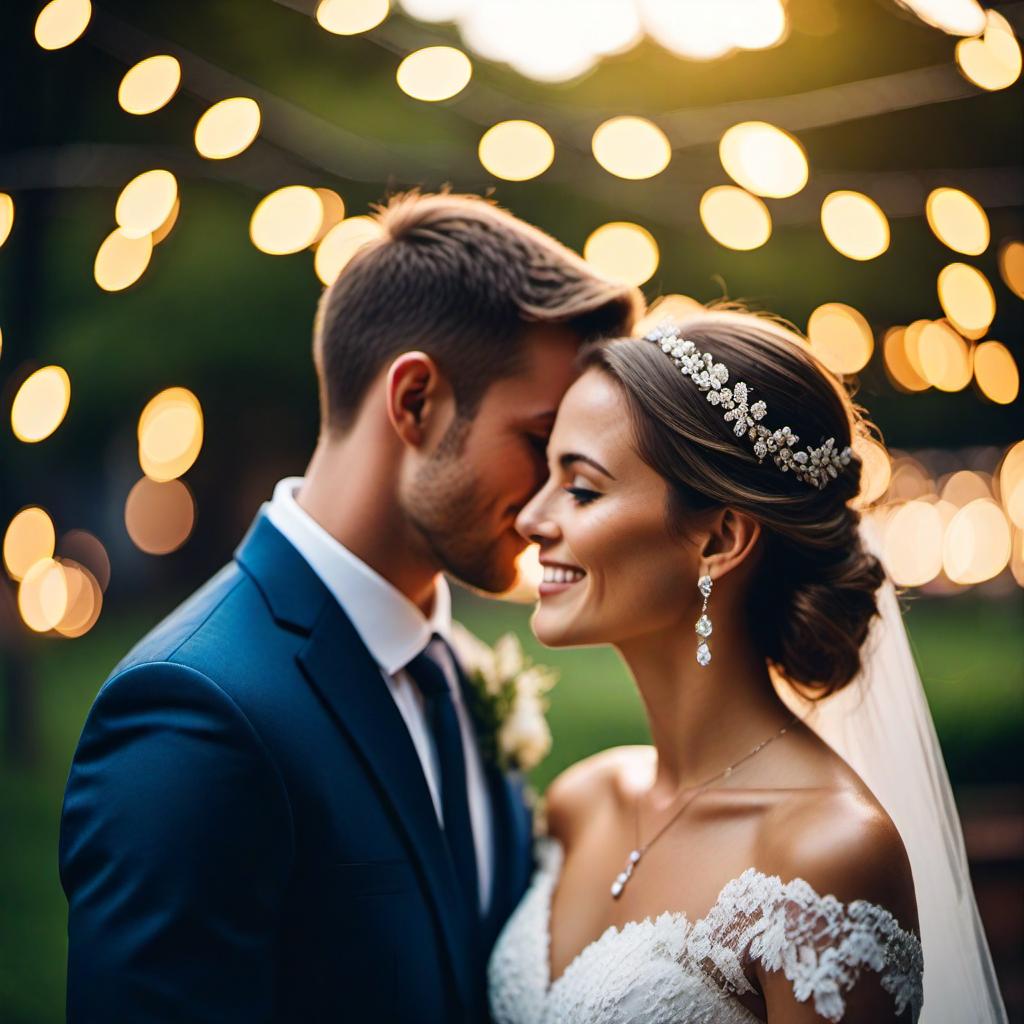}
    \hspace{-5pt}
  \includegraphics[width=0.24\linewidth]{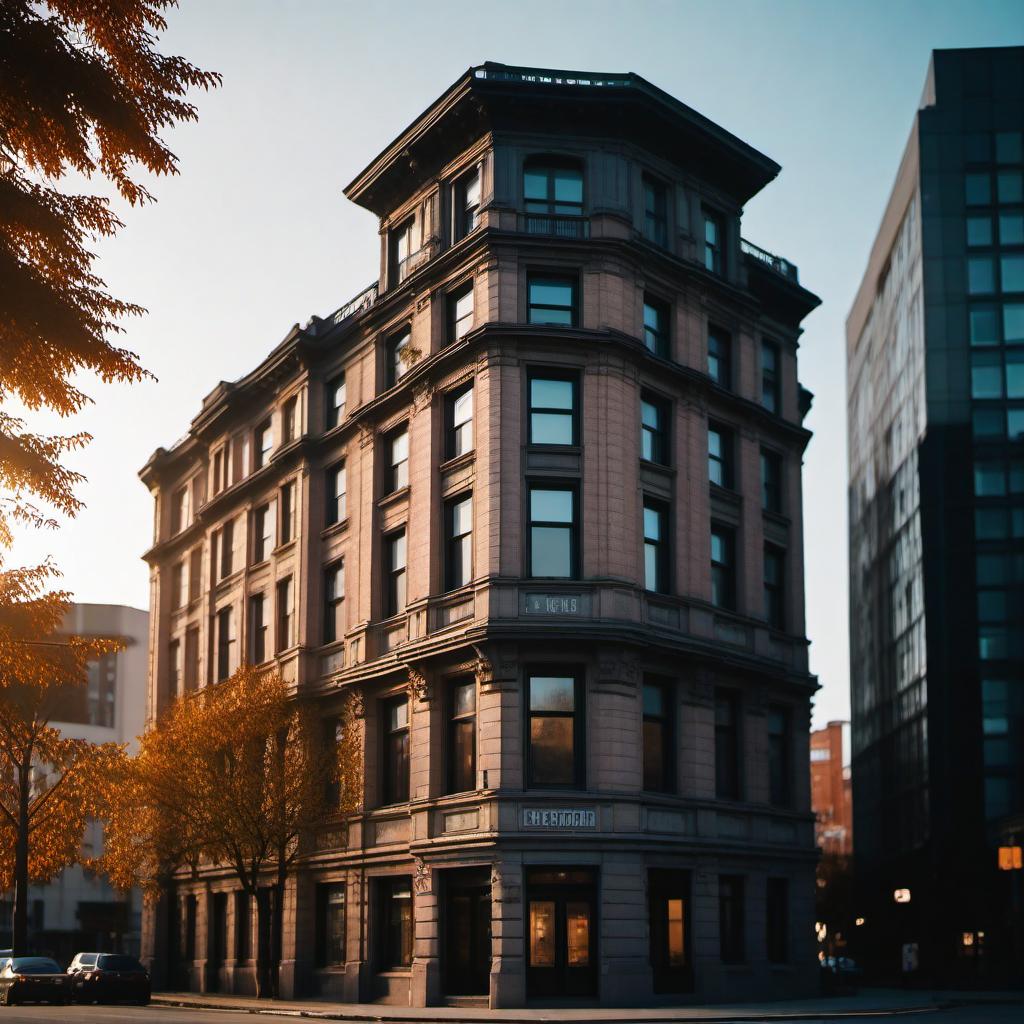}
    \hspace{-5pt}
  \includegraphics[width=0.24\linewidth]{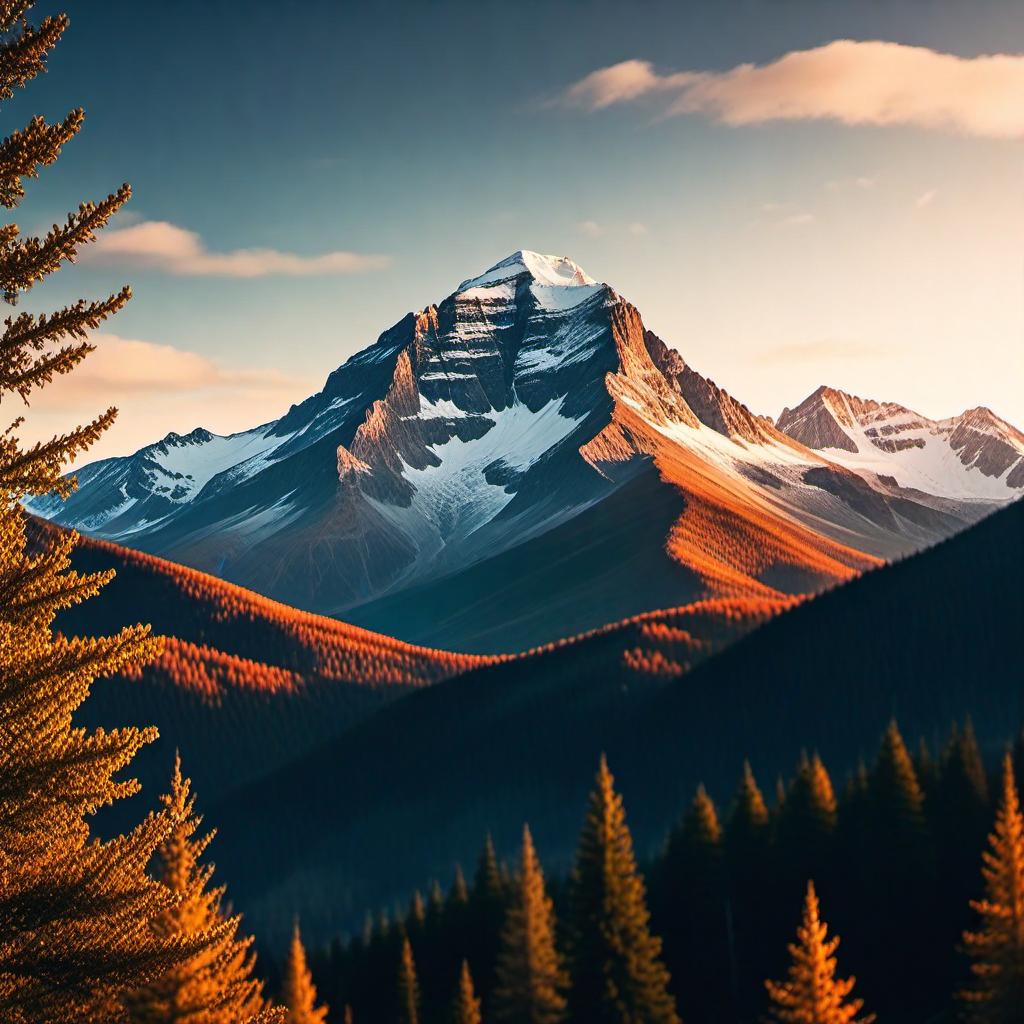}
    \hspace{-5pt}
  \includegraphics[width=0.24\linewidth]{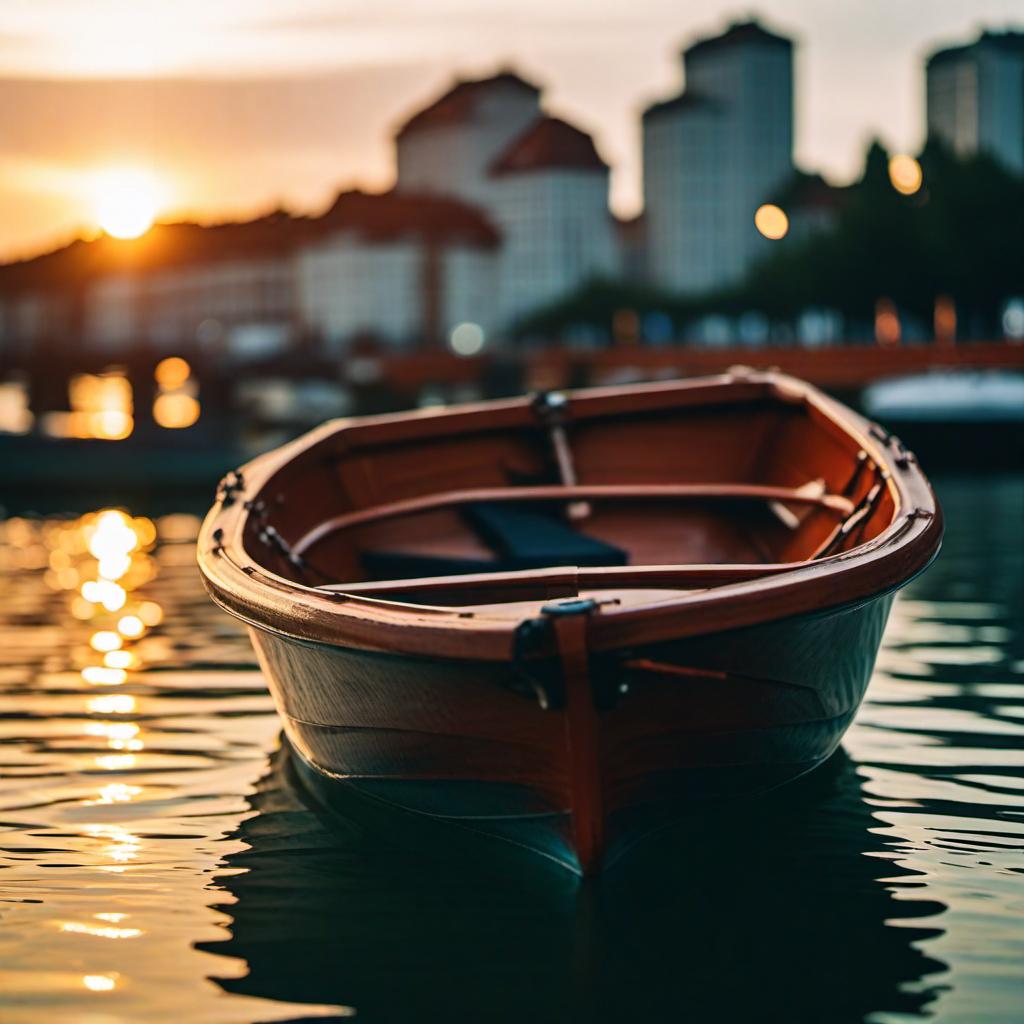}
  \vspace{-1pt}
  \includegraphics[width=0.24\linewidth]{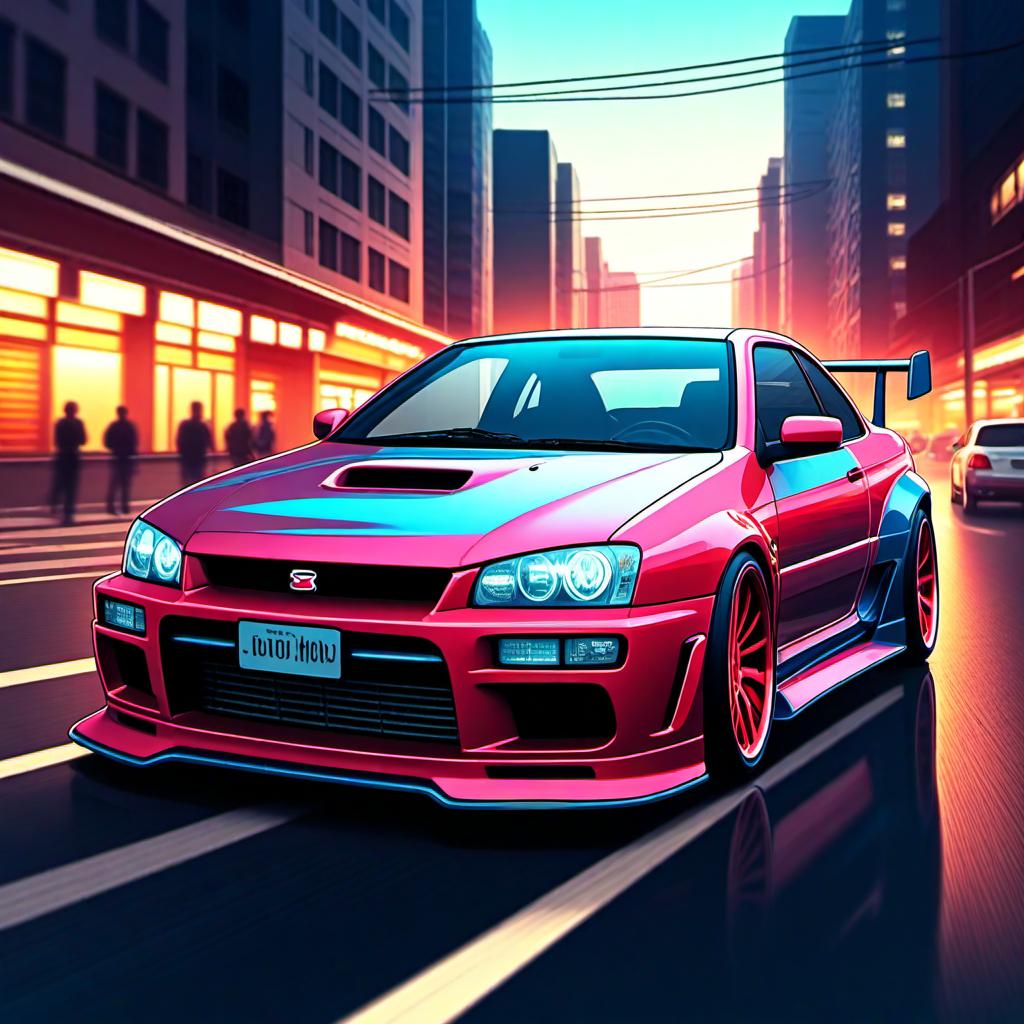}
    \hspace{-5pt}
  \includegraphics[width=0.24\linewidth]{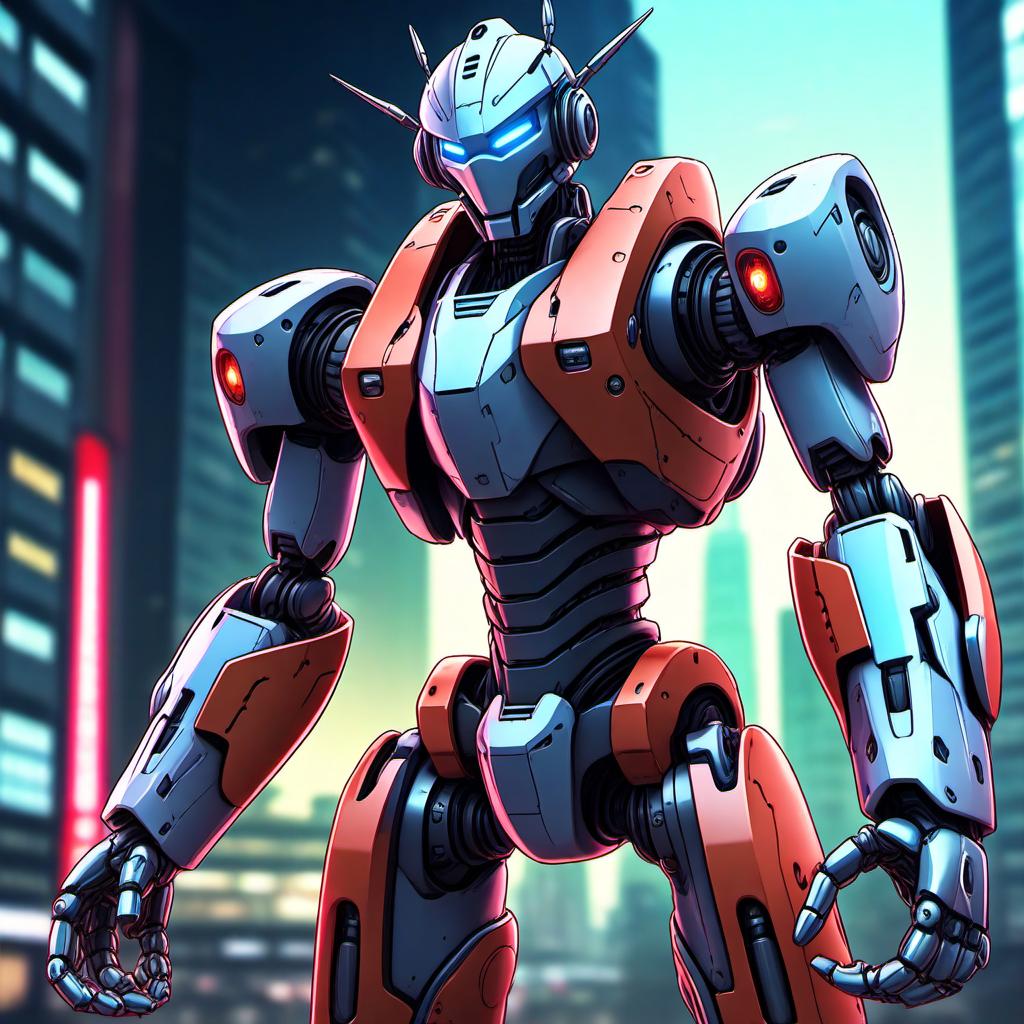}
    \hspace{-5pt}
  \includegraphics[width=0.24\linewidth]{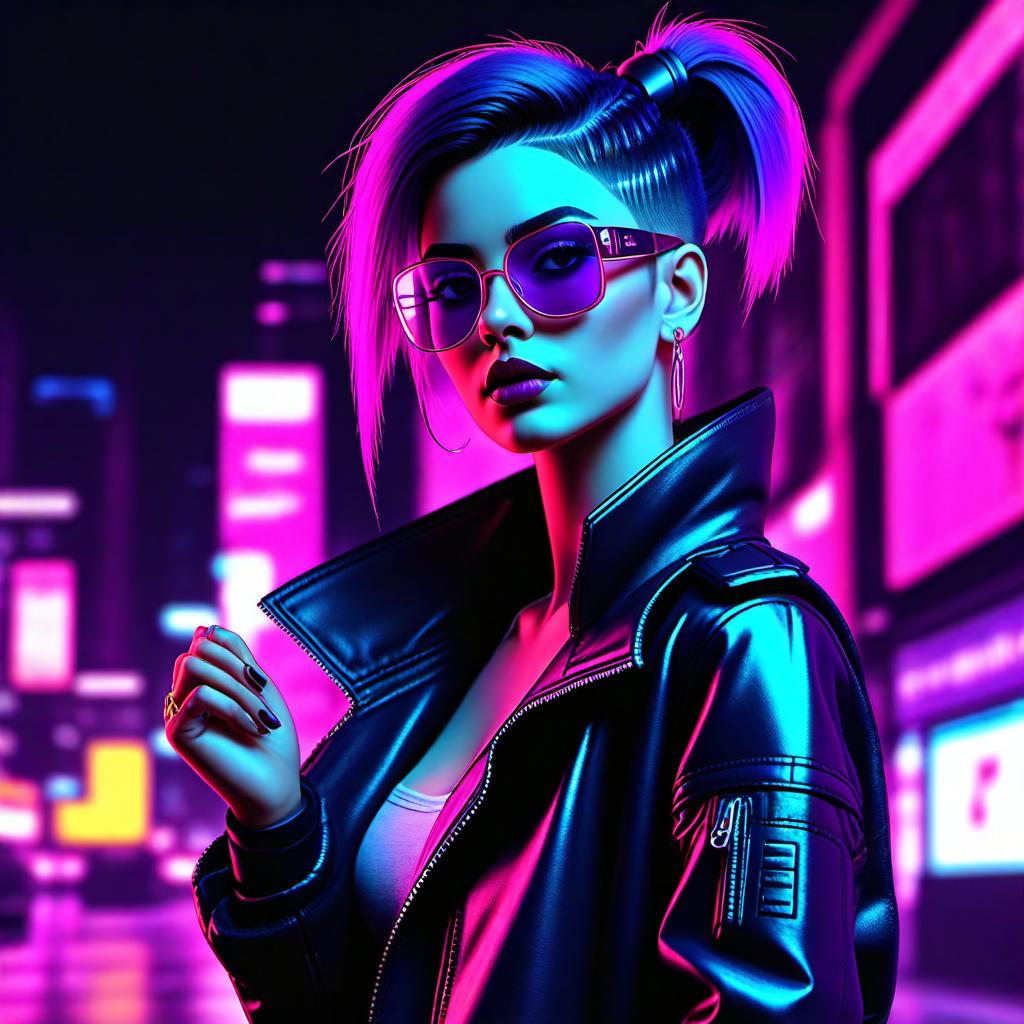}
    \hspace{-5pt}
  \includegraphics[width=0.24\linewidth]{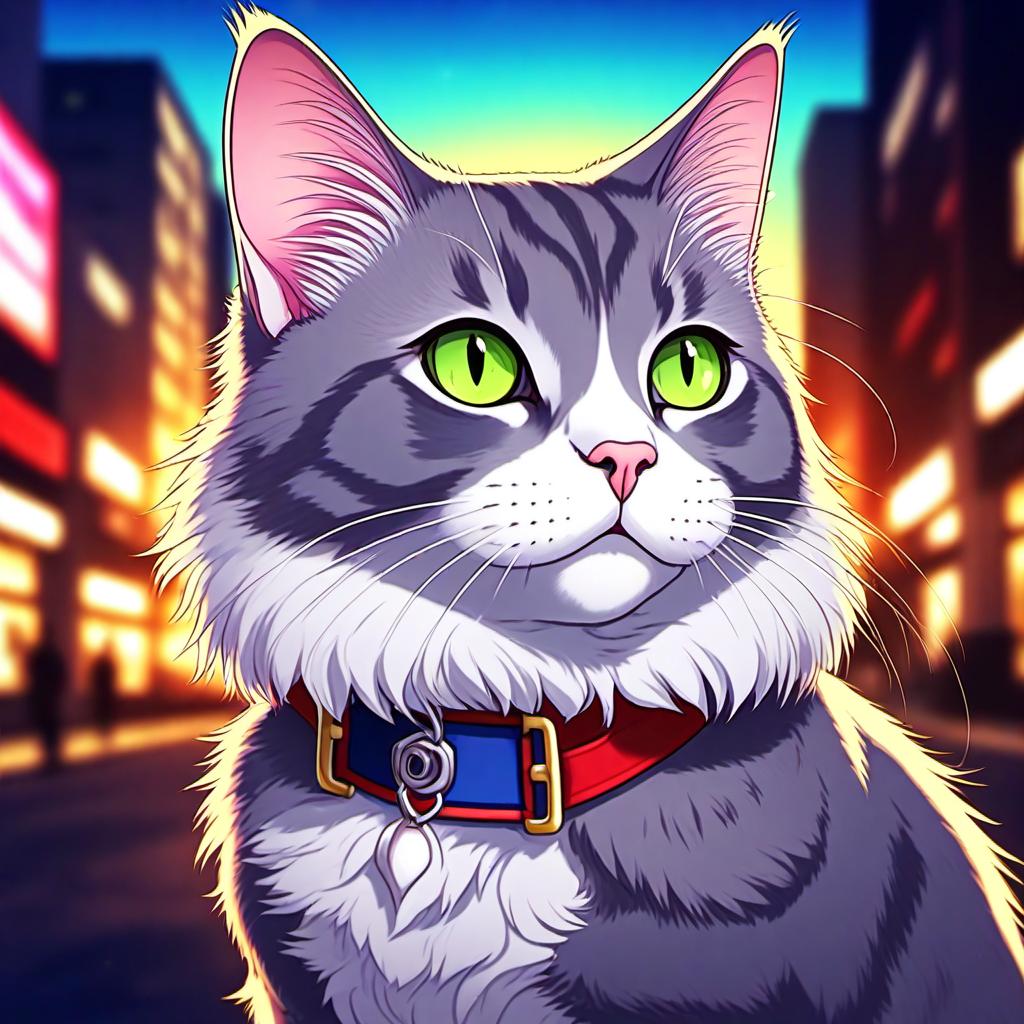}
  \caption{$1024 \times1024$ samples from {TLCM}, distilled from SDXL-base-1.0~\citep{podellsdxl} based on LoRA~\citep{hulora}. 
  From top to bottom, \textbf{2}, \textbf{3}, \textbf{4} and \textbf{8} sampling steps are adopted, respectively.
  Apart from satisfactory visual quality, TLCM can also yield improved metrics compared to strong baselines. 
  }
\end{figure}

\begin{abstract}
Distilling latent diffusion models (LDMs) into ones that are fast to sample from is attracting growing research interest. However, the majority of existing methods face two critical challenges: (\RN{1}) They hinge on long training using a huge volume of real data. 
(\RN{2}) They routinely lead to quality degradation for generation, especially in text-image alignment. 
This paper proposes a novel Training-efficient Latent Consistency Model (TLCM) to overcome these challenges.  
Our method first accelerates LDMs via data-free multistep latent consistency distillation (MLCD), and then data-free latent consistency distillation is proposed to efficiently guarantee the inter-segment consistency in MLCD.  
Furthermore, we introduce bags of techniques, e.g., distribution matching, adversarial learning, and preference learning, to enhance TLCM's performance at few-step inference without any real data.
TLCM demonstrates a high level of flexibility by enabling adjustment of sampling steps within the range of 2 to 8 while still producing competitive outputs compared to full-step approaches.
Notably, TLCM enjoys the data-free merit by employing synthetic data from the teacher for distillation. 
With just 70 training hours on an A100 GPU, a 3-step TLCM distilled from SDXL achieves an impressive CLIP Score of 33.68 and an Aesthetic Score of 5.97 on the MSCOCO-2017 5K benchmark, surpassing various accelerated models and even outperforming the teacher model in human preference metrics. 
We also demonstrate the versatility of TLCMs in applications including image style transfer, controllable generation, and Chinese-to-image generation. 
\end{abstract}

\section{Introduction}

Diffusion models (DMs) have made great advancements in the field of generative modeling, becoming the go-to approach for image, video, and audio generation~\citep{ho2020denoising,kongdiffwave,saharia2022photorealistic}. 
Latent diffusion models (LDMs) further enhance DMs by operating in the latent image space, pushing the limit of high-resolution image and video synthesis~\citep{ma2024latte,peebles2023scalable,podellsdxl,rombach2022high}. 
Despite the high-quality and realistic samples, LDMs suffer from frustratingly slow inference--generating a single sample requires tens to hundreds of evaluations of the model, giving rise to a high cost and bad user experience.

There is growing interest in distilling large-scale LDMs into more efficient ones. 
Concretely, progressive distillation~\citep{lin2024sdxl,meng2023distillation,salimansprogressive} reduces the sampling steps by half in each turn but finally hinges on a set of models for various sampling steps. 
InstaFlow \citep{liu2023instaflow}, UFO-Gen \citep{xu2024ufogen}, DMD~\citep{yin2024one},  and ADD \citep{sauer2023adversarial} target one-step generation, yet losing or weakening the ability to benefit from more (e.g., $>4$) sampling steps. Latent consistency models (LCMs) \citep{luo2023latent} apply consistency distillation~\citep{song2023consistency} on LDMs' reverse-time ordinary differential equation (ODE) trajectories to conjoin one- and multi-step generation, but the image quality degrades substantially, especially in 2-4 steps. 
HyperSD~\citep{ren2024hyper} applies consistency trajectory distillation~\citep{kim2023consistency} in segments of the ODE trajectory, yet suffers from a substantial performance drop in text-image alignment. 
Besides, all these methods rely on a huge volume of high-quality data and long training time, hindering their applicability to downstream scenarios with rare computing and data. 

Before presenting our proposal, we argue that one-step generation may not always be the optimal choice in practical applications. Empirically, sampling with 2-8 steps introduces less than 2 times additional computational time compared to one step for SDXL~\citep{podellsdxl} but can notably enhance the upper limit of sampling quality. 
Moreover, some practical applications typically have a low tolerance for quality degradation and hence can accept a moderate number of sampling steps. 
Thereby, this paper aims to develop a unified model with 2 to 8 sampling steps while achieving quality comparable to full-step models. 
We propose \textbf{T}raining-efficient \textbf{L}atent \textbf{C}onsistency \textbf{M}odels (TLCMs) to achieve this at the expense of minimal computation and training data. 
Technically,
we introduce data-free multistep latent consistency distillation (MLCD) to reduce the sampling steps.
After MLCD, we further employ data-free latent consistency distillation (LCD) for global consistency. 
To enhance LCD, we enforce the consistency of TLCM at sparse predefined timesteps instead of the entire timestep range, which reduces LCD's learning difficulty and accelerates convergence.  
A multistep solver is further explored to unleash the potential of the teacher in LCD.
Besides, we train a latent LPIPS model to constrain the perceptual consistency of the distilled model in latent space. 
To optimize TLCM's performance at few-step inference, we explore 
preference learning, distribution matching, and adversarial learning techniques for regularization in a data-free manner.  

We have performed comprehensive empirical studies to evaluate TLCMs. 
We first assess the image quality on the MSCOCO-2017 5K benchmark. 
We observe the TLCM distilled from SDXL~\citep{podellsdxl} gets an Aesthetic Score (AS)~\citep{AP} of 5.97, and a CLIP Score (CS)~\citep{hessel2021clipscore} of 33.68 with only 3 steps, substantially surpassing 4-step LCM, 8-step SDXL-Lightning~\citep{lin2024sdxl}, and 8-step HyperSD, comparable to 25-step DDIM.  
Additionally, TLCM is obtained by only 70 A100 training hours without any real data, significantly reducing training costs. 
We also demonstrate the versatility of TLCMs in applications including image stylization, controllable generation, and Chinese-to-image generation. 

We summarize our contributions as follows:

\begin{itemize}
    \item We propose TLCMs to accelerate LDMs to generate high-quality outputs within $2-8$ steps, at the expense of minimal training compute and data. 
    \item We establish a data-free multistep latent consistency distillation and improved latent consistency distillation pipeline for fast LDM acceleration. 
    Besides, bags of data-free techniques are incorporated to boost rare-step quality. 
    
    \item TLCM achieves a state-of-the-art CS (33.68) and AS (5.97) in 3 steps, surpassing competing baselines, such as 4-step LCM, 8-step SDXL-Lightning, and 8-step HyperSD.
   \item  TLCMs' versatility extends to scenarios such as image stylization, controllable generation, and Chinese-to-image generation, paving the path for extensive practical applications. 
\end{itemize}

\section{Related works}

\textbf{Diffusion models.} (DMs)~\citep{ho2020denoising,sohl2015deep,song2019generative,song2020improved,songscore} progressively add Gaussian noise to perturb the data, then are trained to denoise noise-corrupted data.
In the inference stage, DMs sample from a Gaussian distribution and perform sequential denoising steps to reconstruct the data.
As a type of generative model, they have demonstrated impressive capabilities in generating realistic and high-quality outputs in text-to-image generation~\citep{podellsdxl,rombach2022high,saharia2022photorealistic}, video generation~\citep{peebles2023scalable}.
To enhance the condition awareness in conditional DMs, the classifier-free guidance (CFG)~\citep{ho2021classifier} technique is proposed to trade off diversity and fidelity.

\textbf{Diffusion acceleration.}
The primary challenges that hinder the practical adoption of DMs is the slow inference involving tens to hundreds of evaluations of the model.

Early works like Progressive Distillation (PD)~\citep{salimansprogressive} and Classifier-aware Distillation (CAD)~\citep{meng2023distillation} explore the approaches of progressive knowledge distillation to compress sampling steps but lead to blurry samples within four sampling steps.  
Consistency models (CMs)~\citep{song2023consistency},
consistency trajectory models (CTMs)~\citep{kim2023consistency} and Diff-Instruct~\citep{luo2024diff} distill a pre-trained DM into a single-step generator, but they do not verify the effectiveness on large-scale text-to-image generation.

Recently, the distillation of large-scale text-to-image DMs has gained significant attention.
LCM~\citep{luo2023latent} extends CM to text-to-image generation with few-step inference but synthesizes blurry images in four steps.
InstaFlow~\citep{liu2023instaflow}, UFOGen~\citep{xu2024ufogen}, BOOT~\citep{gu2023boot}, SwiftBrush~\citep{nguyen2024swiftbrush}, DMD~\citep{yin2023one}, and Diffusion2GAN~\citep{diff2gan}  propose one-step sampling for text-to-image generation but are unable to extend their sampler to multiple steps for better image quality. 

More recently, SDXL-Turbo~\citep{sauer2023adversarial}, SDXL-Lighting~\citep{lin2024sdxl}, and HyperSD~\citep{ren2024hyper} are proposed to further enhance the image quality with a few-step sampling but their training depends on huge high-quality text-image pairs and expensive online training.
Our method not only enables the generation of high-quality samples using a 2-8 steps sampler but also enhances model performance with more inference cost.
Furthermore, our training strategy is resource-efficient and does not require any images.

\textbf{Human preference for text-to-image model.}
ImageReward (IR)~\citep{xu2024imagereward} and Aesthetic Score ~\citep{AP} are proposed to evaluate the human preference for the text-to-image model. Multi-dimensional Preference Score (MPS) ~\citep{zhang2024learning} improves metrics by learning diverse preferences.  To optimize TLCM towards human preference, we incorporate effective reward learning into TLCM  to directly guide model tuning.
\section{Preliminary}
\vspace{-1ex}
\subsection{Diffusion Models}
\vspace{-1ex}
Diffusion models (DMs)~\citep{ho2020denoising,sohl2015deep,songscore} are specified by a predefined forward process that progressively distorts the clean data $x_0$ into a pure Gaussian noise with a Gaussian transition kernel. 
It is shown that such a process can be described by the following stochastic differential equation (SDE)~\citep{karras2022elucidating,songscore}:
\begin{equation}\label{eq:forward}
    dx_t=f(x,t)x_tdt+g(t)dw_t, 
\end{equation}
where $t \in [0,T]$, $w_t$ is the standard Brownian motion, and $f(x,t)$ and $g(t)$
are the drift and diffusion coefficients respectively.
Let $p_t(x_t)$ denote the corresponding marginal distribution of $x_t$. 

It has been proven that this forward SDE possesses the identical marginal distribution as the following probability flow (PF) ordinary differential equation (ODE)~\citep{songscore}:
\begin{equation}
    dx_t=\left[f(x,t)x_t - \frac{1}{2}g^2(t)\nabla{x_t}\log p_t(x_t)\right]dt. 
\end{equation}
As long as we can learn a neural model $\epsilon_\theta(x_{t},{t})$ for estimating the ground-truth score $\nabla{x_t}\log p_t(x_t)$, 
we can then draw samples that roughly follow the same distribution as the clean data by solving the diffusion ODE. 
In practice, the learning of $\epsilon_\theta(x_{t},{t})$ usually boils down to score matching~\citep{songscore}. 

The ODE formulation is appreciated due to its potential for accelerating sampling and has sparked a range of works on specialized solvers for diffusion ODE~\citep{lu2022dpm,lu2022dpm++,songdenoising}. 
Let $\Psi$ denote an ODE solver, e.g., the deterministic diffusion implicit model (DDIM) solver~\citep{songdenoising}.
The sampling iterates by 
\begin{equation}\label{eq:phi}
    x_{t_{n-1}} = \Psi(\epsilon_\theta(x_{t_n},{t_n}),t_{n},t_{n-1}),
\end{equation}
where $\{t_n\}_{n=0}^N$ denotes a set of pre-defined discretization timesteps and $t_N=T, t_0=0$. 

\subsection{Consistency Models}
Consistency model (CM)~\citep{songimproved,song2023consistency} aims at generating images from Gaussian noise within one sampling step. 
Its core idea is to learn a model $f_\theta(x_t,t)$ that directly maps any point $x_t$ on the trajectory of the diffusion ODE to its endpoint. 
To achieve this, CMs first parameterizes $f_\theta(x_t,t)$ as:
\begin{equation}
    f_\theta(x_t,t)=c_{skip}(t)x_t + c_{out}(t)F_\theta(x_t, t),
\end{equation}
where $c_{skip}(t), c_{out}(t)$ is pre-defined to guarantee the boundary condition $f_\theta(x_0,0) = x_0$, and $F_\theta(x_t,t)$ is the neural network (NN) to train.  

CM can be learned from a pre-trained DM $\epsilon_{\theta_0}$ via consistency distillation (CD) by minimizing~\citep{song2023consistency}: 
\begin{equation}
    \mathcal{L}_{CD}=d\big(f_\theta(x_{t_m},t_m),f_{\theta^-}(x_{t_n},t_n)\big),
\end{equation}
where $t_m \sim \mathcal{U}[0, T]$, $x_{t_m} \sim p_{t_m}(x_{t_m})$, $t_n \sim \mathcal{U}[0, {t_m})$, $x_{t_n} = \Psi(\epsilon_{\theta_0}(x_{t_m},{t_m}),t_{m},t_{n})$, $d(.,.)$ is some distance function, and $\theta^-$ is the exponential moving average (EMA) of $\theta$. 
Typically, $x_{t_n}$ is obtained by a single-step solver (SS) $\Psi$.

 Multistep consistency models (MCMs)~\citep{heek2024multistep} generalize CMs by splitting the entire range $[0, T]$ into multiple segments and performing consistency distillation individually within each segment. 
Formally, MCMs first define a set of milestones $\{t_\mathrm{step}^s\}_{\mathrm{s} = 0}^M$ ($M$ denotes the number of segments), and minimize the following multistep consistency distillation (MCD) loss:
\begin{equation}
\label{eq:mcd}
\mathcal{L}_{MCD}=d\big(\mathrm{DDIM}(x_{t_m}, f_\theta(x_{t_m},t_m),t_m, t_\mathrm{step}^s), \mathrm{DDIM}(x_{t_n}, f_{\theta^-}(x_{t_n},t_n),t_n, t_\mathrm{step}^s )\big),
\end{equation}
where $s$ is uniformly sampled from $\{0, \dots, M\}$, $t_m \sim \mathcal{U}[t_\mathrm{step}^s, t_{\mathrm{step}}^{s+1}]$, $t_n =t_m-1$, and $\mathrm{DDIM}(x_{t_m}, f_\theta(x_{t_m},t_m),t_m, t_\mathrm{step}^s)$ means one-step DDIM transformation from state $x_{t_m}$ at timestep $t_m$ to timestep $t_\mathrm{step}^s$ based on the estimated clean image $f_\theta(x_{t_m},t_m)$~\citep{songdenoising}.

\begin{figure}[t]
    \centering
    \includegraphics[width=1\linewidth]{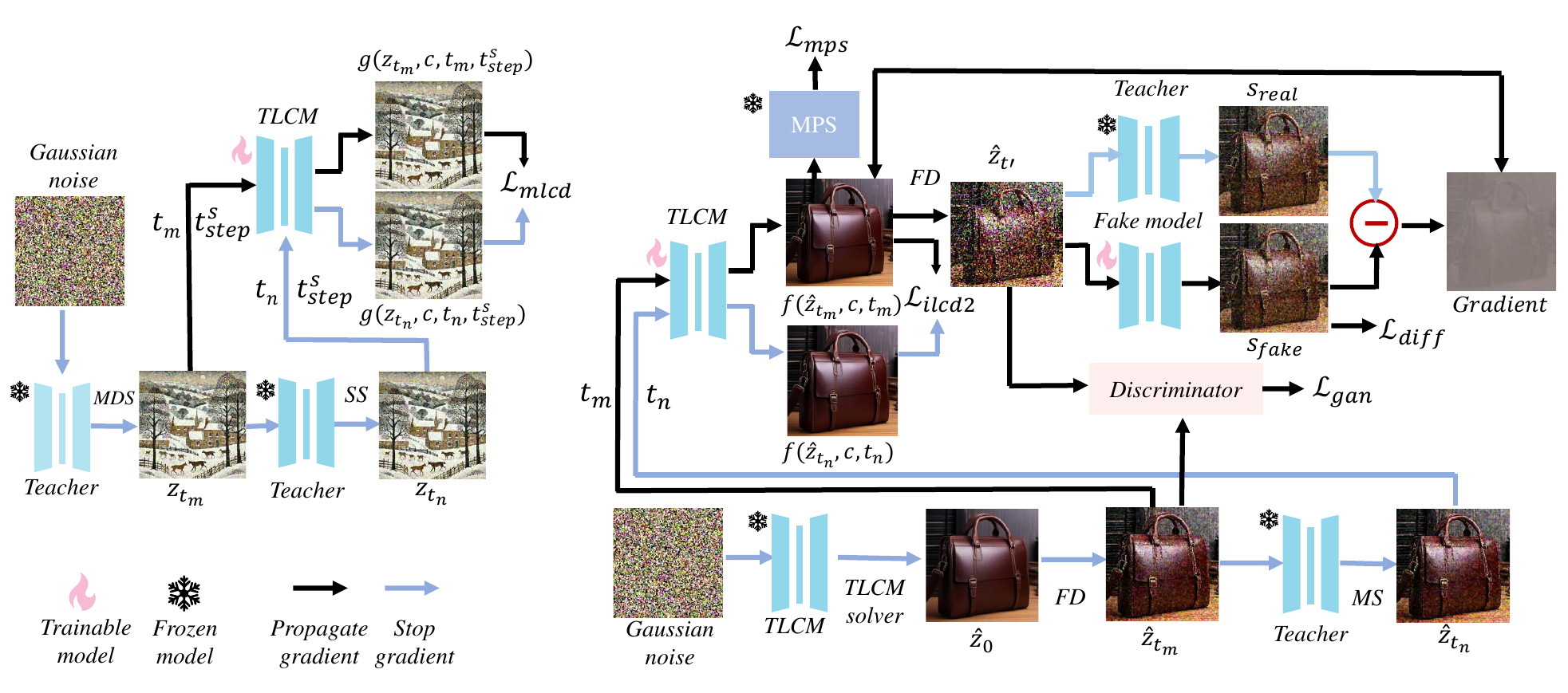}
    \caption{The overview for training TLCM. Data-free multistep latent consistency distillation is first used to accelerate LDM, obtaining initial TLCM (left part of the overview). Then, improved data-free latent consistency distillation is proposed to enforce the global consistency of TLCM.  MPS optimization, DM, and adversarial learning are exploited to promote TLCM's performance in a data-free manner (right part of the overview). Note that we omit the Latent LPIPS model for brevity. }\label{fig:sys}
 \end{figure}

\section{Methodology}
Our target is to accelerate LDM into a few-step model, with performance competitive to the full-step teacher. 
The learning procedure should be executed  with 
cheap cost in a data-free manner.  
In this section, we propose a novel and unified Training-efficient Latent Consistency Model (TLCM) with  2-8 step inference.  
We begin by introducing data-free multistep latent consistency distillation. 
Subsequently, we discuss data-free latent consistency distillation to enforce the global consistency of TLCM. 
Lastly, we explore various techniques to promote TLCM's performance in a data-free manner. The overview of our training pipeline is presented in Figure~\ref{fig:sys}.

\subsection{Data-free Multistep Latent Consistency Distillation}
We consider distilling representative pre-trained LDMs, e.g., SDXL~\citep{podellsdxl}.
Previous LCM~\citep{luo2023latent} has distilled SDXL into a few-step model, but it results in a big performance drop since it is hard to learn the mapping between an arbitrary state of the entire ODE trajectory to the endpoint. 
Drawing inspiration from MCM, we split the entire range [0, $T$]
into $M$ segments, and then only enforce the consistency at each separate segment. 
To speed up convergence, we change the skipping step ($skip$) =1 in MCM into 20. 
The EMA module is removed to save memory consumption.   
Let $z_t$ denote the states at timestep $t$ in the latent space. 
We abuse $\epsilon_{\theta_0}(z_t, c, t)$ and $f_\theta(z_t, c, t)$ to denote the pre-trained LDM and target model respectively, where $c$ refers to the generation condition.  
We formulate the multistep latent consistency distillation (MLCD) loss as:
\begin{equation}
    \label{eq: mlcd}
    \mathcal{L}_{mlcd}=\Vert g_{\theta}(z_{t_m},t_m,t_{step}^s,c)- \text{nograd}(g_{\theta}(z_{t_n},t_n,t_{step}^s,c)\Vert_2^2, 
\end{equation}
where  $g_{\theta}(z_{t_m},t_m,t_{step}^s,c)= \mathrm{DDIM}\big(z_{t_m},f_{\theta}(z_{t_m},c,t_m),t_m,t_\mathrm{step}^s\big)$ represents initial TLCM.
Given that CFG~\citep{ho2021classifier} is critical for high-quality text-to-image generation, we integrate it to  MLCD by:
\begin{equation}\label{eq:ddim}
    z_{t_n}=\Psi(\hat{\epsilon}_{\theta_0}(z_{t_m},c,w,t_m),t_m,t_n),  
\end{equation}
where 
$ \hat{\epsilon}_{\theta_0}(z_t,c,w,t):=\epsilon_{\theta_0}(z_t,\emptyset,t) +w(\epsilon_{\theta_0}(z_t,c,t)-\epsilon_{\theta_0}(z_t,\emptyset,t))$ with $w$ as the guidance scale.

However, this training procedure depends on huge high-quality data, which limits its applicability in scenarios where such data is inaccessible. 
To deal with this problem, we propose to draw samples from the teacher model as training data. 
Specifically, instead of obtaining $z_{t_m}$ via adding noise to $z_0$ as in MCM and HyperSD, 
we initialize $z_T$ as pure Gaussian noise $\epsilon$ and perform denoising with off-the-shelf ODE solvers based on the teacher model $\epsilon_{\theta_0}$ to obtain $z_{t_m}$.
Intuitively, with this strategy, we leverage and distill only the denoising ODE trajectory of the teacher without concerning the forward one. 
The latent state $z_{t_m}$ can be acquired from $\epsilon$  by a single denoising step, but we empirically observe that this naive strategy is unable to accelerate LDM with desirable performance, due to poor quality of  $z_{t_m}$. 
Theoretically, $z_{t_m}$  contains less noise for smaller ${t_m}$. Therefore, we design a multistep denoising strategy (MDS) to predict $z_{t_m}$, which executes more sampling iterations for smaller ${t_m}$ to get cleaner $z_{t_m}$. 
At this stage, the DDIM solver is used to estimate the ODE trajectory and generate samples from pure Gaussian noise. 
We present the details in Algorithm~\ref{alg:al1} in Appendix~\ref{alg}.  

\subsection{Improved  Data-free  Latent Consistency Distillation }
After a round of distillation on $M$ segments, TLCM can naturally produce high-quality samples through $M$-step sampling. 
However, it is empirically observed that the performance decreases when using fewer steps, which is probably because of the larger discretization error caused by long sampling step sizes. 
To alleviate this, we advocate explicitly teaching TLCM to capture the mapping between the states that cross segments.  Upon this goal, we propose data-free latent consistency distillation to promote the model to be consistent across the predefined timesteps.  

We do not compile TLCM to keep consistency across the entire timestep range [0, $T$] since it is hard to learn the mapping that transforms any point along the trajectory into real data.  Instead, we improve raw LCD by only keeping consistency at the predefined $M$ timesteps, which makes LCD much easier to learn the mapping. 
Naturally, the skipping step $skip$ is changed to $T/M$. 
The big  $skip$ offers an additional advantage that further accelerates model convergence.  
Benefiting from the pre-trained TLCM, we can fast yield clean data $\hat{z}_0$ via a few-step ($q$-step) sampler, such as 4 steps, eliminating the requirement of real data. 
The latent state $\hat{z}_{t_m}$ is obtained by adding noise to $\hat{z}_0$ in the   forward diffusion process, where $t_m \in \{t_\mathrm{step}^s\}_{\mathrm{s} = 1}^M$. We formulate this procedure as 
\begin{equation}
\label{eq: tlcm}
    \hat{z}_{t_m}=FD(TLCM(\epsilon, T,c),t_m), \quad \epsilon \in \mathcal{N}(0,I),
\end{equation}
where $FD$ and $TLCM$ denote forward diffusion and multistep iterations by TLCM.
Then, an ODE solver is used to estimate latent state $\hat{z}_{t_n}$ from  $\hat{z}_{t_m}$.
Raw LCD adopts a one-step solver to predict $\hat{z}_{t_n}$. 
We argue that it restricts the capability of the teacher due to discretization error, especially for big $skip$.   
As a result, we explore a multistep solver (MS) to unleash the potential of the teacher.  
Concretely, the time interval $T/M$ is uniformly divided into $p$ parts, and then $p$-step DDIM with CFG is used to calculate $\hat{z}_{t_n}$. 
The improved data-free LCD loss in stage 2 is:
\begin{equation}
    \label{eq:ilcd2}
    \mathcal{L}_{ilcd2}=\Vert \big(f_{\theta}(\hat{z}_{t_m},c,t_m))- \text{nograd}\big(f_{\theta}(\hat{z}_{t_n},c,t_n))\big)\Vert_2^2.
\end{equation}
We present the details in Algorithm~\ref{alg:al2} in appendix~\ref{alg}.  
Surprisingly, our improved data-free LCD only costs 2K-iteration training to achieve convergence. 

\subsection{Incorporating Bag of Techniques into TLCM in Data-free Manner}
\textbf{Latent LPIPS.} 
Typical LCD directly adopts mean square error loss ($\mathcal{L}_{mse}$) to enforce consistency in the latent space, but it can not capture perceptual features.  LPIPS~\citep{zhang2018unreasonable}  
can extract the features matching human perceptual responses. 
Meanwhile, it has been widely used as an effective regression loss across many image translation tasks.  
Thereby, we aim to integrate LPIPS into our distillation pipeline to enhance TLCM's performance. 
However, LPIPS  is built in the pixel space, and hence we have to reconstruct latent codes to pixel space to use LPIPS. 
To reduce training time, we train a latent LPIPS (L-LPIPS) model, which 
computes perceptual features in latent space. 
The latent LPIPS model adopts the VGG network by 
changing the input to 4 channels and   removing the 3 max-pooling layers, as the latent space in LDM is already
8× downsampled. 
The model is trained from scratch on BAPPS dataset~\citep{zhang2018unreasonable}.   
Based on  L-LPIPS, the outputs of the model $g_\theta$ and $f_\theta$ are first fed into the L-LPIPS model, whose outputs are used to calculate consistency loss via Equation~(\ref{eq: mlcd}) or Equation~(\ref{eq:ilcd2}).   

\textbf{MPS optimization.}
Since TLCMs transform the points on the trajectory to clean samples $\hat{x}_0$, we can naturally directly maximize the feedback of the scorer on the sample $s(\hat{x}_0, c)$. 
Considering that multi-dimensional preference score~\citep{zhang2024learning} can measure diverse human preferences, we leverage it to improve TLCM towards human preference. 
Formally, we optimize the following MPS loss ($\mathcal{L}_{mps}$):
\begin{equation}
    \mathcal{L}_{mps} = \max(s_0-s(\hat{x}_0,c_{pos}), 0)+\max(s(\hat{x}_0,c_{neg}), 0),
\end{equation}
where $c_{pos}$ represents the text condition corresponding to the images while $c_{pos}$ denotes the irrelevant texts. $\mathcal{L}_{mps}$ maximizes $s(\hat{x}_0, c_{pos})$ with margin $s_0$ and simultaneously minimizes $s(\hat{x}_0, c_{neg})$ with margin 0.
The gradients are directly back-propagated from the scorer to model parameters $\theta$ for optimization. 
We do not use ImageReward or AS to optimize TLCM, because we find IR tends to cause overexposure and AS results in oversaturation for generated images.

\textbf{Distribution matching.} 
Distribution matching~\citep{yin2023one} is proposed to transform LDM into a one-step model. 
We effectively integrate it into our distillation method to enhance the performance of TLCM. 
To remove the need of real data,  we exploit Equation~\ref{eq: tlcm} to get noisy latent  $\hat{z}_t$. 
Data-free DM loss in $\mathcal{L}_{dfdm}$ is applied to optimize TLCM at sparse-step inference as 
\begin{equation}
    \mathcal{L}_{dfdm}=-\mathbb{E}_{t,\epsilon,\hat{z}_t}[s_{real}(FD(f_\theta(\hat{z}_t,t,c),t'))-s_{fake}(FD(f_\theta(\hat{z}_t,t,c),t'))\nabla_{\theta}f_\theta(\epsilon)],
\end{equation}
 where $s_{real}$ and $s_{fake}$ denote the pre-trained score model and fake score model,  both initialized by SDXL.  The model $s_{fake}$ is finetuned on synthetic data $\hat{z}_0$ through noise prediction loss $\mathcal{L}_{diff}$ in DM~\citep{yin2023one}.

\textbf{Adversarial learning.}
For high-resolution text-to-image
generation, considering the high data dimensionality and
complex data distribution, simply using MSE loss  fails to capture data discrepancy precisely, thus providing imperfect consistency
constraints. 
We propose to use GAN loss to enforce the distribution consistency. 
Unlike previous methods needing real data to execute adversarial learning, we exploit Equation~\ref{eq: tlcm} to obtain $\hat{z}_t$. 
The student model  $f_{\theta}$ denoises $\hat{z}_t$ by one step, obtaining $\widetilde{z}_0$.  
Through discriminator $D$, the GAN loss $\mathcal{L}_{gan}$ is formulated  as 
\begin{equation}
    \mathcal{L}_{gan}=\log(D(FD(\hat{z}_0,t'))-\log(D(FD(\widetilde{z}_0,t'))).
\end{equation}

\section{Experiments} 
\subsection{Implementation Details}
We use the prompts from LAION-Aesthetics- 6+ subset of LAION-5B~\citep{schuhmann2022laion} to train our model.  We train the
model with 12000 iterations for data-free MLCD and 2000 iterations for data-free LCD.  After LCD,  MPS optimization runs 500 iterations with a batch size of 8. Then, DM and adversarial learning are used to improve TLCM with 1000 iterations with a batch size of 4. 
The whole procedure uses AdamW optimizer and 4 A100. Only MLCD adopts a learning rate of 1e-4 and the other stages use a learning rate of 1e-5. The discriminator adopts a learning rate of 1e-4 and AdamW optimizer.
The initial segment number $M$ is 8 and $s_0$ for MPS is 16. We set the guidance scale $w$ in CFG as 8.0, the denoising steps $p=3$ for the teacher to compute $\hat{z}_{t_n}$, and $q=4$ for TLCM to compute $\hat{z}_0$. 
As for model configuration, we use SDXL~\citep{podellsdxl} as teacher to estimate trajectory while student model $f_\theta$ is also initialized by SDXL. The discriminator is also initialized by SDXL. 
We train a unified Lora instead of UNet in all the distillation stages for convenient transfer to downstream applications.

\subsection{Main Results}

We quantitatively compare our
method with both the DDIM~\citep{songdenoising} baseline and acceleration
approaches including LCM~\citep{luo2023latent}, SDXL-Turbo~\citep{sauer2023adversarial}, SDXL-Lightning~\citep{lin2024sdxl}, HyperSD~\citep{ren2024hyper}, 
CS~\citep{hessel2021clipscore} with ViT-g/14 backbone, 
AS~\citep{AP}, IR~\citep{xu2024imagereward}, and Fréchet Inception Distance (FID) are exploited as objective metrics.  
The evaluation is performed on MSCOCO-2017 5K validation dataset~\citep{lin2014microsoft}. 
All methods perform zero-shot validation except for HyperSD since it utilizes the MSCOCO-2017 dataset for training.  
Only SDXL-Turbo produces 512-pixel images while the others generate 1024-pixel images.  We only report FID for reference and do not analyze it since FID on COCO is not reliable for evaluating text-to-image models~\citep{sauer2023adversarial,ren2024hyper}.

\begin{table}[t]
\caption{Zero-shot performance comparison on MSCOCO-2017 5K validation datasets with the state-of-the-art methods. All models adopt SDXL architecture. Time: inference time (second) on A100.  TH: Training hours using A100. TI: Training images. }
\label{sdxl}
\begin{center}
\begin{tabular}{lllllllll}
\\ \hline 
Method  &Step & Time  &FID & CS   & AS & IR &TH & TI\\
\hline
DDIM~\citep{songdenoising} & 25 &3.29&23.29 &33.97 &5.87&0.82 &0&0 \\
LCM~\citep{luo2023latent} & 4 &0.71&27.09 &32.53&5.85&0.51&-&-\\
SDXL-Turbo~\citep{sauer2023adversarial} &4 &0.38&28.52&33.35&5.64&0.83&-&- \\
SDXL-Turbo~\citep{sauer2023adversarial} &8 &0.61&29.64&32.81&5.78&0.82&-&- \\
SDXL-Lightning~\citep{lin2024sdxl}&4 &0.71&27.90& 32.90&5.63&0.72 &-&$>$12M \\
SDXL-Lightning~\citep{lin2024sdxl}&8 &0.99 &27.04 &32.74&5.95&0.71 &-&$>$12M \\
HyperSD~\citep{ren2024hyper}&4 &0.71 &34.45 &32.64&5.52&1.15 &600&$>$12M \\
HyperSD~\citep{ren2024hyper}&8 &0.99 &35.94 &32.41&5.83&1.14 &600&$>$12M \\
TLCM &2&0.58&27.50 & 33.18 &5.90&0.97&70&0\\ 
TLCM &3&0.65&29.12 & 33.68 &5.97 &   1.00&70&0\\
TLCM &4&0.71&30.33&33.52 & 6.06 & 1.01  &70&0\\
TLCM &5&0.78&30.90 & 33.69 &6.04  & 1.01&70&0\\
TLCM &6&0.85& 30.98& 33.71 & 6.07 & 1.01&70&0\\
TLCM &8&0.99&32.40 & 33.53 &6.08& 1.02   &70&0 \\
\hline
\end{tabular}
\end{center}
\end{table}
The metrics of various methods are listed in Table~\ref{sdxl}. We use ``-" to represent a metric when it is missing in the corresponding paper.  
We can observe that our TLCM  only costs 70 A 100 training hours, even without any data. Compared to other methods, TLCM significantly reduces training resources, which is very valuable for most laboratories and scenarios when real data are inaccessible.   our 3-step TLCM presents superior CS, AS, and IR than 4-8 step's LCM~\citep{luo2023latent}, SDXL-Lightning~\citep{lin2024sdxl}.  
These results indicate our TLCM's synthetic images are much better aligned with texts and human preference than LCM, SDXL-Lightning.  Excitingly, our 3-step TLCM outperforms the 25-step teacher in terms of AS and IR, and achieves comparable CS value, demonstrating that TLCM almost reserves all the information in the teacher and even introduces new human preference knowledge via the proposed distillation method.  
Our 3-step TLCM shows much higher CS than the 4-8 step HyperSD, indicating HyperSD loses much information in the distillation procedure because it fails to sufficiently ensure consistency constraint.  
We notice IR value of HyperSD is higher than our TLCM. 
This is because IR model has been used to optimize HyperSD.  
Moreover, we can see the performance of SDXL-Turbo drop with respect to CS and  IR  when increasing sampling steps. 
This is because it is designed for specific steps.  
Instead, our TLCM can improve at least one metric with additional steps. 
This is valuable since image quality is the primary consideration when affordable computation resource is determined in real applications. 

\begin{figure}
    \centering
    \parbox{0.16\linewidth}{\small \centering DDIM\\Step=25}
    \parbox{0.16\linewidth}{\small \centering LCM\\Step=4}
    \parbox{0.16\linewidth}{\small \centering SDXL-Turbo\\Step=4}
    \parbox{0.16\linewidth}{\small \centering SDXL-Lightning\\Step=4}
    \parbox{0.16\linewidth}{\small \centering HyperSD\\Step=4}
    \parbox{0.16\linewidth}{\small \centering TLCM (\textbf{Ours})\\Step=4}
    
    \includegraphics[width=0.16\linewidth]{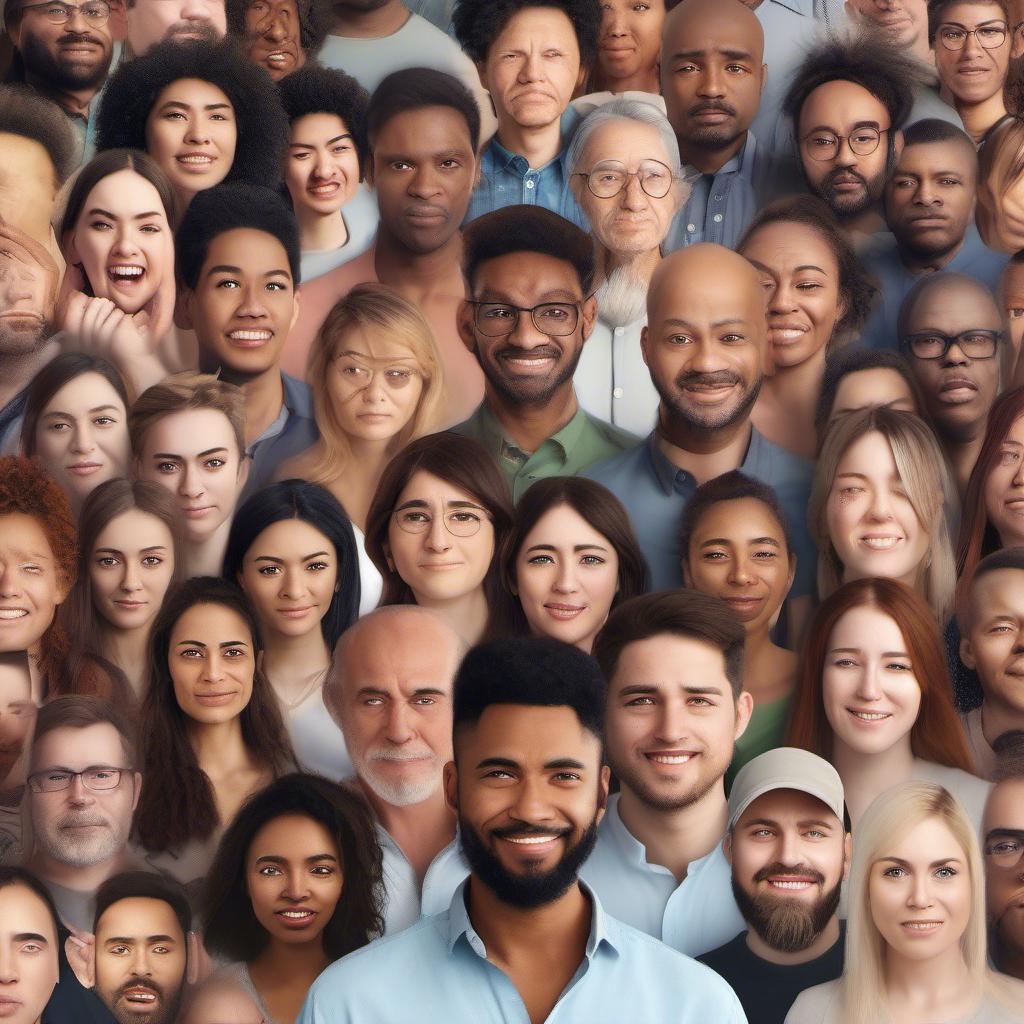}
    \includegraphics[width=0.16\linewidth]{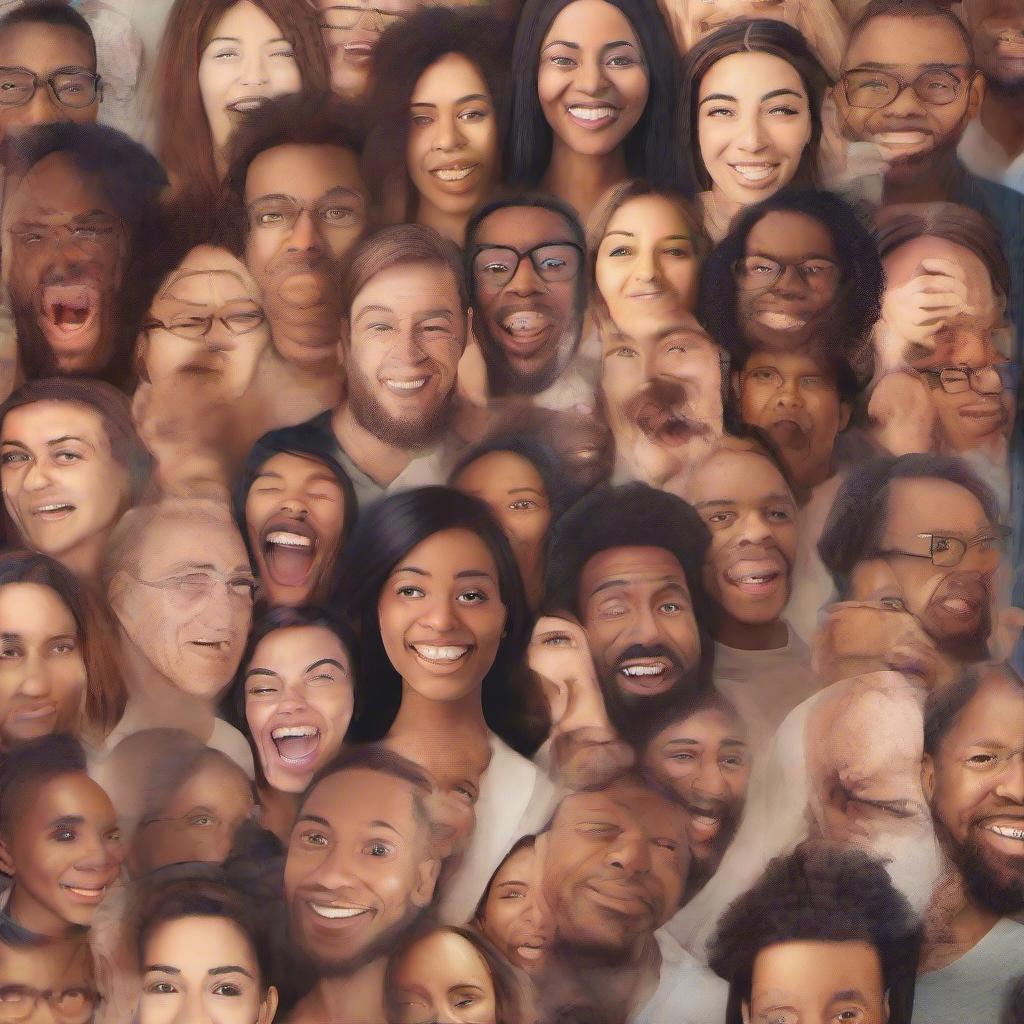}
    \includegraphics[width=0.16\linewidth]{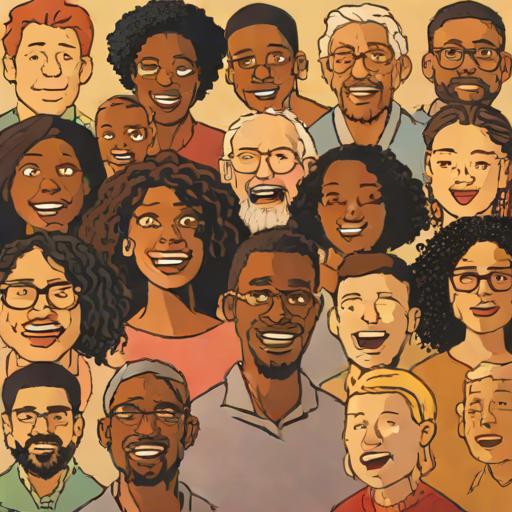}
    \includegraphics[width=0.16\linewidth]{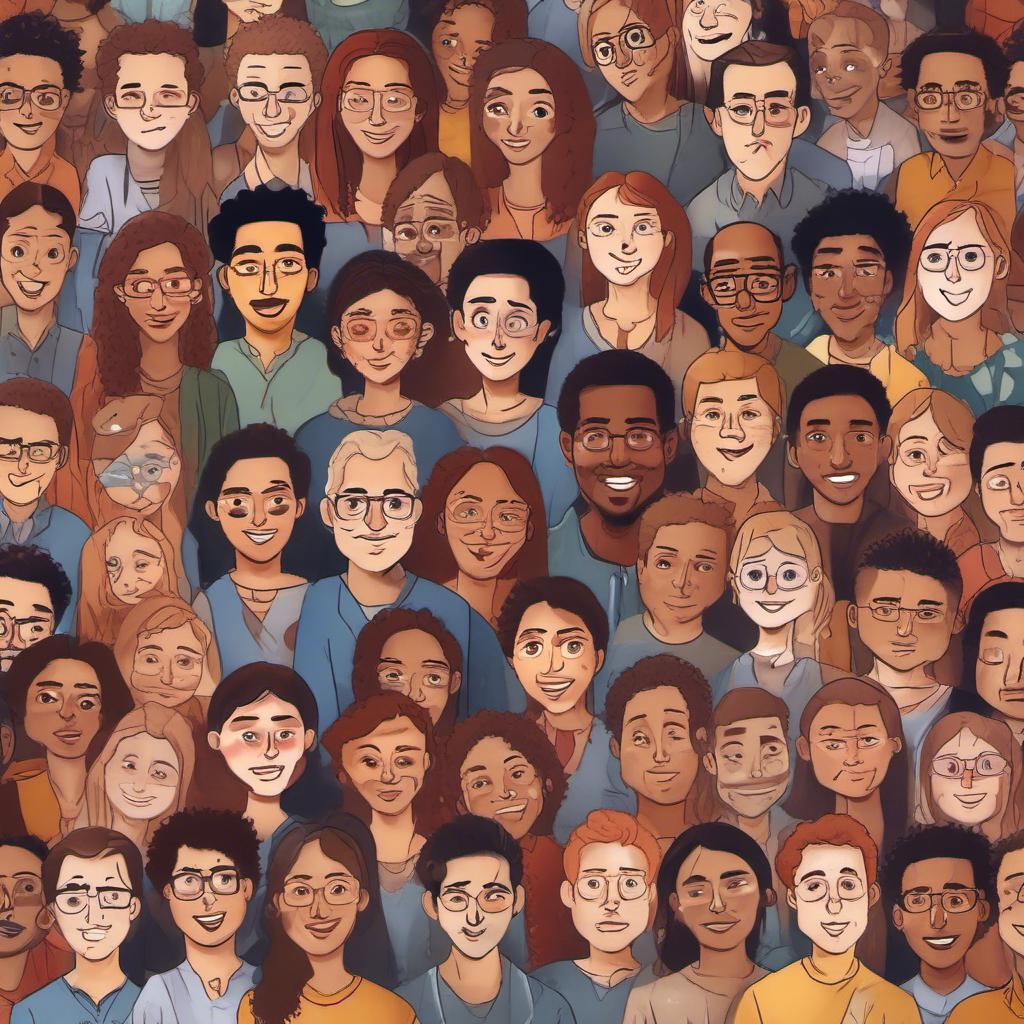}
    \includegraphics[width=0.16\linewidth]{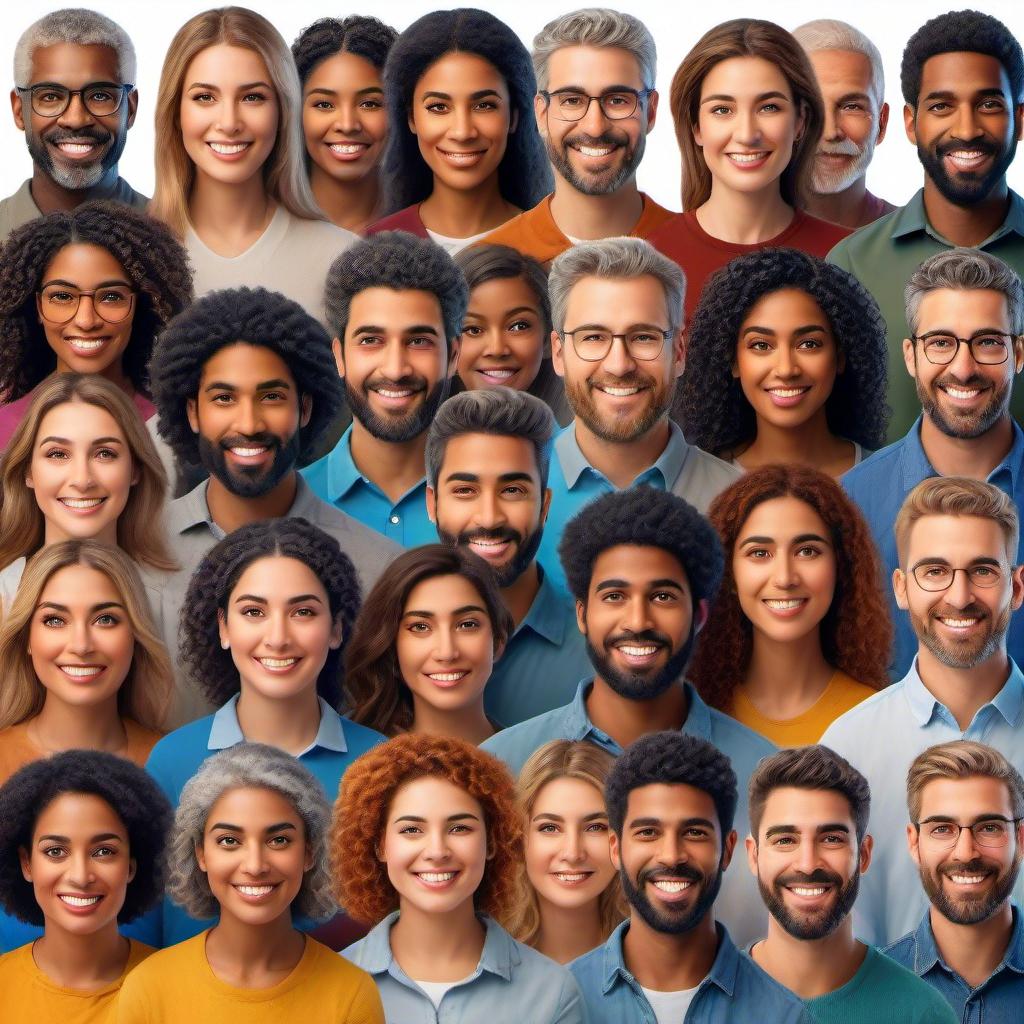}
    \includegraphics[width=0.16\linewidth]{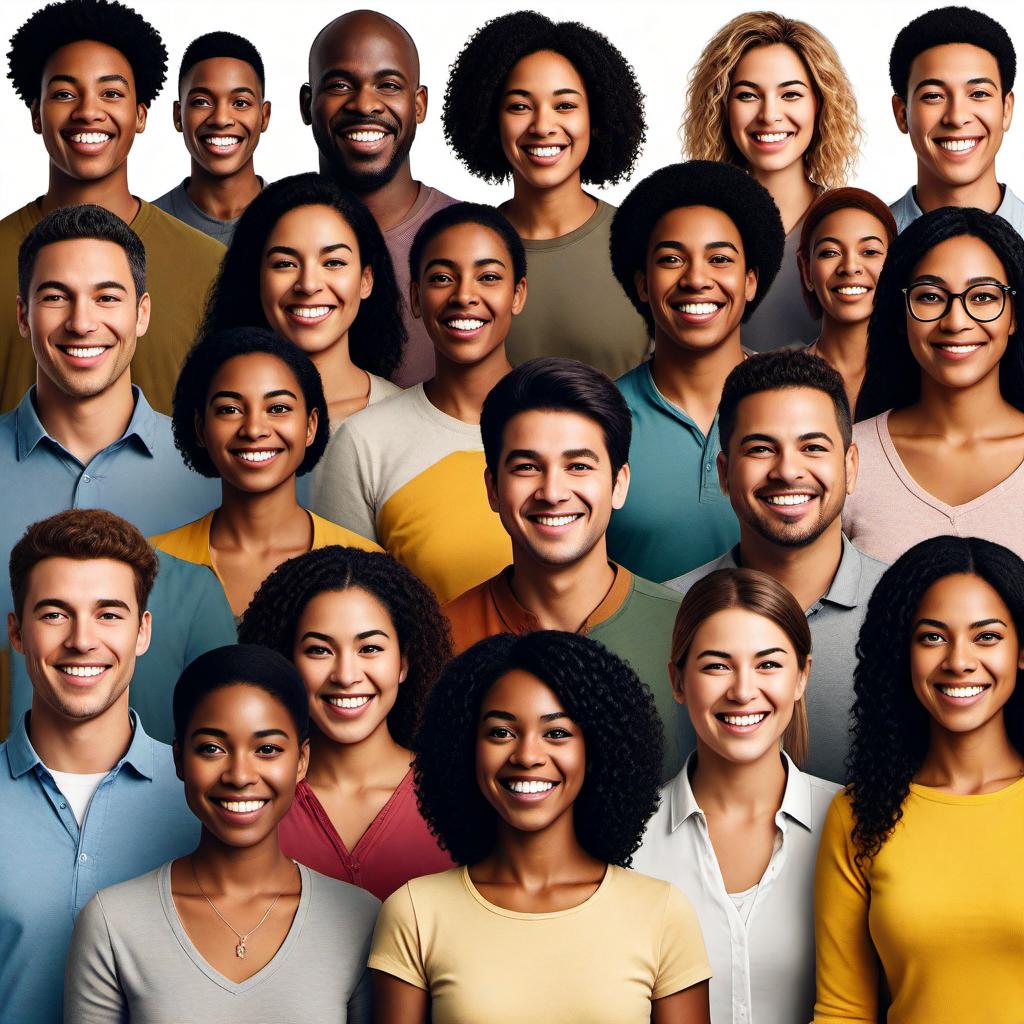}
    \parbox{\linewidth}{\centering a high-resolution image or illustration of a diverse group of people facing me.}

    \includegraphics[width=0.16\linewidth]{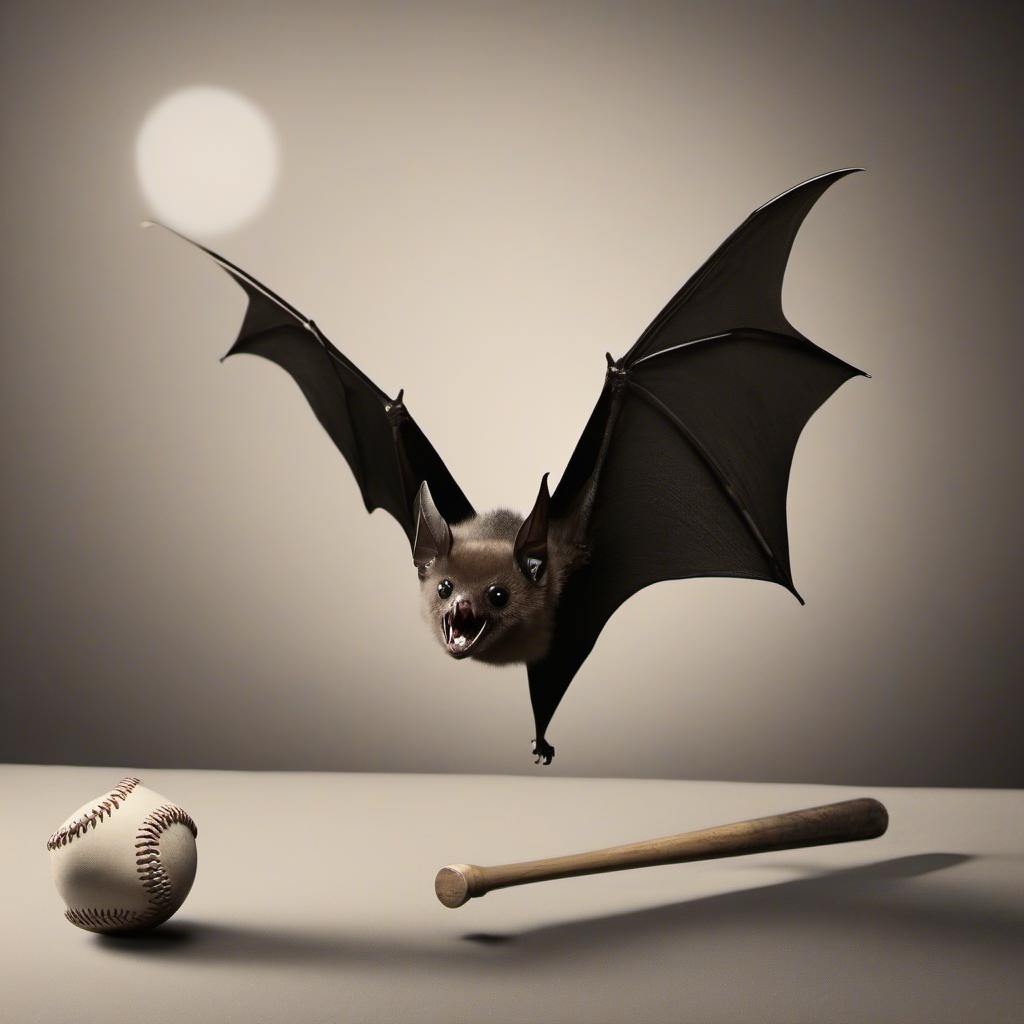}
    \includegraphics[width=0.16\linewidth]{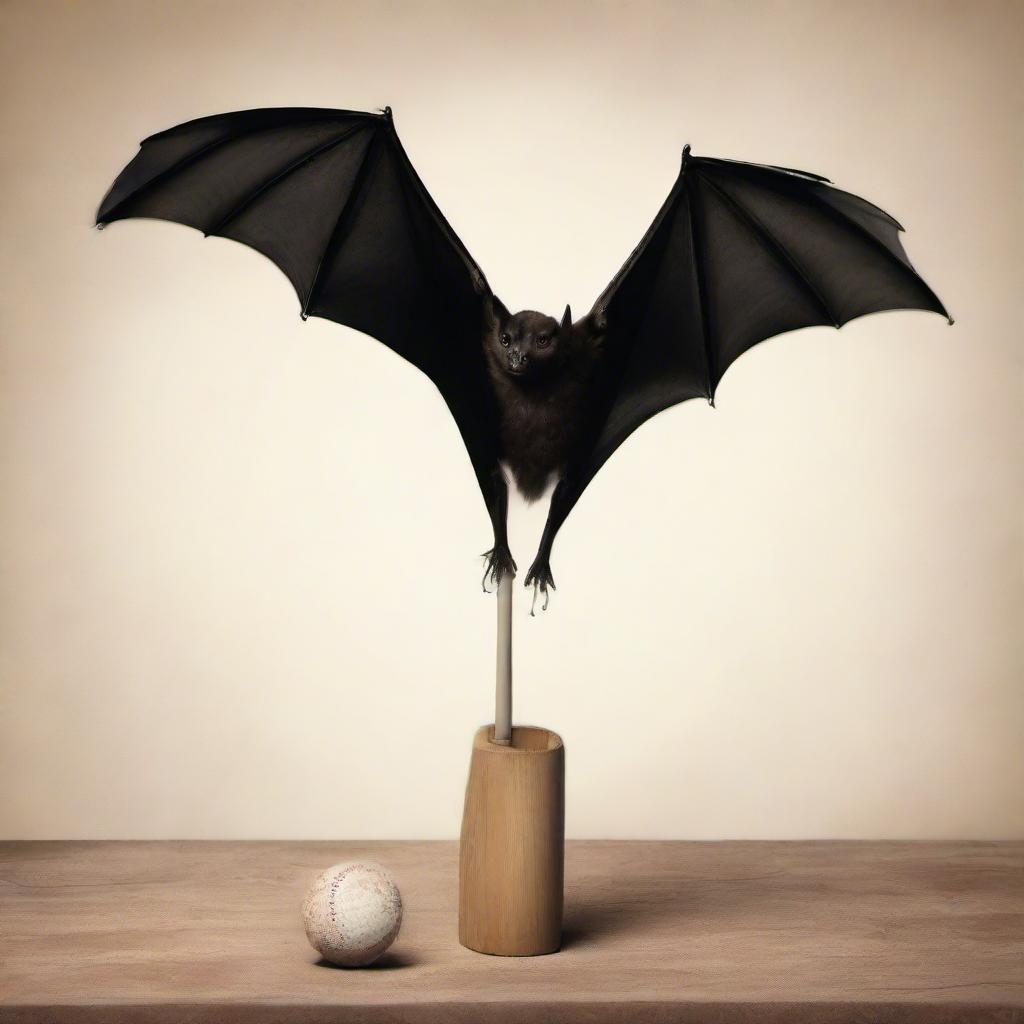}
    \includegraphics[width=0.16\linewidth]{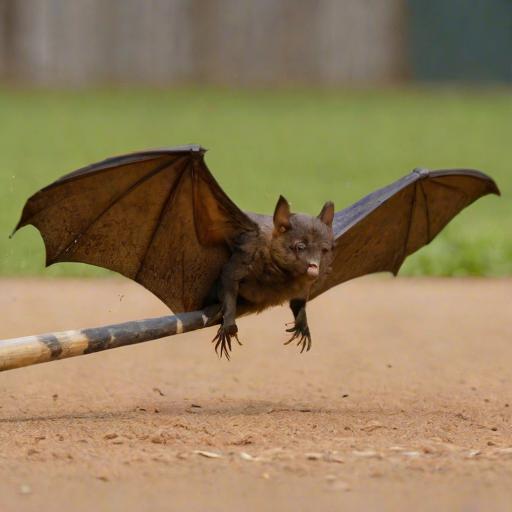}
    \includegraphics[width=0.16\linewidth]{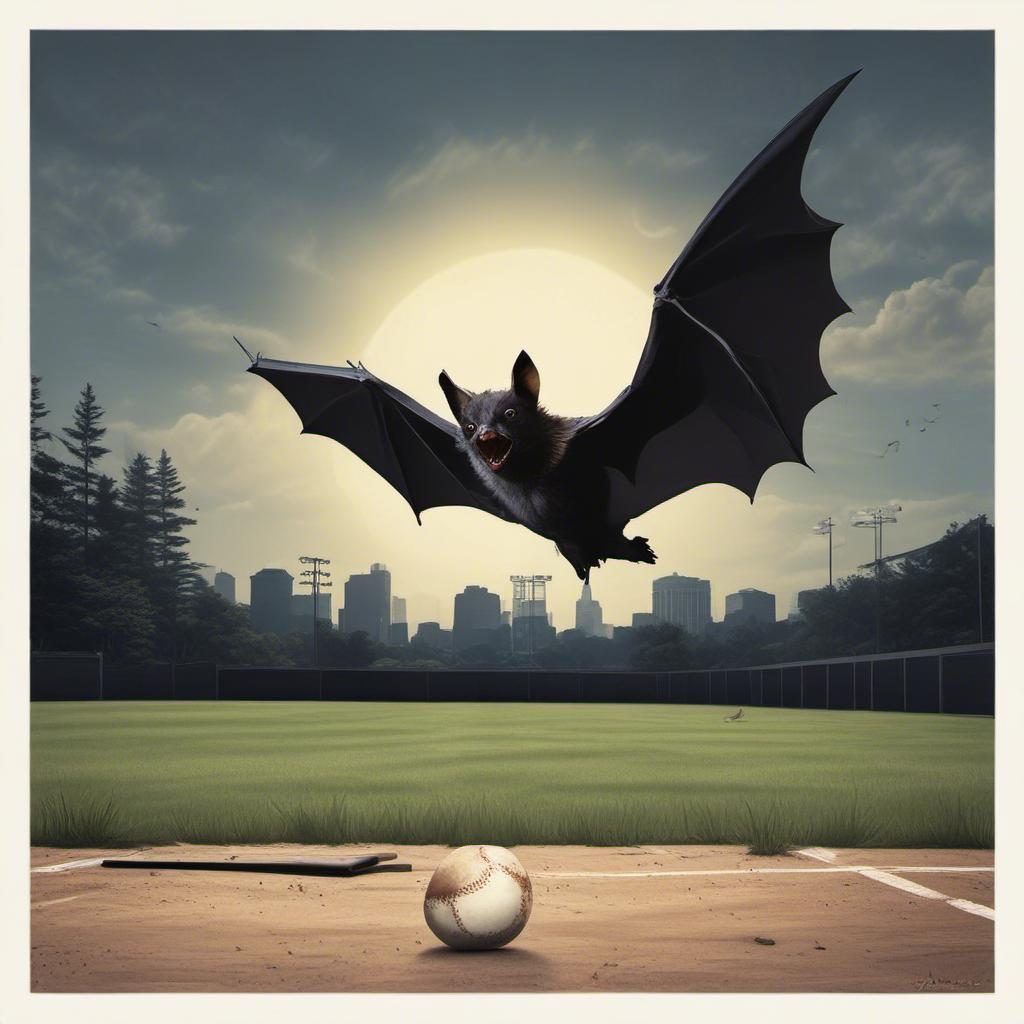}
    \includegraphics[width=0.16\linewidth]{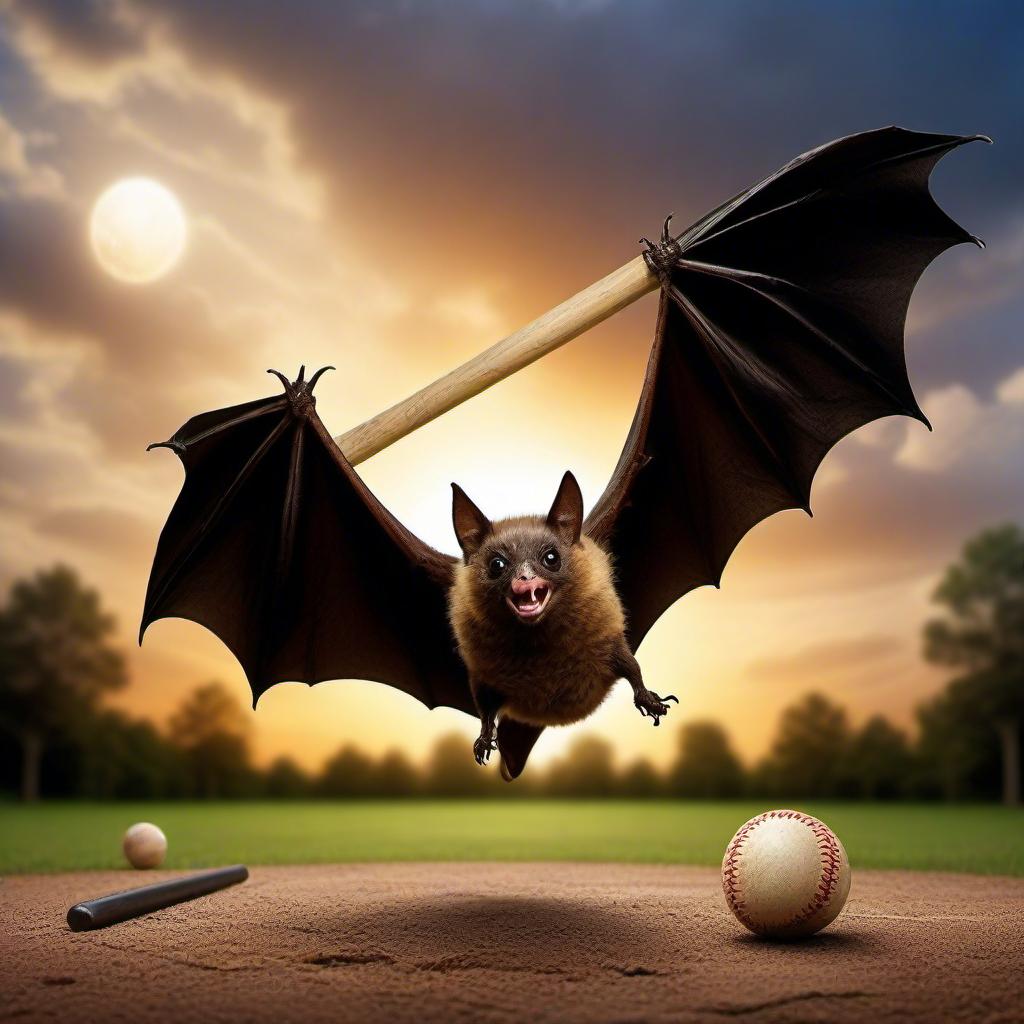}
    \includegraphics[width=0.16\linewidth]{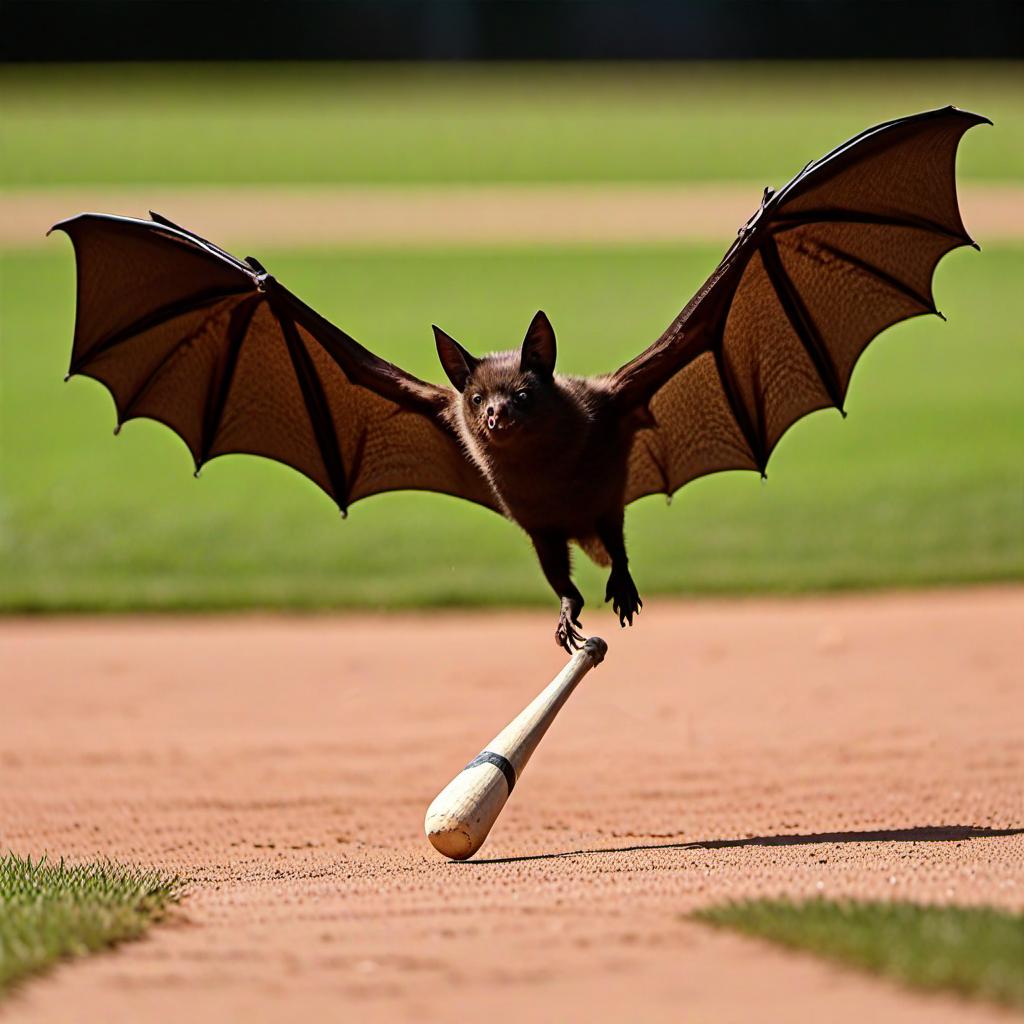}
    \parbox{\linewidth}{\centering A bat landing on a baseball bat.}

    \includegraphics[width=0.16\linewidth]{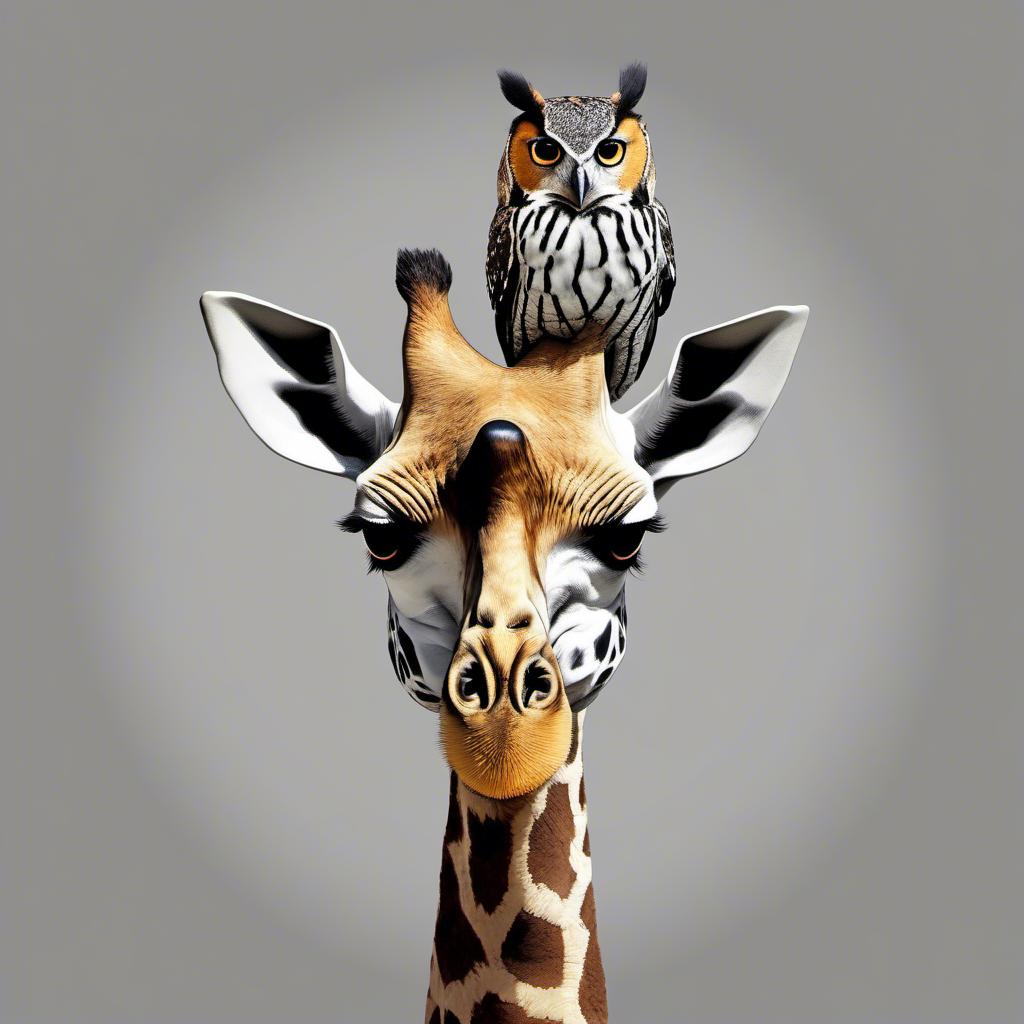}
    \includegraphics[width=0.16\linewidth]{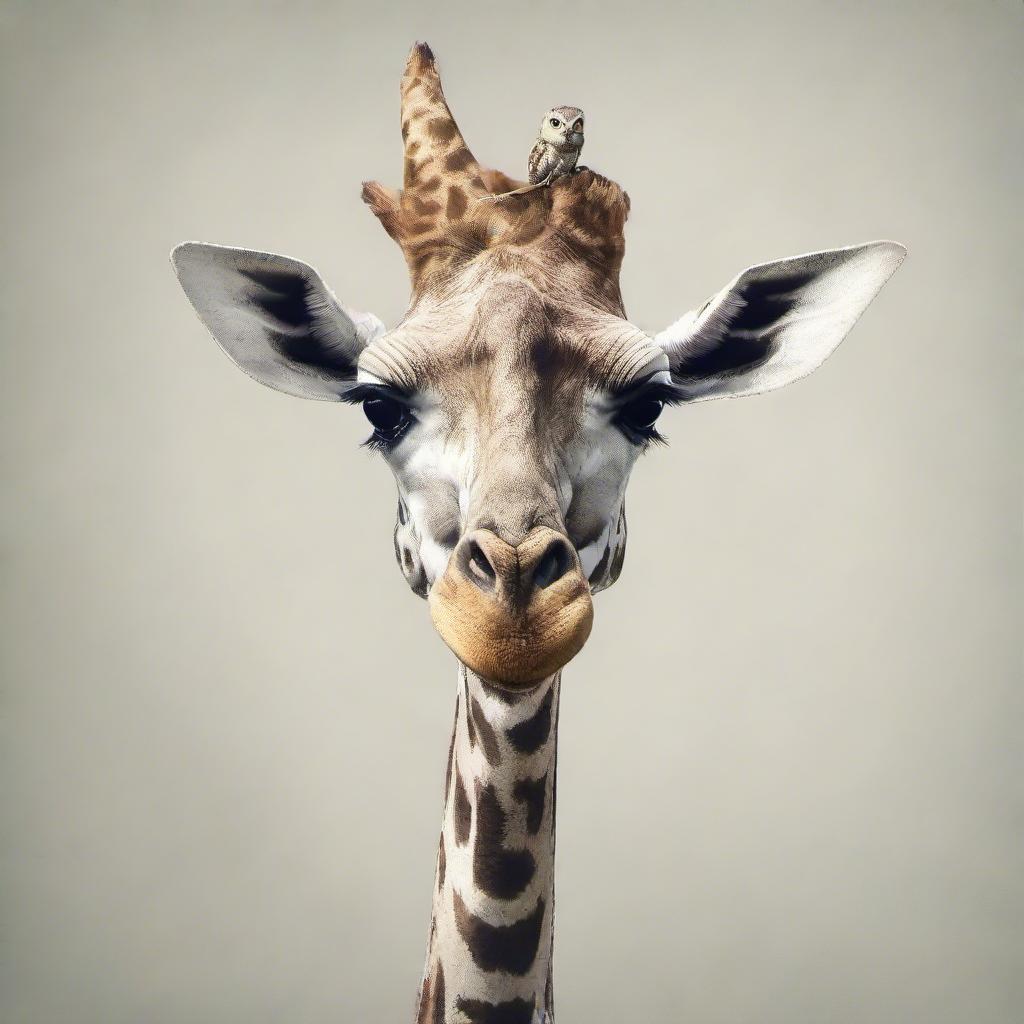}
    \includegraphics[width=0.16\linewidth]{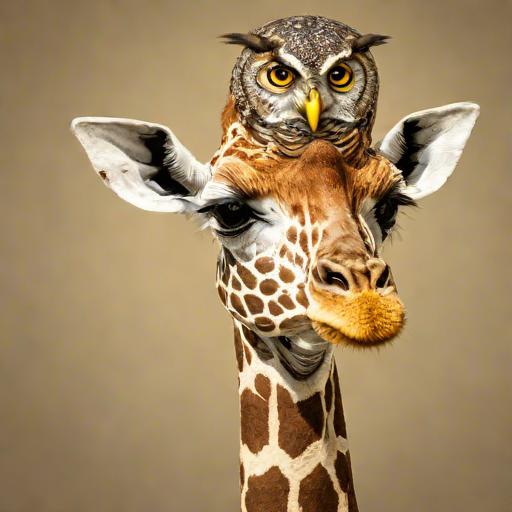}
    \includegraphics[width=0.16\linewidth]{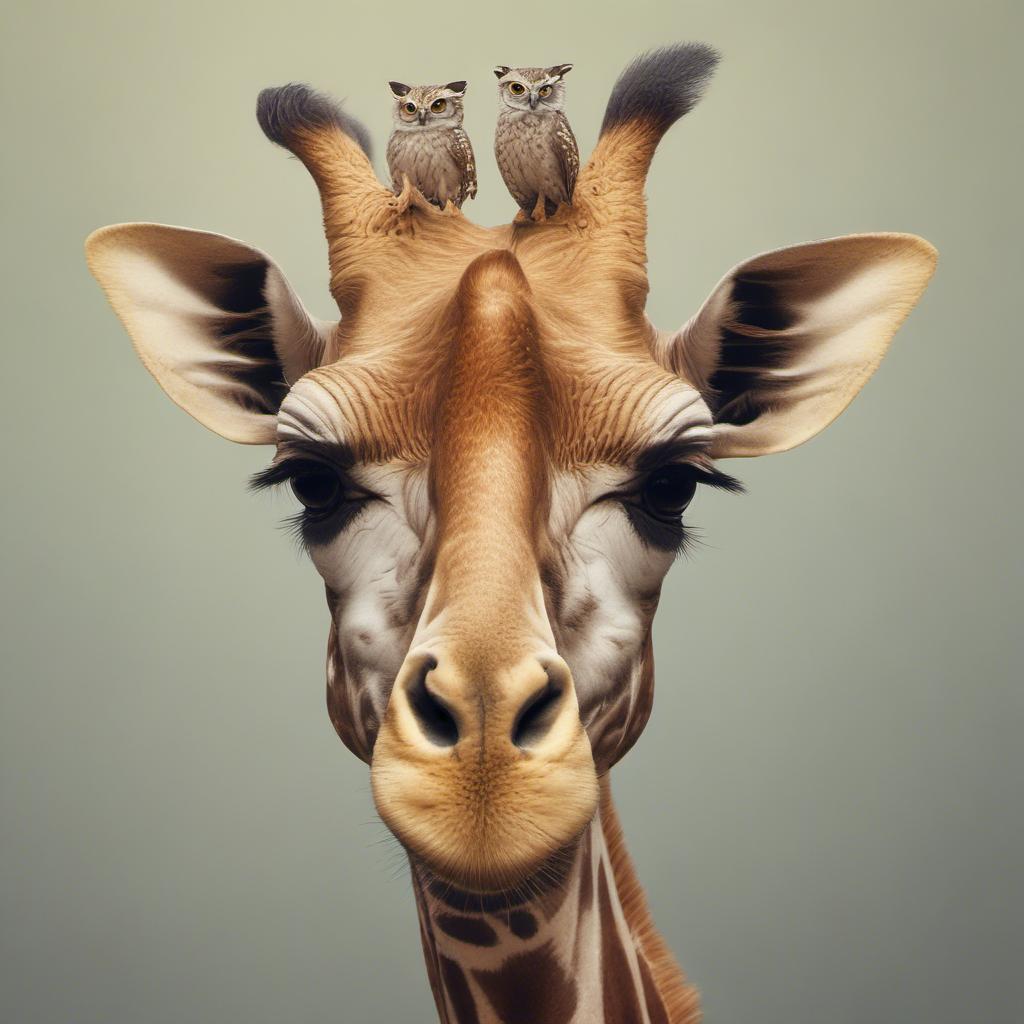}
    \includegraphics[width=0.16\linewidth]{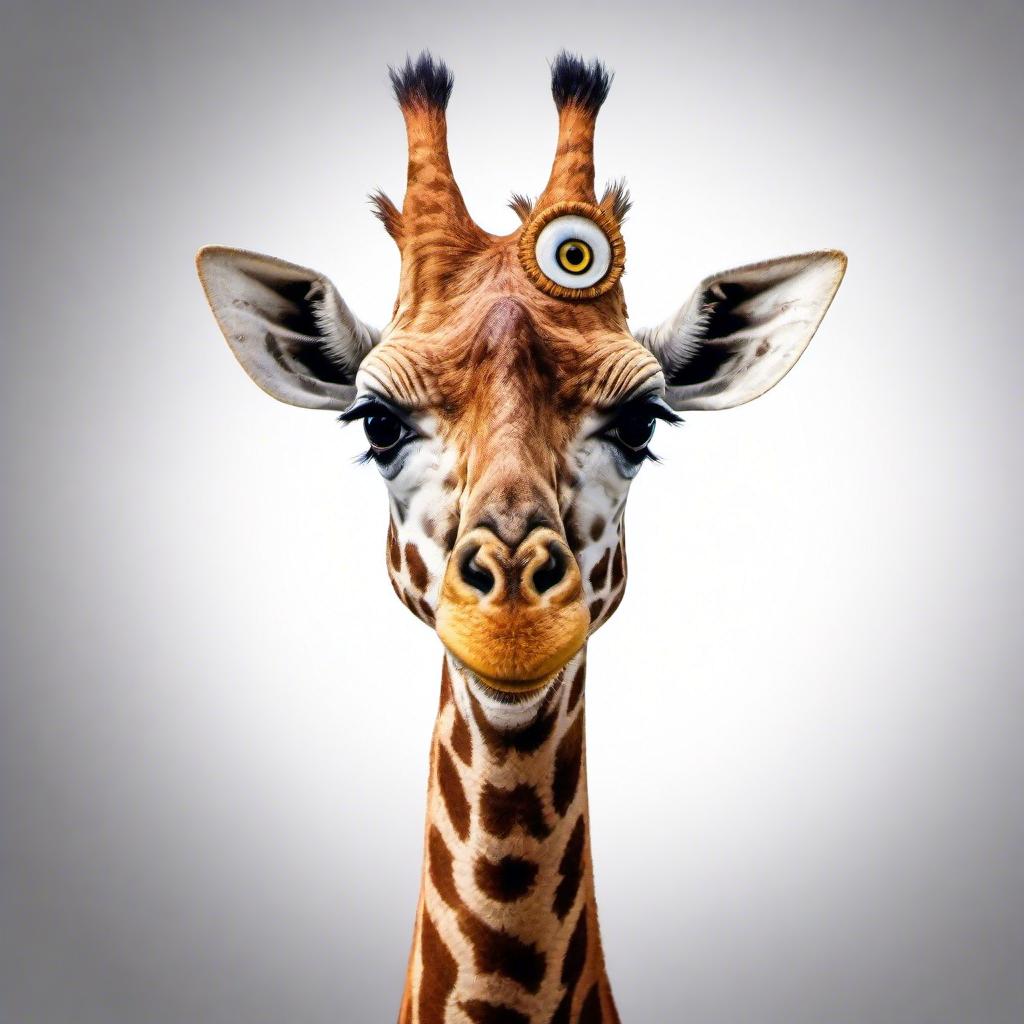}
    \includegraphics[width=0.16\linewidth]{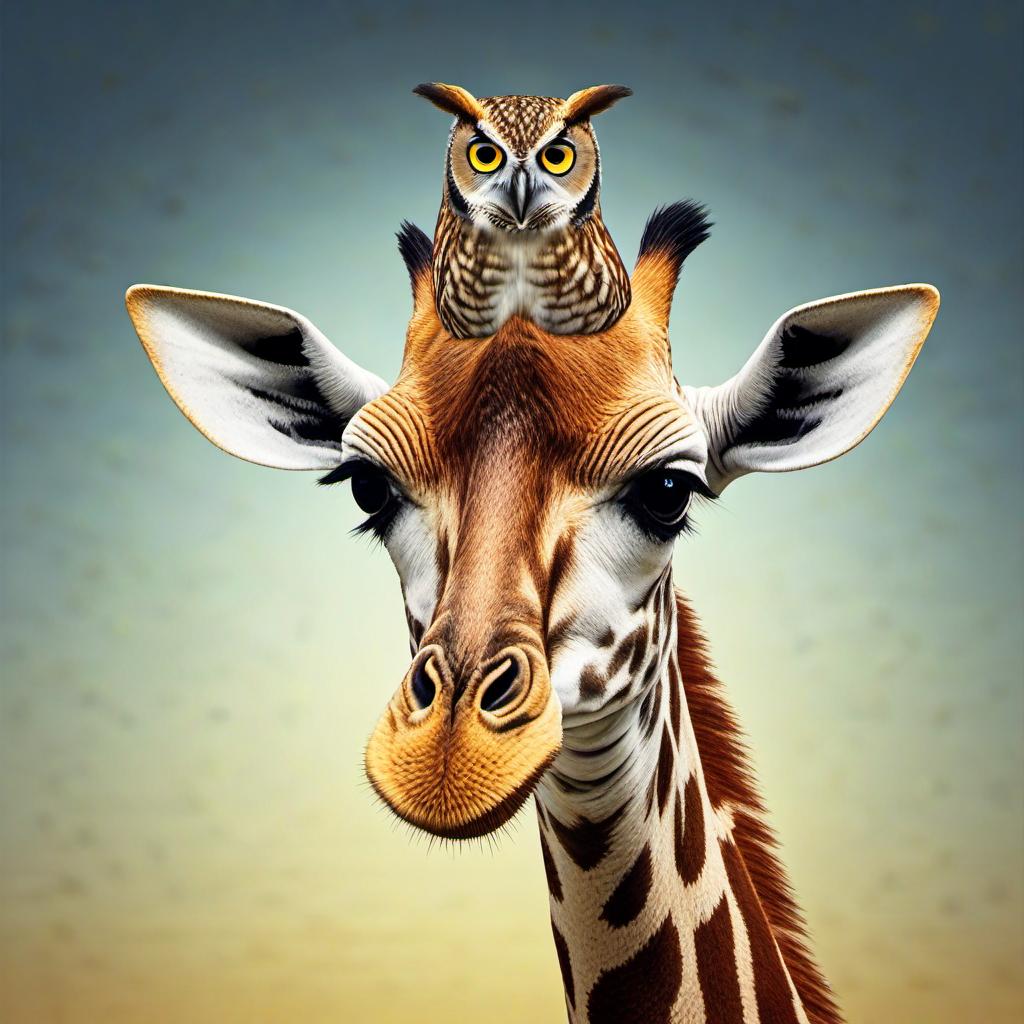}
    \parbox{\linewidth}{\centering A giraffe with an owl on its head.}

    \includegraphics[width=0.16\linewidth]{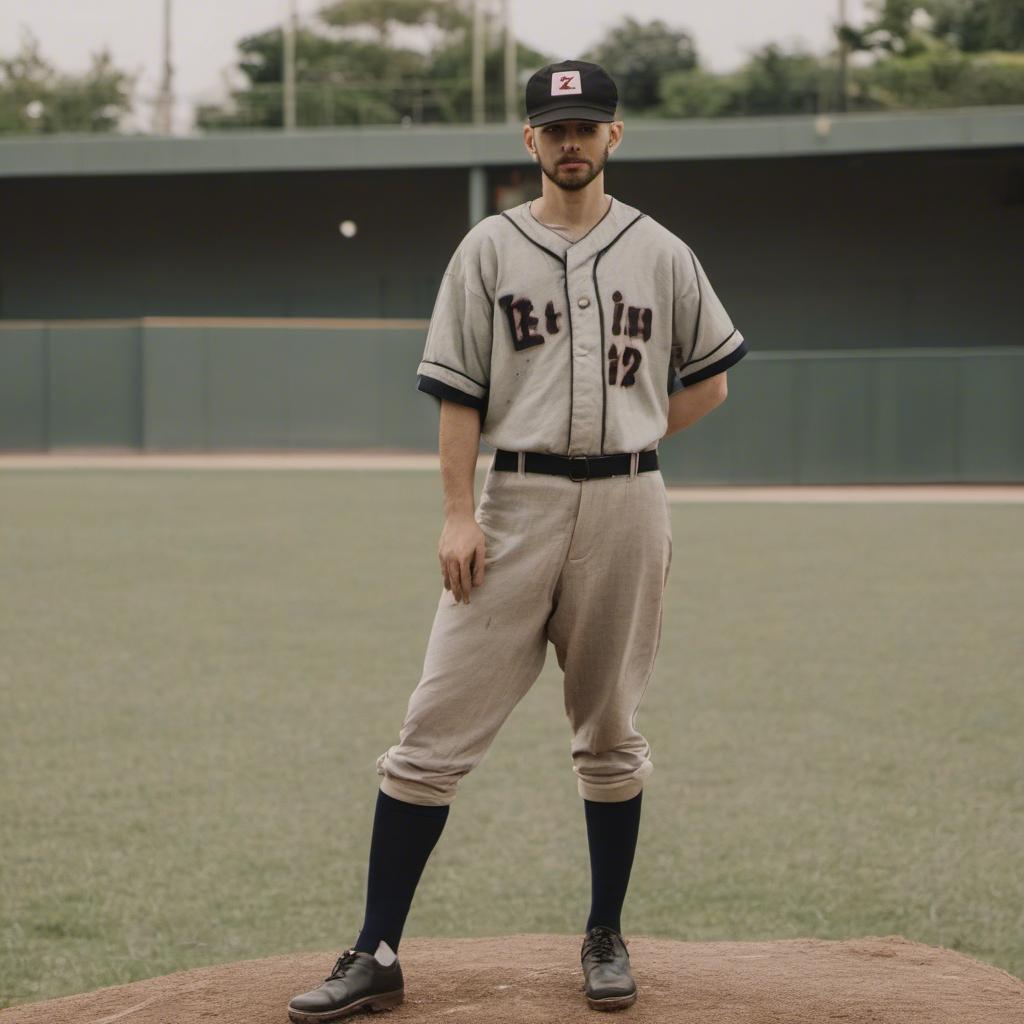}
    \includegraphics[width=0.16\linewidth]{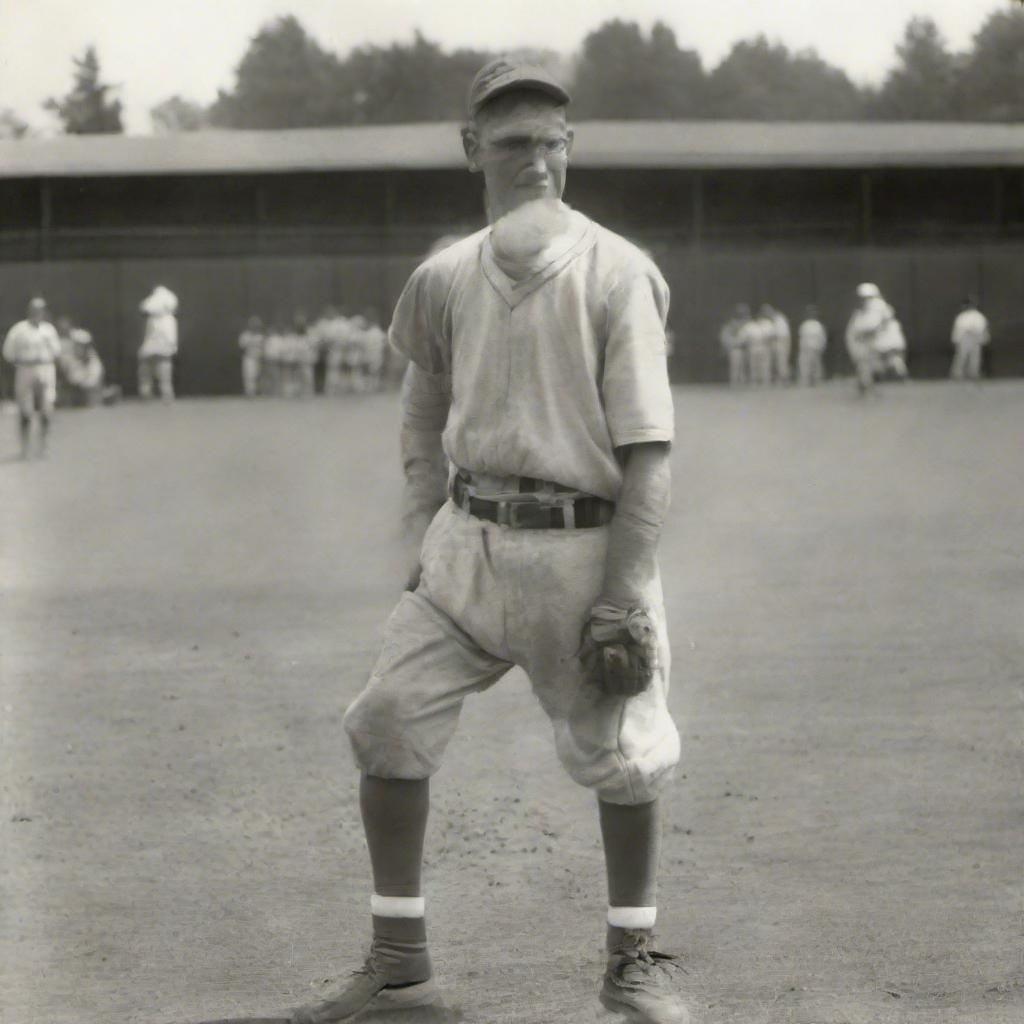}
    \includegraphics[width=0.16\linewidth]{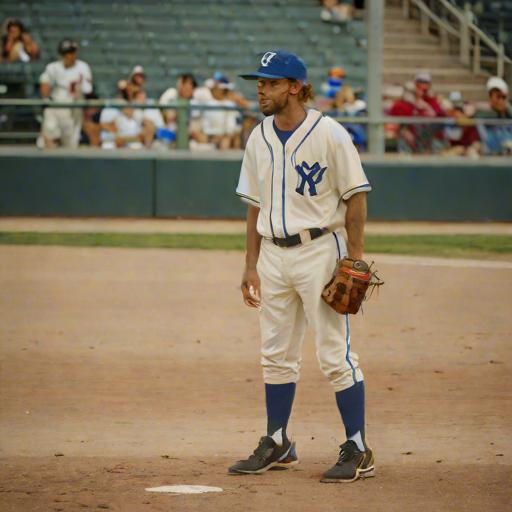}
    \includegraphics[width=0.16\linewidth]{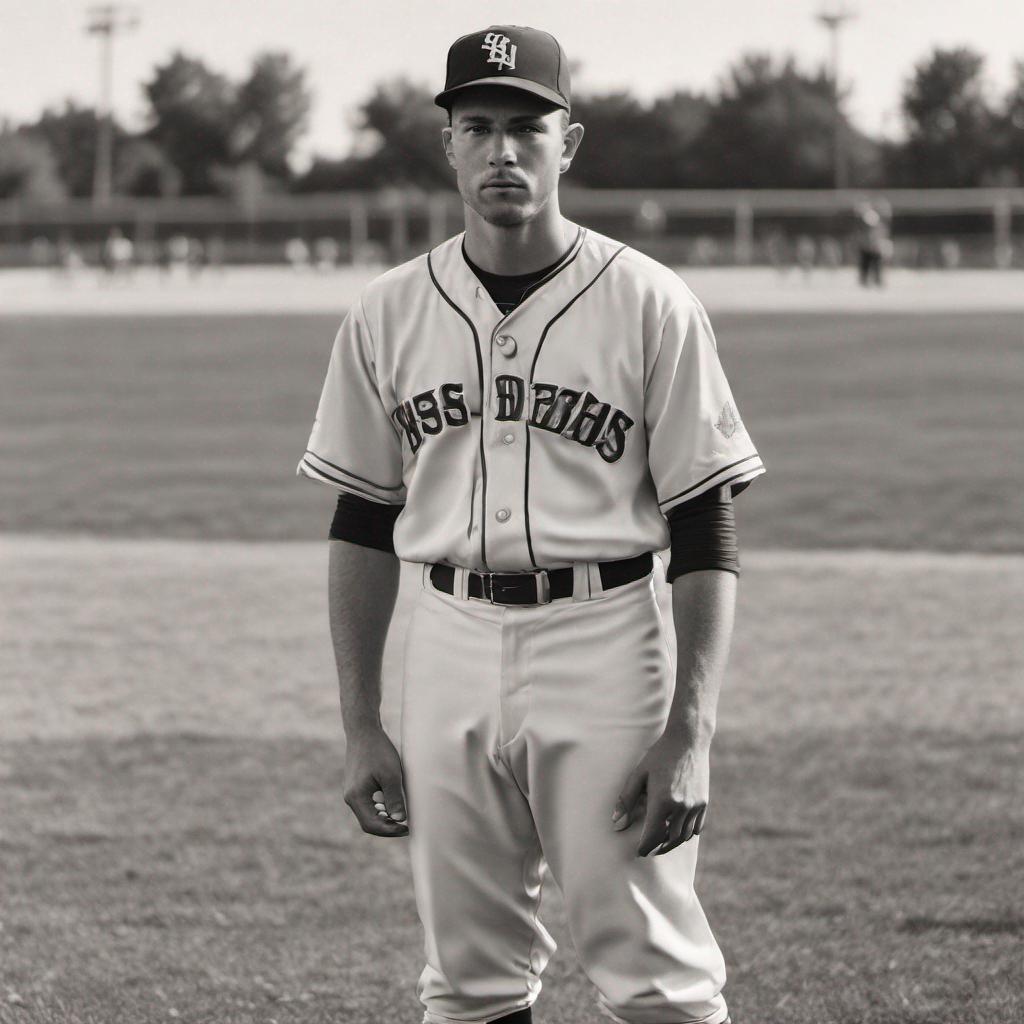}
    \includegraphics[width=0.16\linewidth]{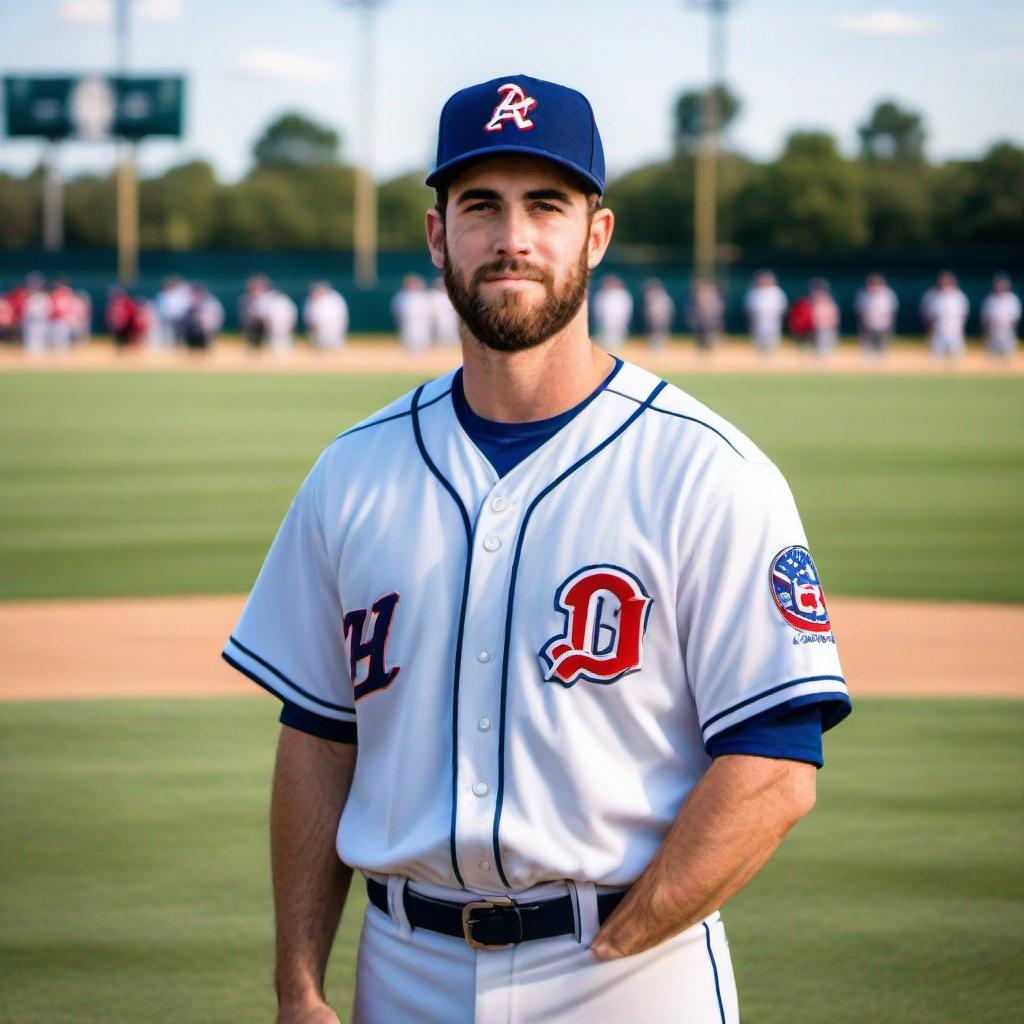}
    \includegraphics[width=0.16\linewidth]{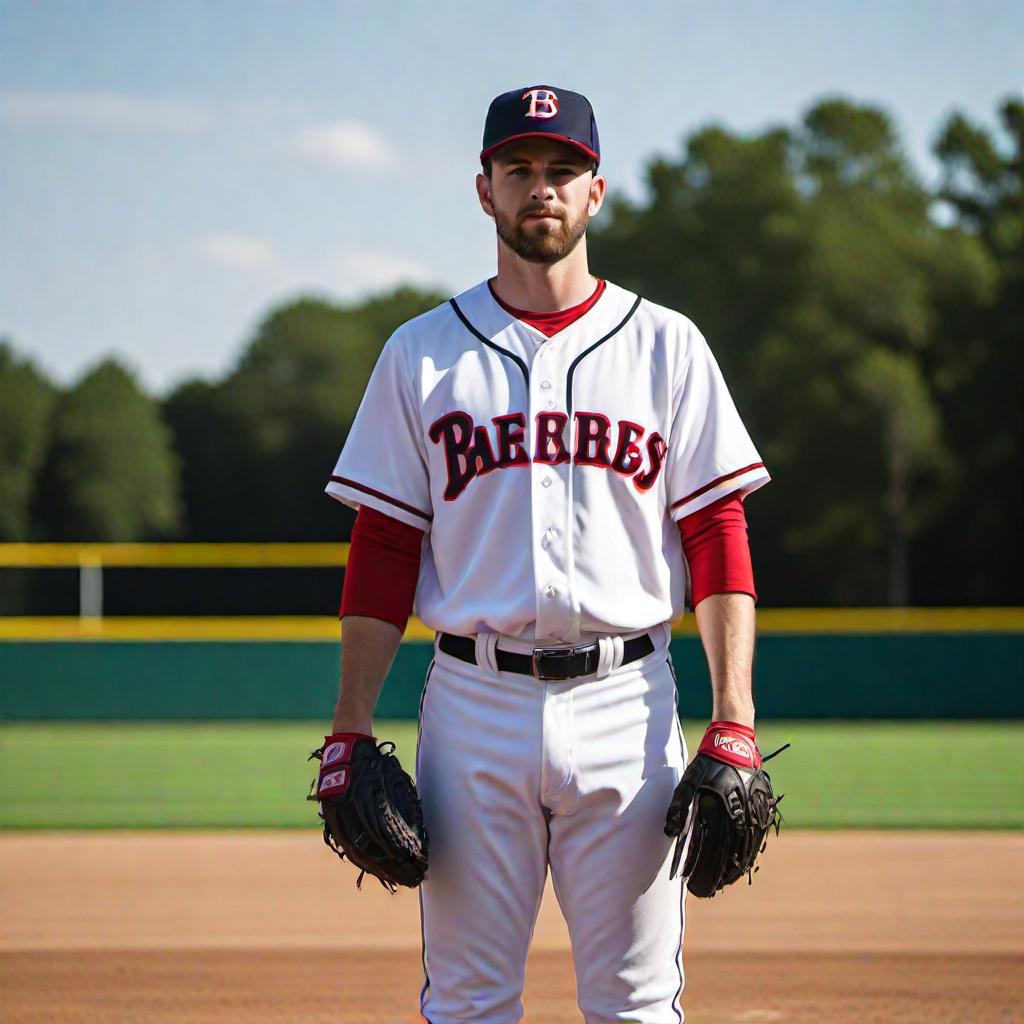}
    \parbox{\linewidth}{\centering A baseball player standing on the field in uniform.}

    \includegraphics[width=0.16\linewidth]{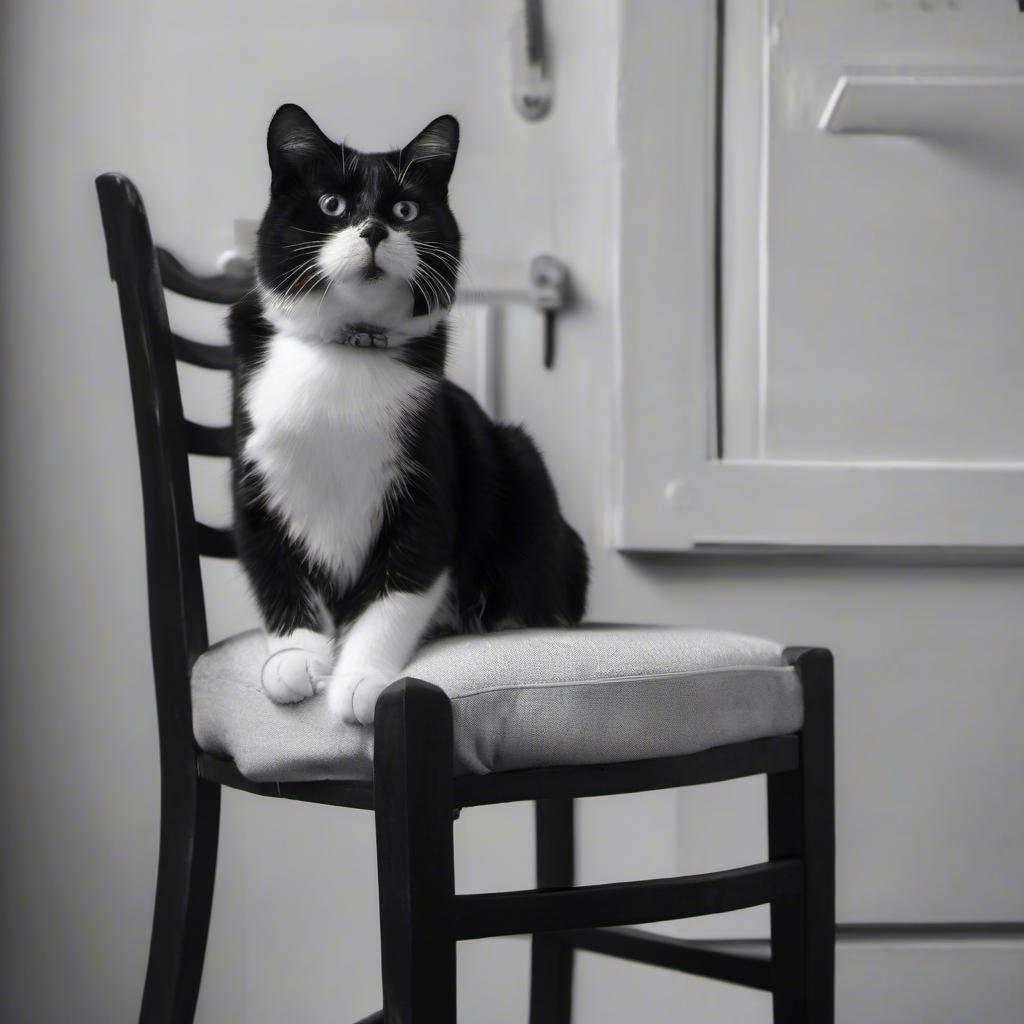}
    \includegraphics[width=0.16\linewidth]{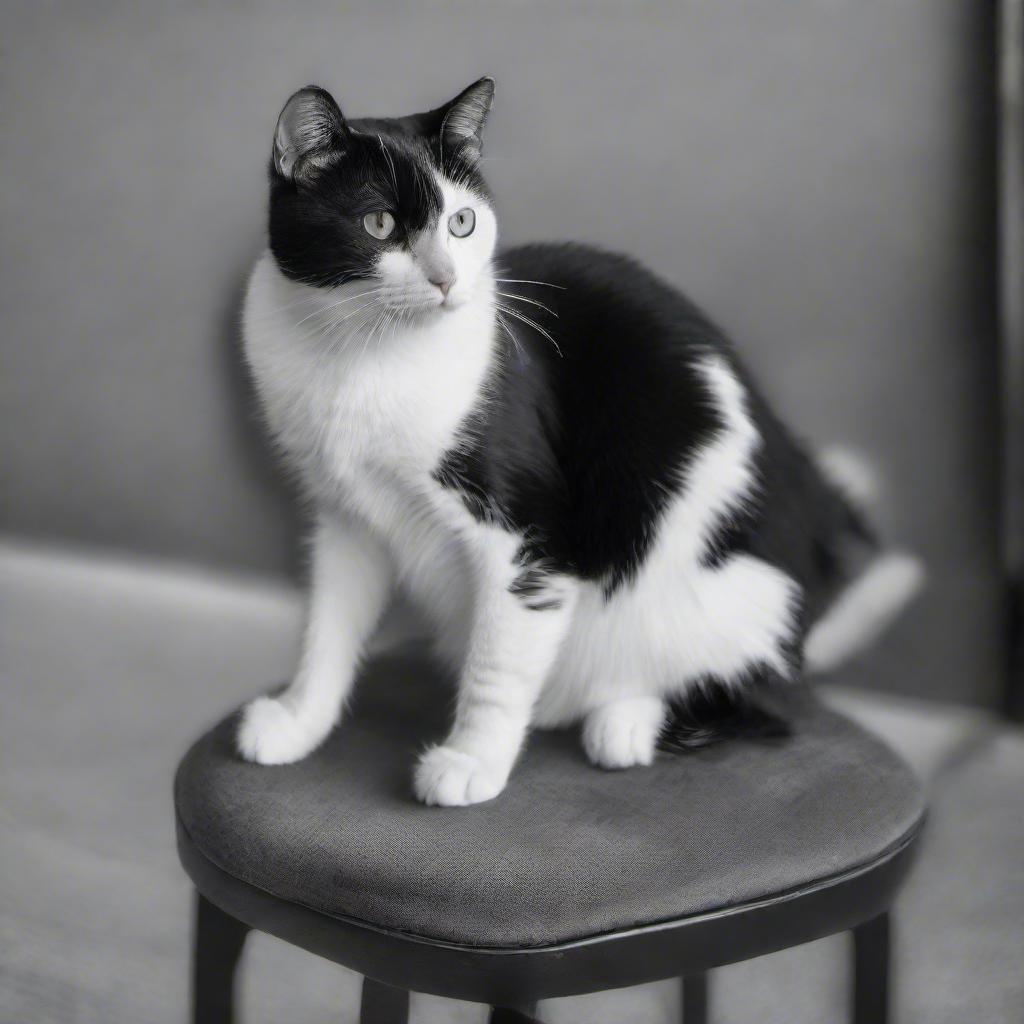}
    \includegraphics[width=0.16\linewidth]{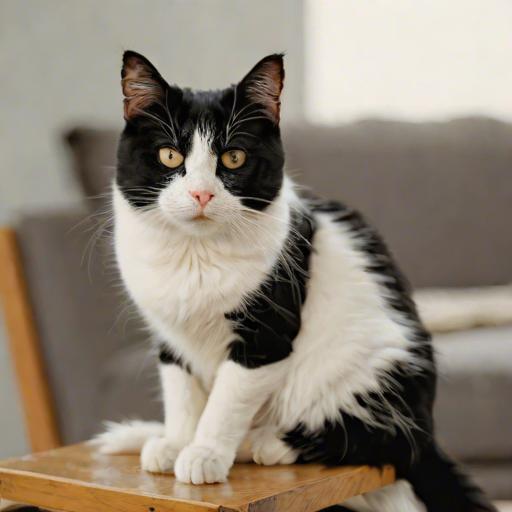}
    \includegraphics[width=0.16\linewidth]{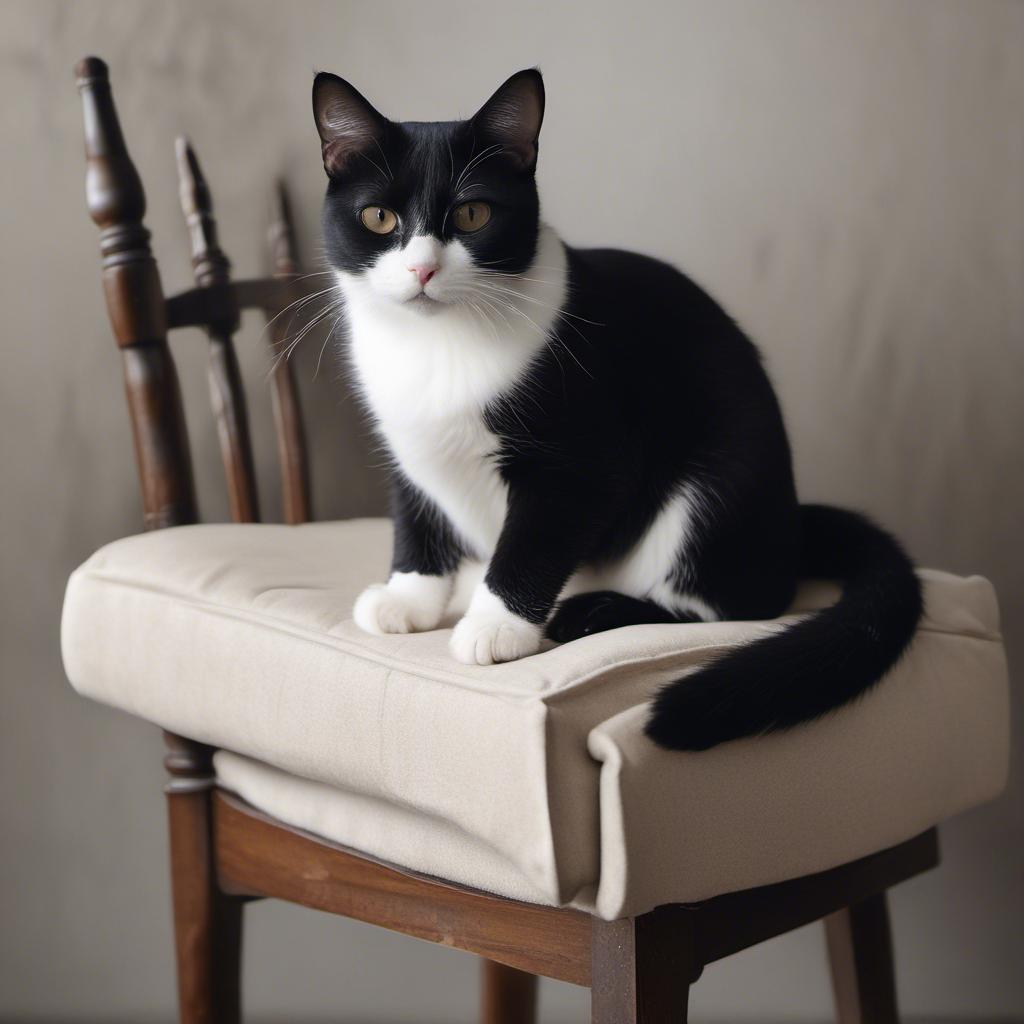}
    \includegraphics[width=0.16\linewidth]{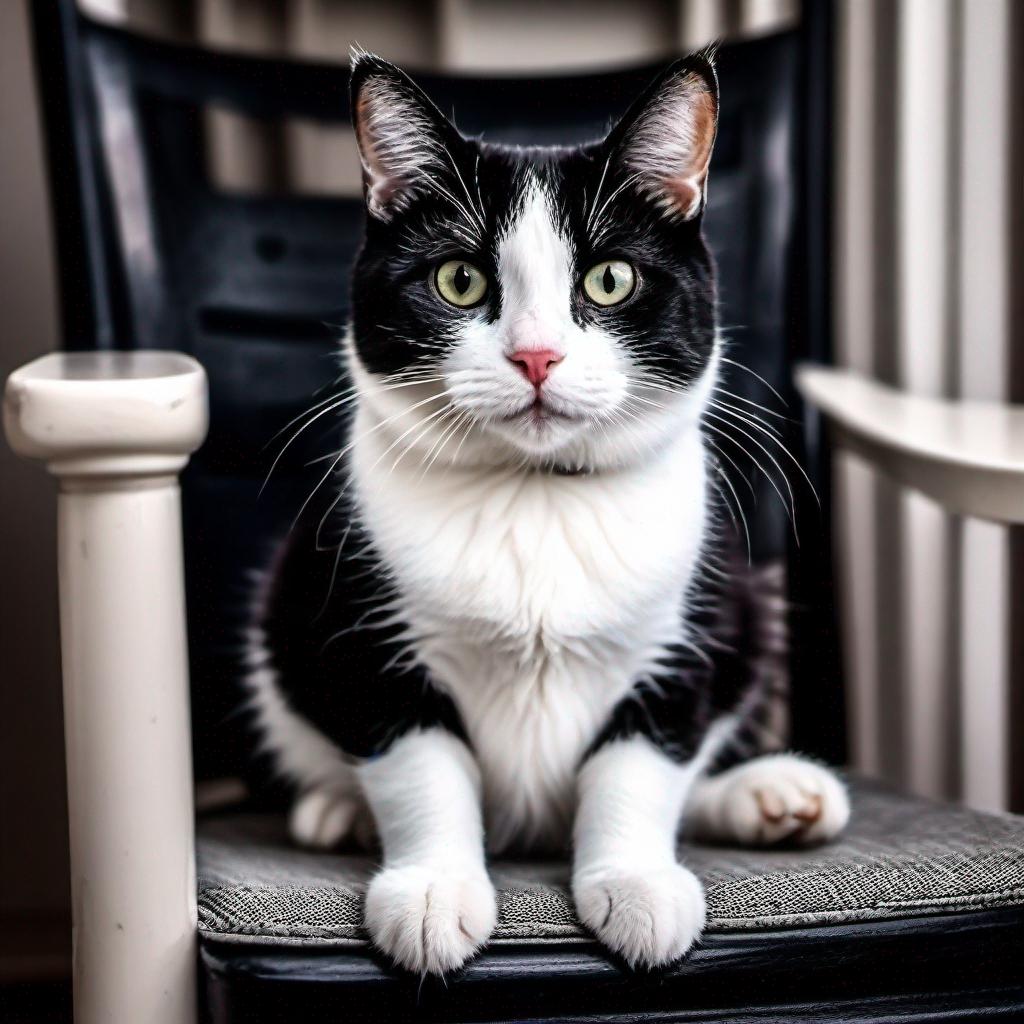}
    \includegraphics[width=0.16\linewidth]{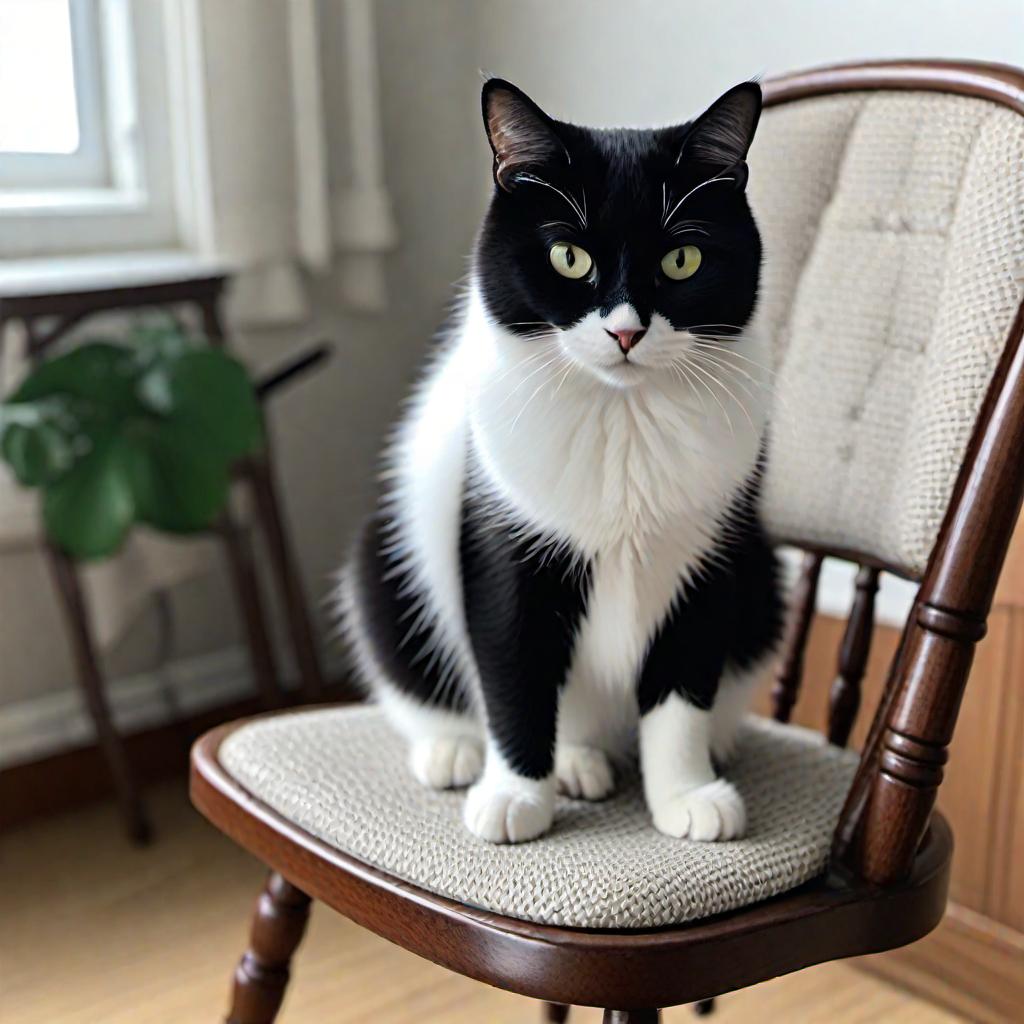}
    \parbox{\linewidth}{\centering A black and white cat sitting on top of a chair.}

    \includegraphics[width=0.16\linewidth]{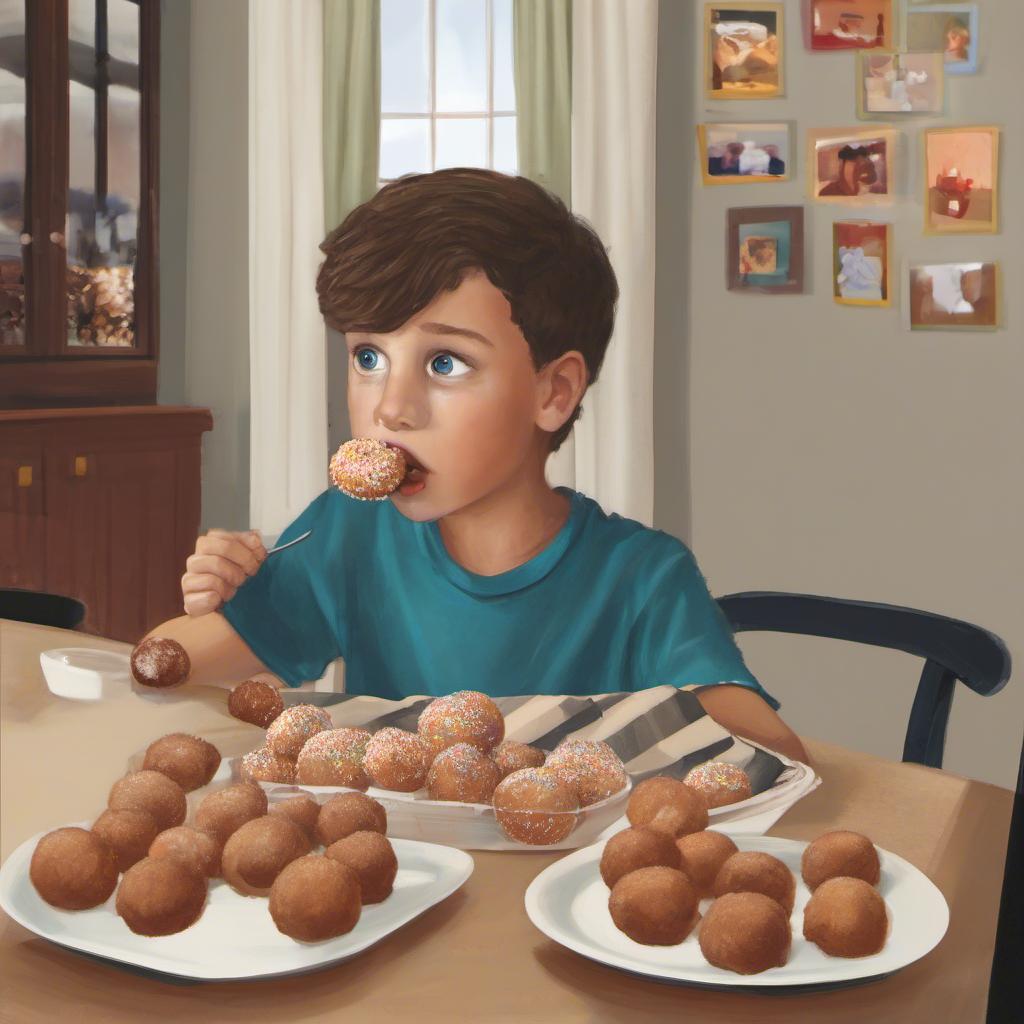}
    \includegraphics[width=0.16\linewidth]{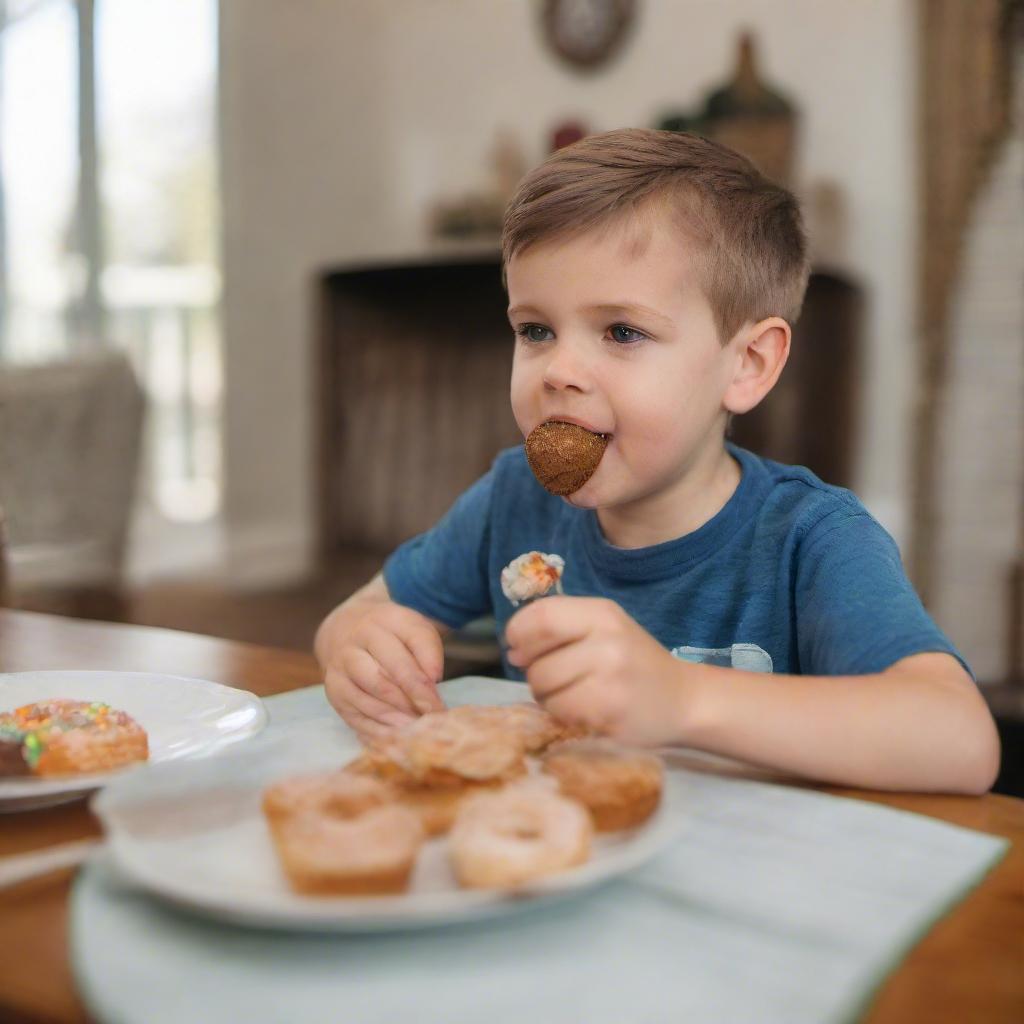}
    \includegraphics[width=0.16\linewidth]{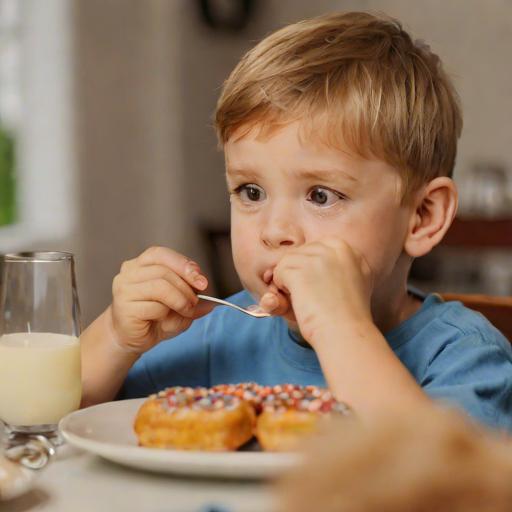}
    \includegraphics[width=0.16\linewidth]{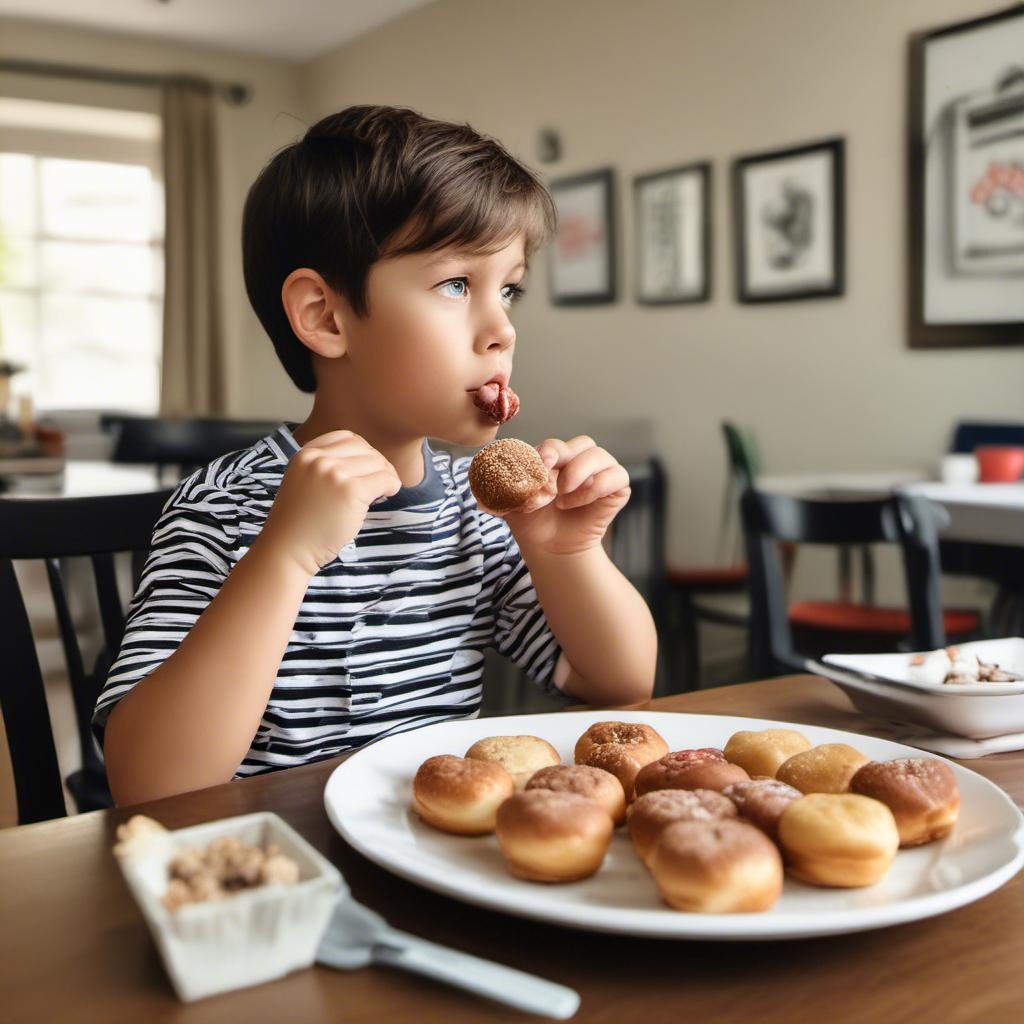}
    \includegraphics[width=0.16\linewidth]{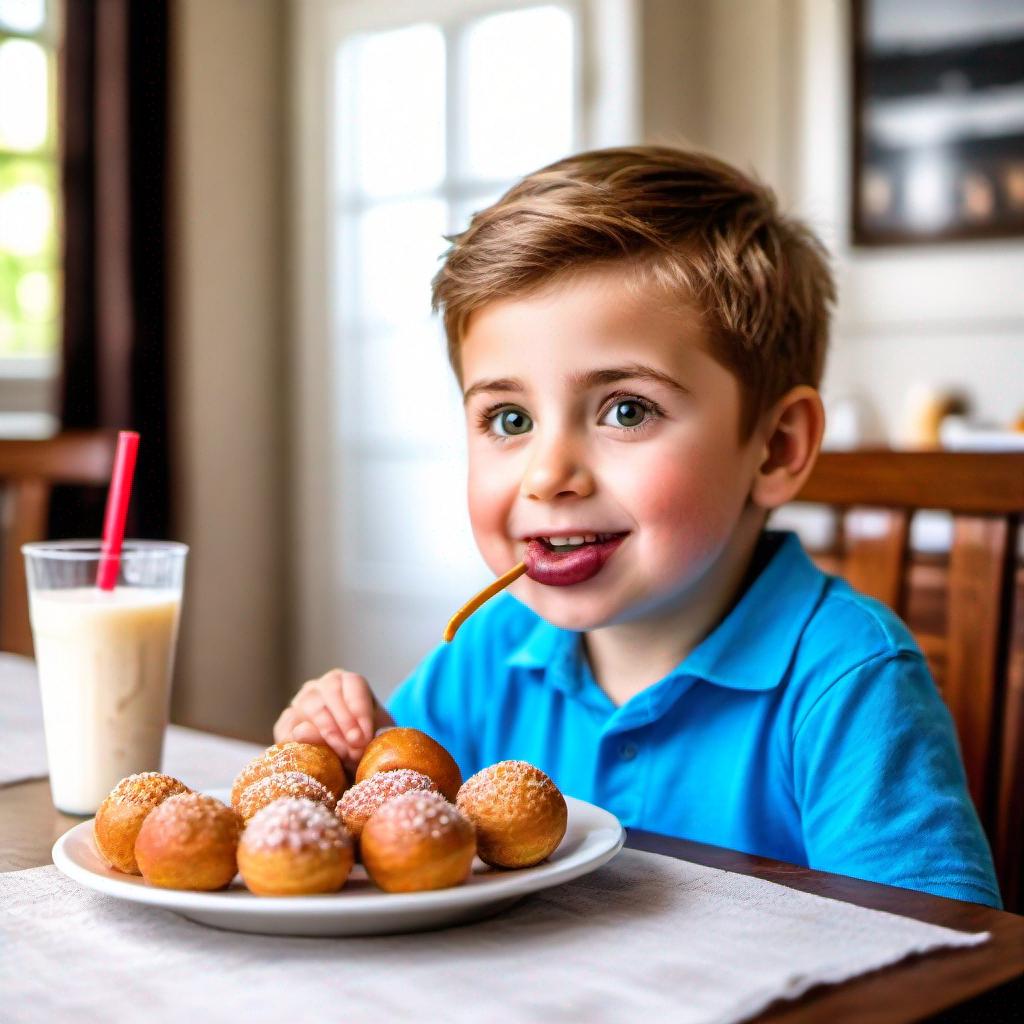}
    \includegraphics[width=0.16\linewidth]{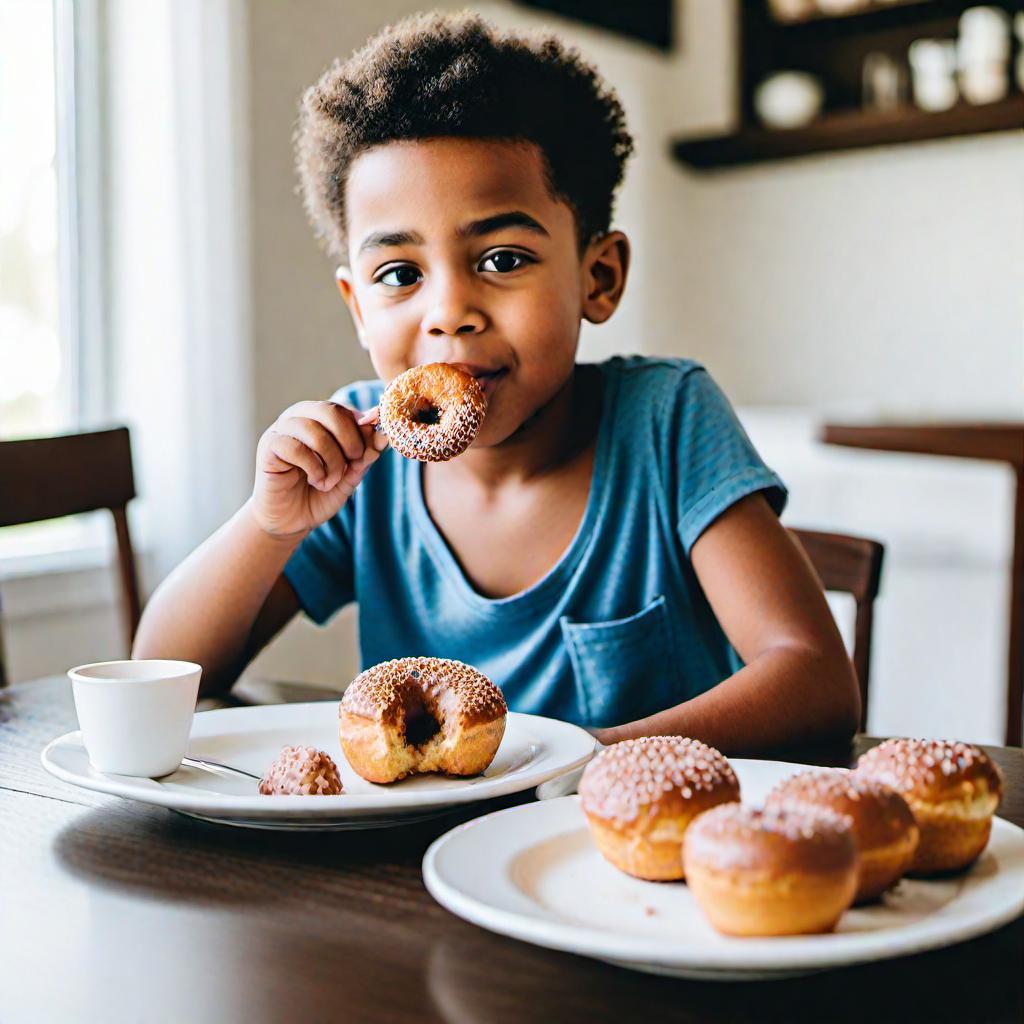}
    \parbox{\linewidth}{\centering A boy is eating donut holes while sitting at a dinner table. }

    \includegraphics[width=0.16\linewidth]{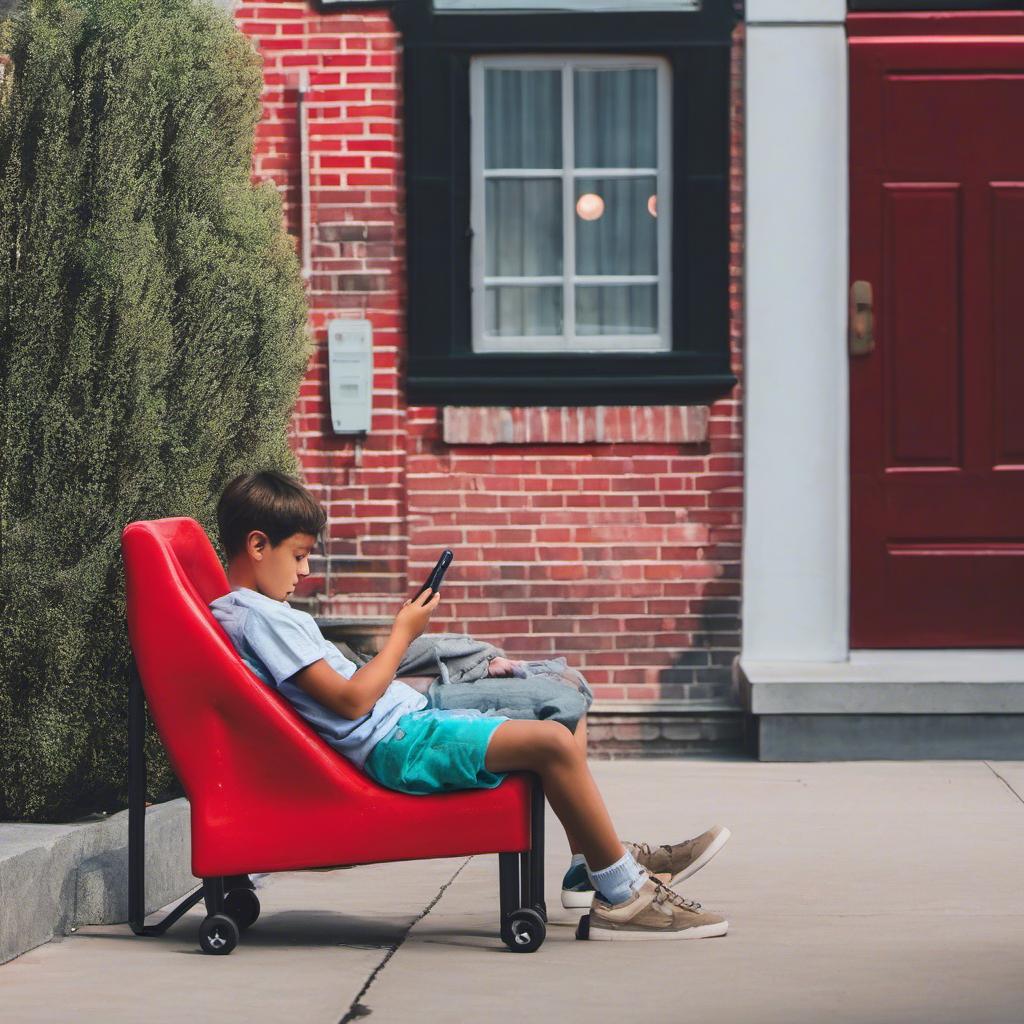}
    \includegraphics[width=0.16\linewidth]{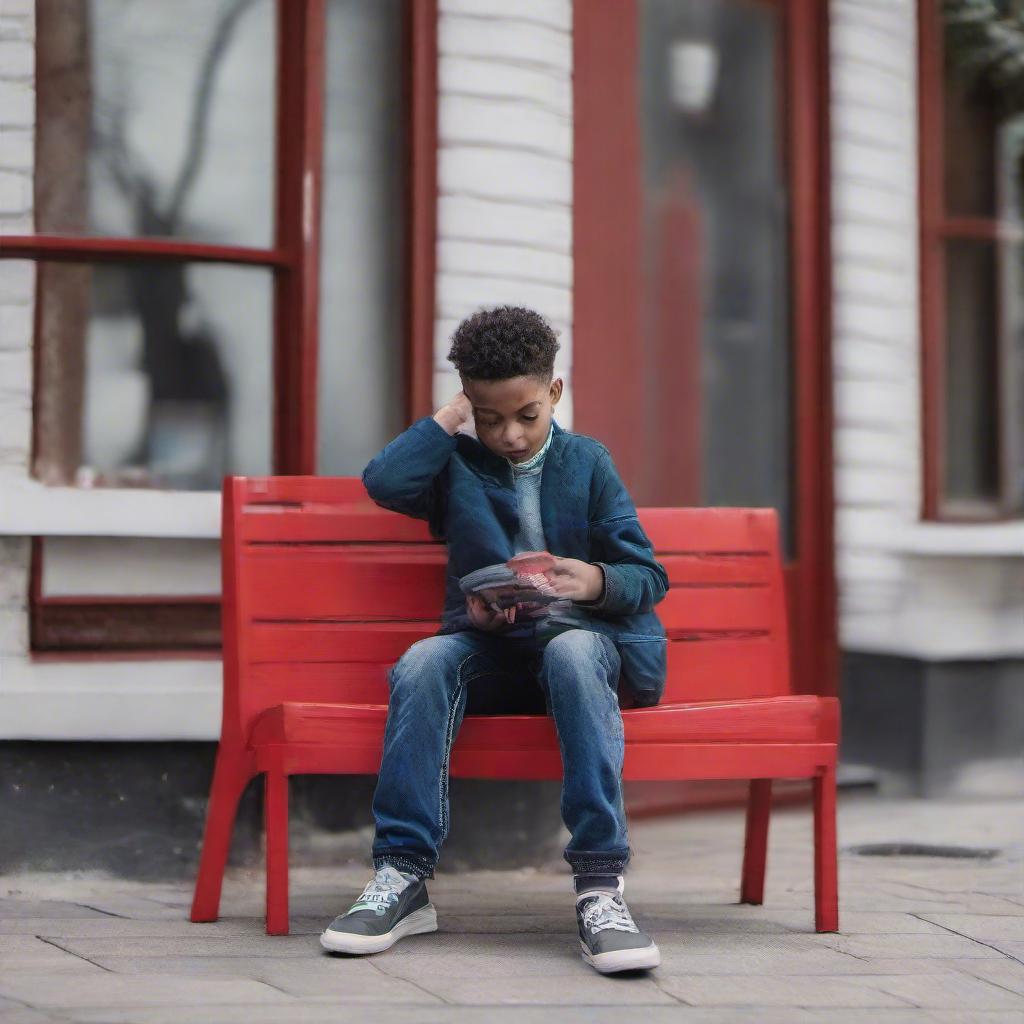}
    \includegraphics[width=0.16\linewidth]{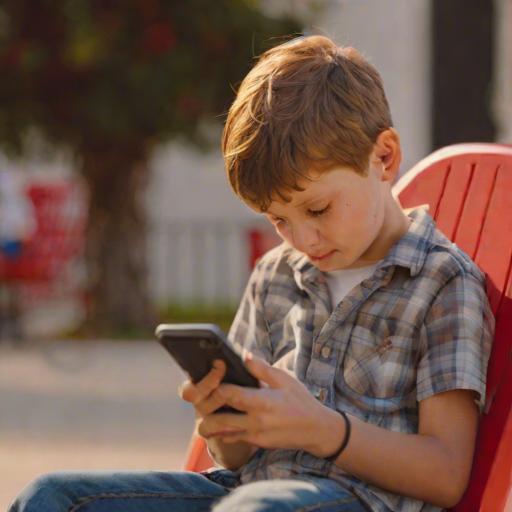}
    \includegraphics[width=0.16\linewidth]{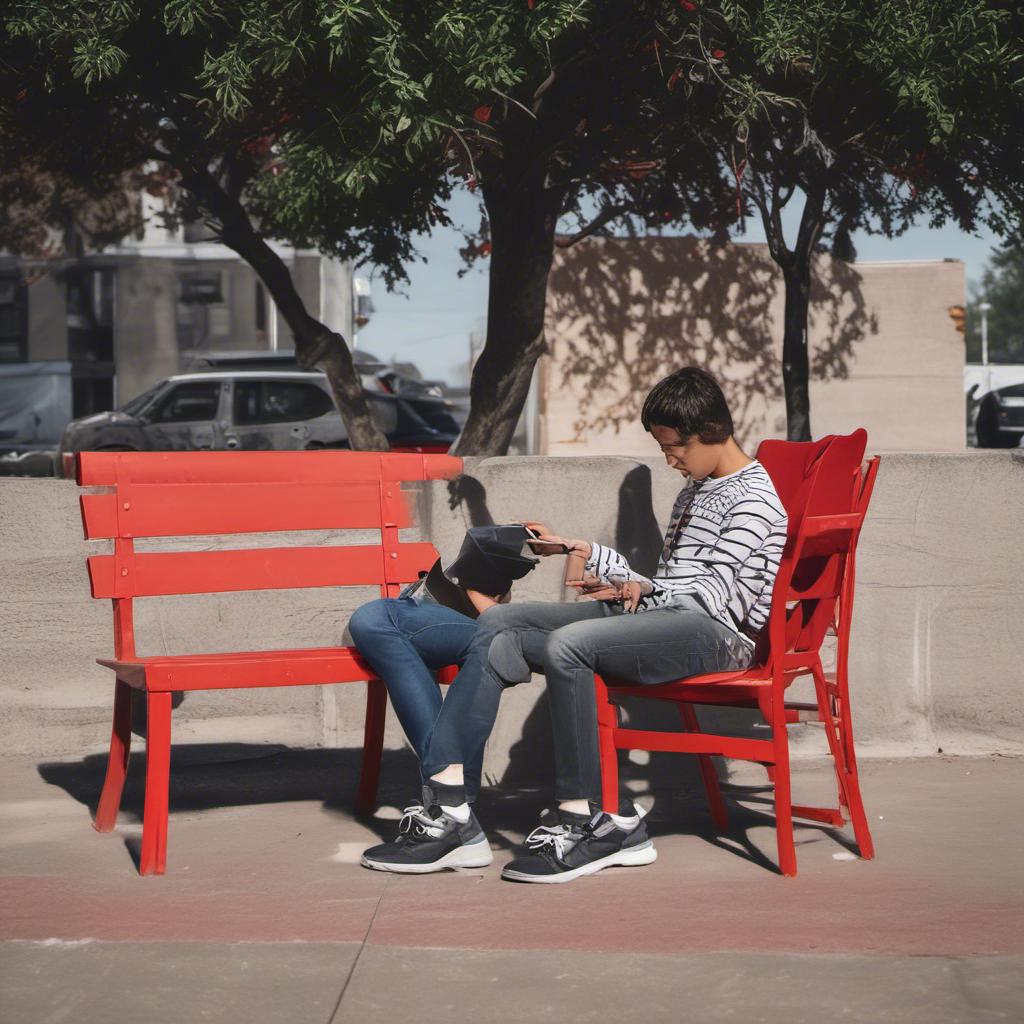}
    \includegraphics[width=0.16\linewidth]{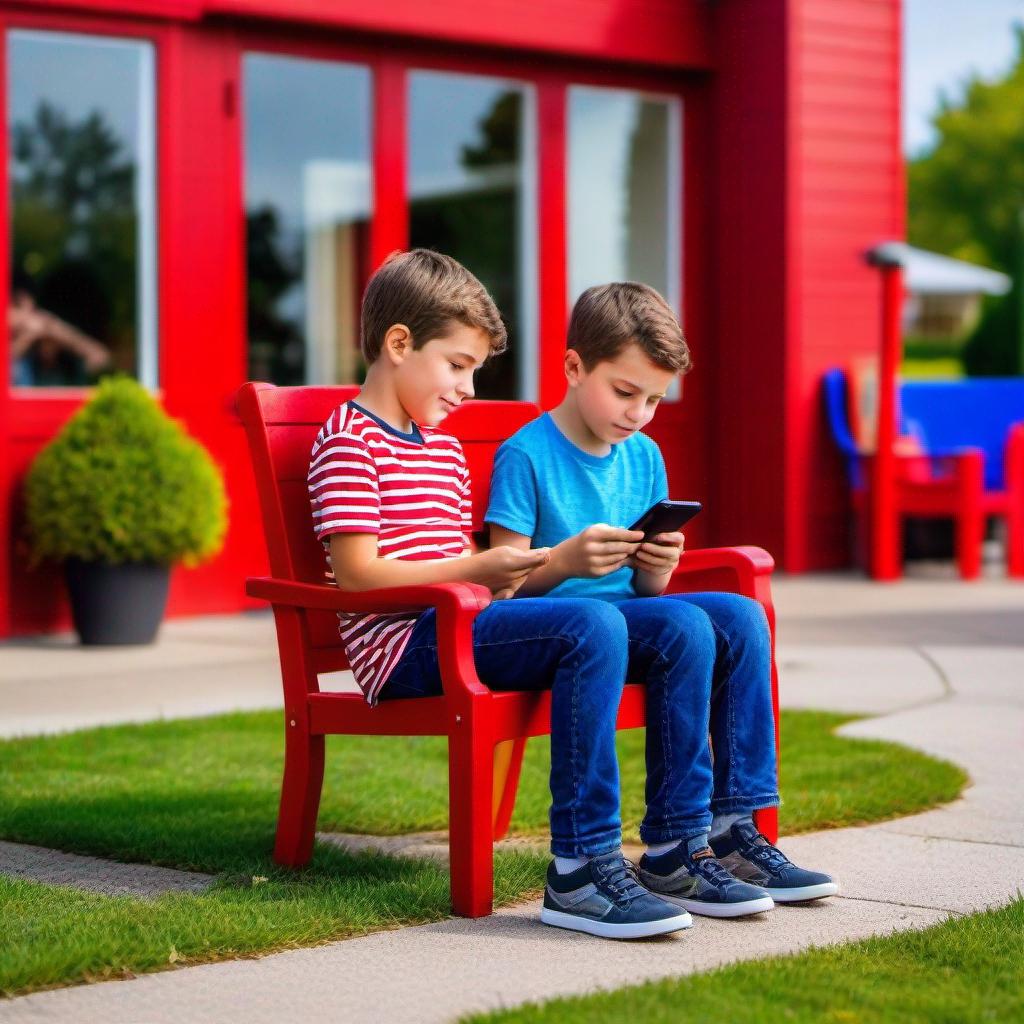}
    \includegraphics[width=0.16\linewidth]{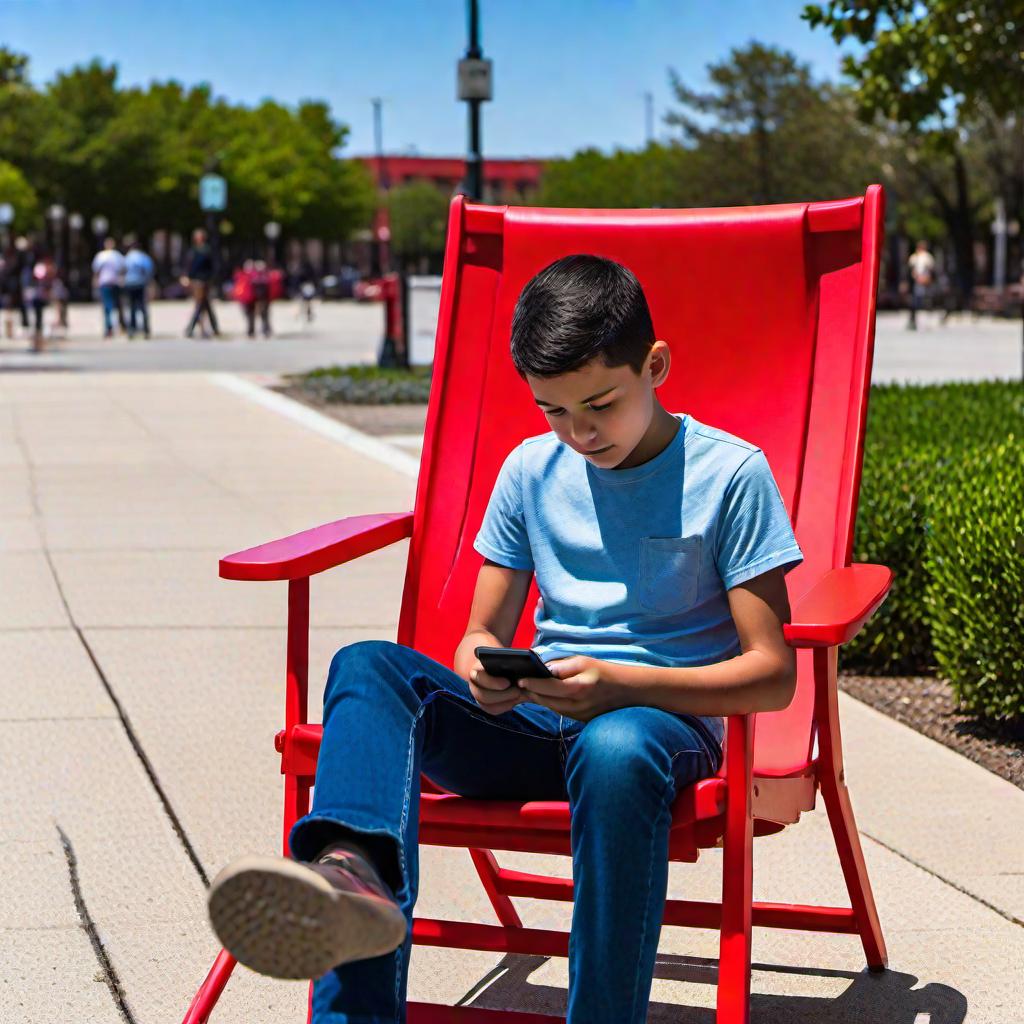}
    \parbox{\linewidth}{\centering A boy on his phone outside near a red chair.}

    \caption{Visual comparison between our TLCM and the state-of-the-art methods. Zoom in for more details.}
    \label{fig:comparison}
\end{figure}

We present the visual comparisons in Figure~\ref{fig:comparison}.
Under the same conditions, we observe that the images generated by TLCM have better image quality and maintain higher semantic consistency on more challenging prompts, which also leads to greater human preference.

\subsection{Ablation Study}
To analyze the key components of our method, we make a
thorough ablation study to verify the effectiveness of the
proposed TLCM. 
Table~\ref{ablation} depicts the performance of TLCM's variants. 

\textbf{Data-free multistep latent consistency distillation.} 
As shown in Table~\ref{table:ablation}, only using $\mathcal{L}_{lcd-s}$ which computes $z_{t_m}$ by single step for LCD  achieves CS score of 31.61, AS of 5.89,  indicating our data-free method is able to accelerate LDM with good quality. 
 Changing  $\mathcal{L}_{lcd-s}$ to  single-step denoising MLCD $\mathcal{L}_{mlcd-s}$,   all metrics are improved.  This result verifies that  MLCD has a stronger capability to accelerate LDM than LCD. 
 This is because it is hard for data-free LCD to enforce consistency across the entire timestep range while data-free MLCD alleviates this by performing LCD within predefined multiple segments.

\textbf{Denoising strategy.} 
We can observe from  Table~\ref{ablation} that   $\mathcal{L}_{mlcd-m}$ 
substantially enhances the performance of $\mathcal{L}_{mlcd-s}$, verifying that the proposed multistep denoising strategy is critical to perform data-free MLCD. The probable reason is our multistep MDS yields better initial latent codes, whereas the latent codes have better quality with smaller timesteps. 

\textbf{Latent LPIPS.} 
As outlined in  Table~\ref{ablation}, $\mathcal{L}_{mlcd-s}$  using L-LPIPS introduces gains on all metrics. This result denotes it is more powerful to enforce consistency in latent LPIPS space than raw latent space as latent LPIPS can make perceptual consistency.

\textbf{Data-free latent consistency distillation in stage 2.}
In ~\ref{ablation}, $\mathcal{L}_{lcd2}$ represents using multistep solver in LCD to enforce consistency across the entire timestep range. We can see that $\mathcal{L}_{lcd2}$  significantly improves  CS values of TLCM trained in stage 1. This is because $\mathcal{L}_{lcd2}$   achieves inter-segment consistency of TLCM.   
The performance is further enhanced by substituting $\mathcal{L}_{lcd2}$ with $\mathcal{L}_{ilcd2}$. The reason lies in that it is easier to make consistency along the sparse predefined timesteps than the entire timestep range. 

\textbf{MHP optimization.} 
Table~\ref{ablation}  shows that adding $\mathcal{L}_{mhp}$ to the losses in line 7 introduces gains in terms of CS and IR. This result indicates that our MHP optimization method is capable of improving the text-image alignment and human preference of TLCM.  

\textbf{Data-free DM.} 
We can see in Table~\ref{ablation} using our data-free DM loss   $\mathcal{L}_{dfdm}$ leads to the performance improvements on all metrics. 
This result demonstrates that our DM in a data-free way is compatible with the proposed distillation method, boosting TLCM's performance. 

\textbf{Discriminator.} 
We also observe in  Table~\ref{ablation}  that discriminator loss $\mathcal{L}_{gan}$ improves CS, AS, and IR since the discriminator facilitates consistency in probability distribution space, which is critical for the low-step regime.

\textbf{Teacher's inference steps of  data-free latent  consistency distillation in stage 2.} 
In Table~\ref{ablation1},  we study the effect concerning the teacher's sampling steps of data-free  LCD in stage 2. The results show as the sampling step increases from 1 to 4, the performance is consistently improved. 
Therefore, it is crucial to perform multi-step denoising to estimate $\hat{z}_{t_{n}}$. The reason is that multi-step solvers are capable of reducing discretization error for big skipping step. 

\begin{table}[t]\small
\setlength{\tabcolsep}{3pt}
\caption{Ablation study of TLCM with respect to  latent LPIPS, data-free LCD with single denoising  step  ($\mathcal{L}_{lcd-s}$),  data-free MLCD  with single denoising iteration ($\mathcal{L}_{mlcd-s}$), data-free MLCD with MDS ($\mathcal{L}_{mlcd-m}$), data-free  LCD in stage 2 ($\mathcal{L}_{lcd2}$), improved data-free  LCD in stage 2 ($\mathcal{L}_{ilcd2}$), data-free DM  ($\mathcal{L}_{dfdm}$),   
 multi-dimensional human preference ($\mathcal{L}_{mhp}$), adversarial ($\mathcal{L}_{gan}$). All the models adopt a 4-step sampler and SDXL backbone.}
\label{ablation}
\begin{center}
\begin{tabular}{llllllllllllll}\label{table:ablation}
\\ \hline 
L-LPIPS & $\mathcal{L}_{lcd-s}$  &$\mathcal{L}_{mlcd-s}$  &$\mathcal{L}_{mlcd-m}$ &$\mathcal{L}_{lcd2}$&$\mathcal{L}_{ilcd2}$   &$\mathcal{L}_{mhp}$ &$\mathcal{L}_{dfdm}$    &$\mathcal{L}_{gan}$ & CS & FID&AS & IR \\
\hline
&\checkmark &&&&&&&&  31.61 &32.90&5.89&0.41\\
&&\checkmark &&&&&&& 31.76&27.01&5.98&0.58 \\
\checkmark&&\checkmark &&&&&&& 31.99&27.61 & 5.92&0.61\\
\checkmark&&&\checkmark &&&&&& 32.31&30.99&6.01&0.69\\
\checkmark&&&\checkmark &\checkmark&&&&& 32.74&32.05&6.00&0.72\\

\checkmark&&&\checkmark&~ &\checkmark&&&& 33.06&25.44&5.96&0.77 \\
\checkmark&&&\checkmark&~ &\checkmark&\checkmark&&&33.16&28.40&6.01&0.90 \\
\checkmark&&&\checkmark&~ &\checkmark&\checkmark&\checkmark&~&33.32&30.58&6.03&0.97  \\
\checkmark&&&\checkmark&~ &\checkmark&\checkmark&\checkmark&\checkmark&33.52& 30.33&6.06&1.01 \\
\hline
\end{tabular}
\end{center}
\end{table}

\begin{table}[t]\small
\caption{Performance comparison  of the teacher's sampling steps for data-free LCD in stage 2.}
\label{ablation1}
\begin{center}
\begin{tabular}{llllllllll}
\\ 
\hline 
Step& CS & FID&AS & IR &Step& CS & FID&AS & IR \\
\hline
1& 32.78  & 26.19 &5.95 & 0.66&  2&32.97&25.73&5.95&0.71\\

3&33.06 &25.44  &5.96 & 0.77 &4&33.10&25.18&5.97&0.78\\

\hline
\end{tabular}
\end{center}
\end{table}

\section{Conclusion}
In this paper, we propose Training-efficient Latent Consistency Model (TLCM), a novel approach for accelerating text-to-image
latent diffusion models using only 70 A100 hours,  without requiring any text-image paired data. 
TLCM can generate high-quality, delightful images with only 2-8 sampling steps and achieve better image quality than baseline methods while being compatible with image style transfer, controllable generation, and Chinese-to-image generation.


\bibliography{iclr2025_conference}
\bibliographystyle{iclr2025_conference}
\newpage
\appendix
\section{Appendix}

\subsection{Algorithms}\label{alg}

\begin{algorithm}[ht]
   \caption{Data-free multistep latent consistency  distillation}\label{alg:al1}
        \textbf{Input:} Gaussian noise $\epsilon$,  timestep ${t_m}$, segment index $s$, teacher model $\epsilon_{\theta_0}$, student model $g_{\theta}$, text condition $c$, segment number $M$ \\
        Initialize $z_{T} $ with $\epsilon$, calculate denoising steps $L=M-s$,
        time interval $\triangle T=(T-t_m)/L$ \\
        \For {$i\quad \text{in} \quad \{0,1,\cdots,L-1\} $ }
            {Calculate $t=T-i*\triangle T,\quad t_{m'}=t-\triangle T$\\
            Calculate $z_{t_{m'}}=\Psi(\hat{\epsilon}_{\theta_0}(z_{t},c,w,t),t,t_{m'})$
            } 
        Calculate $z_{t_n}$ using Equation~(\ref{eq:ddim}) \\
        Perform MLCD using Equation~(\ref{eq: mlcd})
\end{algorithm}

\begin{algorithm}[ht]
   \caption{Data-free latent consistency  distillation in stage 2}\label{alg:al2}
        \textbf{Input:} Gaussian noise $\epsilon$,  timestep ${t_m}$, teacher model $\epsilon_{\theta_0}$, student model $f_\theta$, text condition $c$, segment number $M$, denoising step of teacher $p$, denoising step $q$ of student, diffusion coefficient sequence $\alpha_{1:T}$, timestep milestones $\{t_{step}^s\}_{s=0}^M$\\
        Initialize $\hat{z}_T $ with $\epsilon$  and timestep $t$ with $T$\\
        \For {$i\quad \text{in} \quad \{0,1,\cdots,q-1\} $ }
            {Calculate $\hat{z}_0=\dfrac{\hat{z}_t-\sqrt{1-\alpha_t}f_{\theta}(\hat{z}_t,t,c)}
            {\sqrt{\alpha_t}}$\\
            Calculate $t=T-T/q\times(i+1)$, 
            Calculate  $\hat{z}_t=\sqrt{\alpha_t} \hat{z}_0 + \sqrt{1-\alpha_t}\epsilon $ 
            }

        Randomly sample $t_m$ from $\{t_{step}^s\}_{s=1}^M$,
        detach $\hat{z}_0$  and calculate $\hat{z}_{t_m}$ by forward diffusion $\hat{z}_{t_m}=\sqrt{\alpha_{t_m}} \hat{z}_0 + \sqrt{1-\alpha_{t_m}}\epsilon$\\
        \For {$i\quad \text{in} \quad \{0,1,\cdots,p-1\} $ }
        {
            Calculate $t_1=t_m-(T/M)/p\times i$ and $t_2=t_m-(T/M)/p\times (i+1)$\\
            Calculate $\hat{z}_{t_2}$ using Equation~(\ref{eq:ddim}) based on current state $\hat{z}_{t_1}$
        }
        Perform LCD using Equation~(\ref{eq:ilcd2})
\end{algorithm}

\begin{figure}[ht]
    \centering
    \parbox{0.18\linewidth}{\centering{Source}}
    \parbox{0.18\linewidth}{\centering{Japanese comics}}
    \parbox{0.18\linewidth}{\centering{Ink and wash \\style}}
    \parbox{0.18\linewidth}{\centering{Pixar\\dreamworks}}
    \parbox{0.18\linewidth}{\centering{Van Gogh's \\paintings}}
    
    \includegraphics[width=0.18\linewidth]{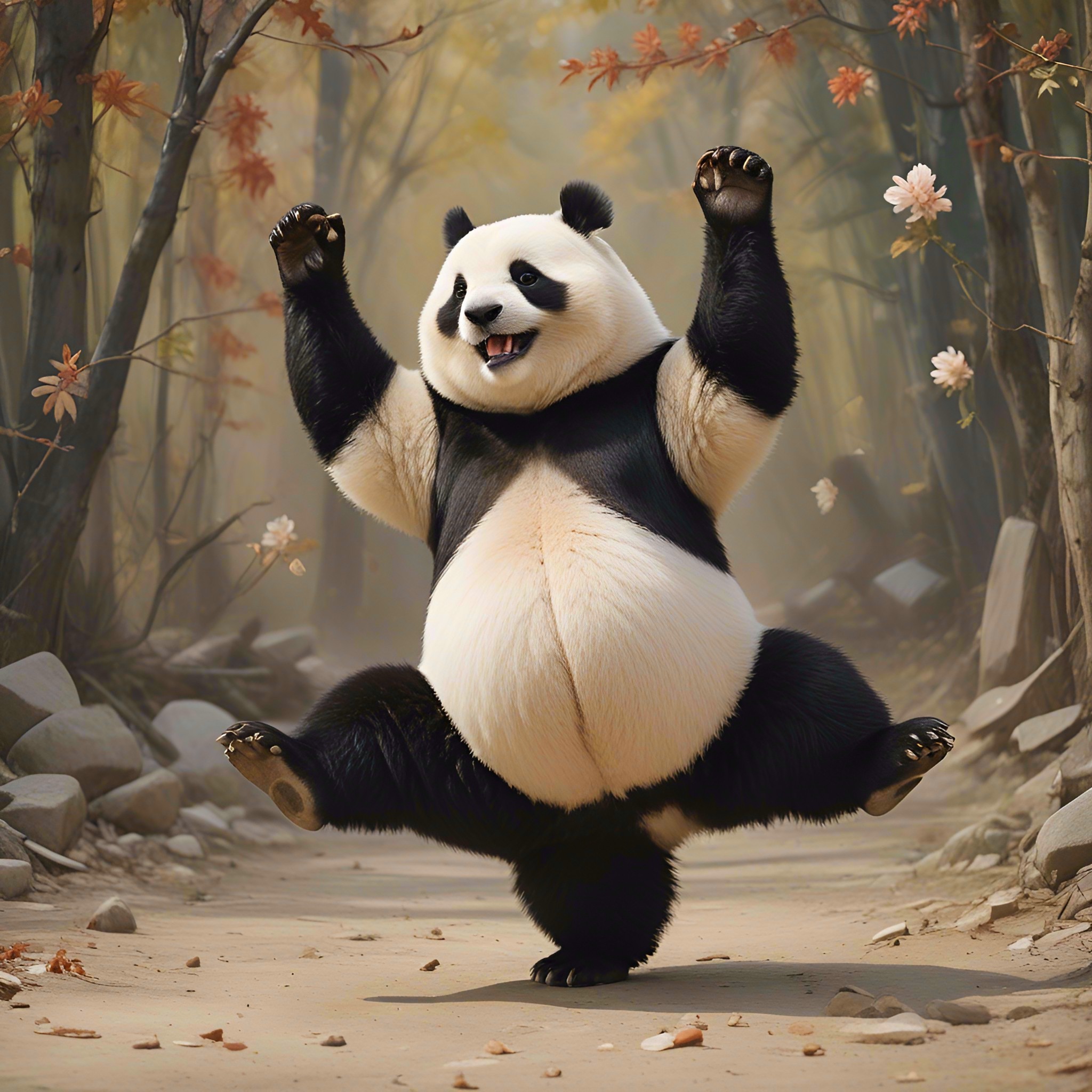}
    \includegraphics[width=0.18\linewidth]{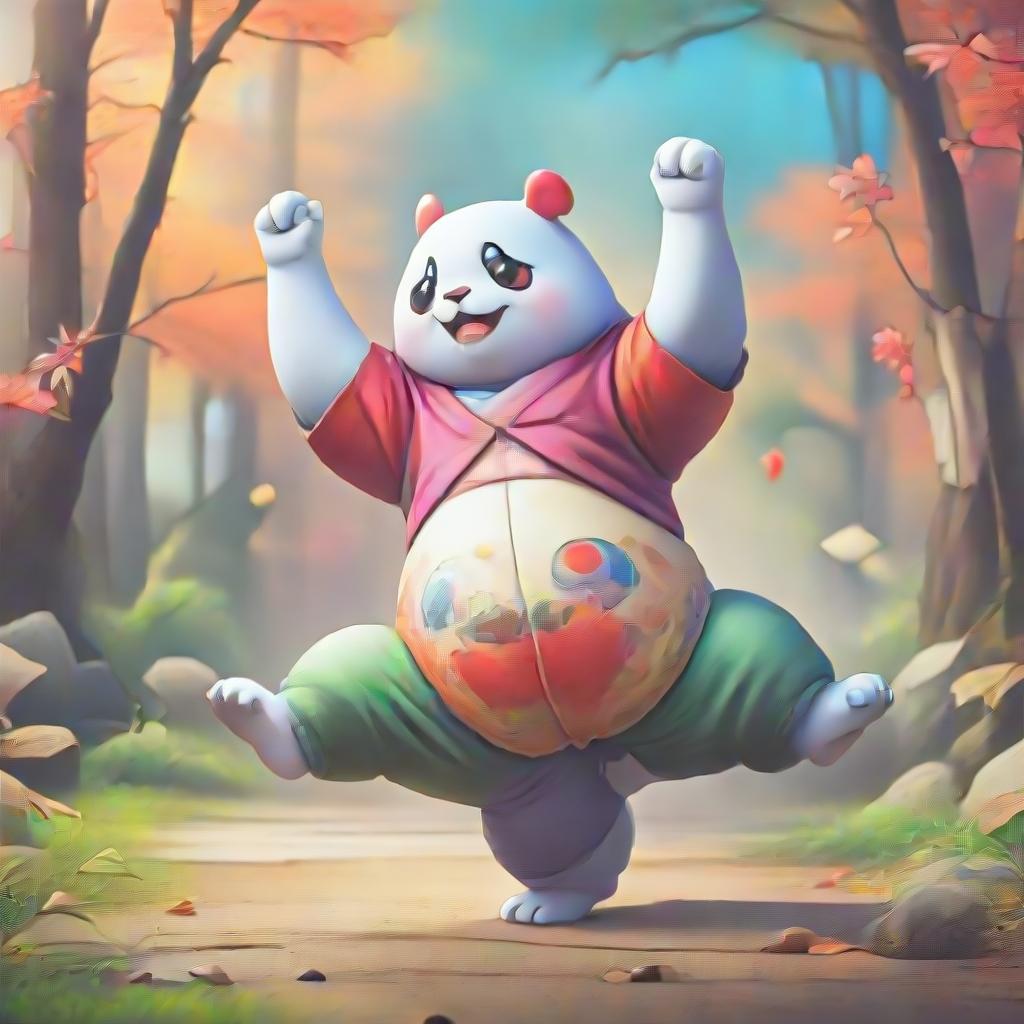}
    \includegraphics[width=0.18\linewidth]{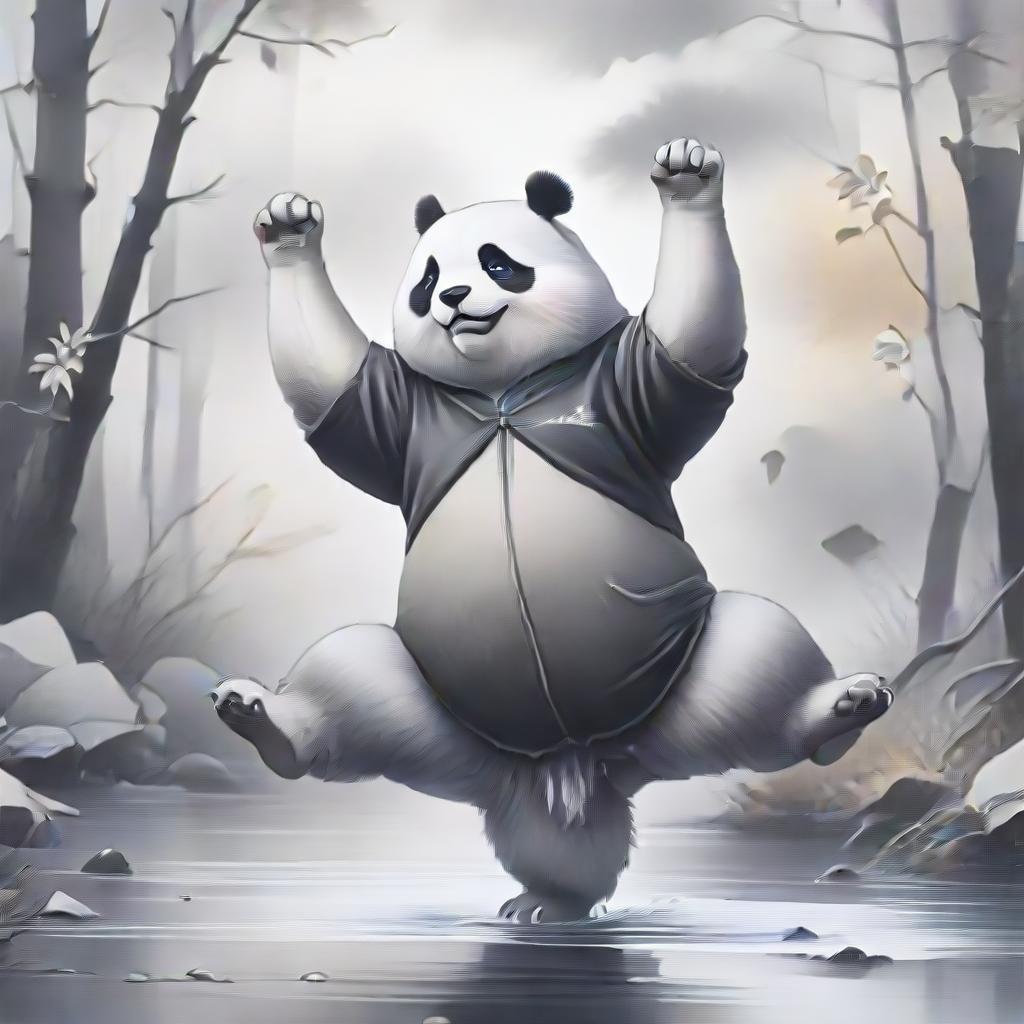}
    \includegraphics[width=0.18\linewidth]{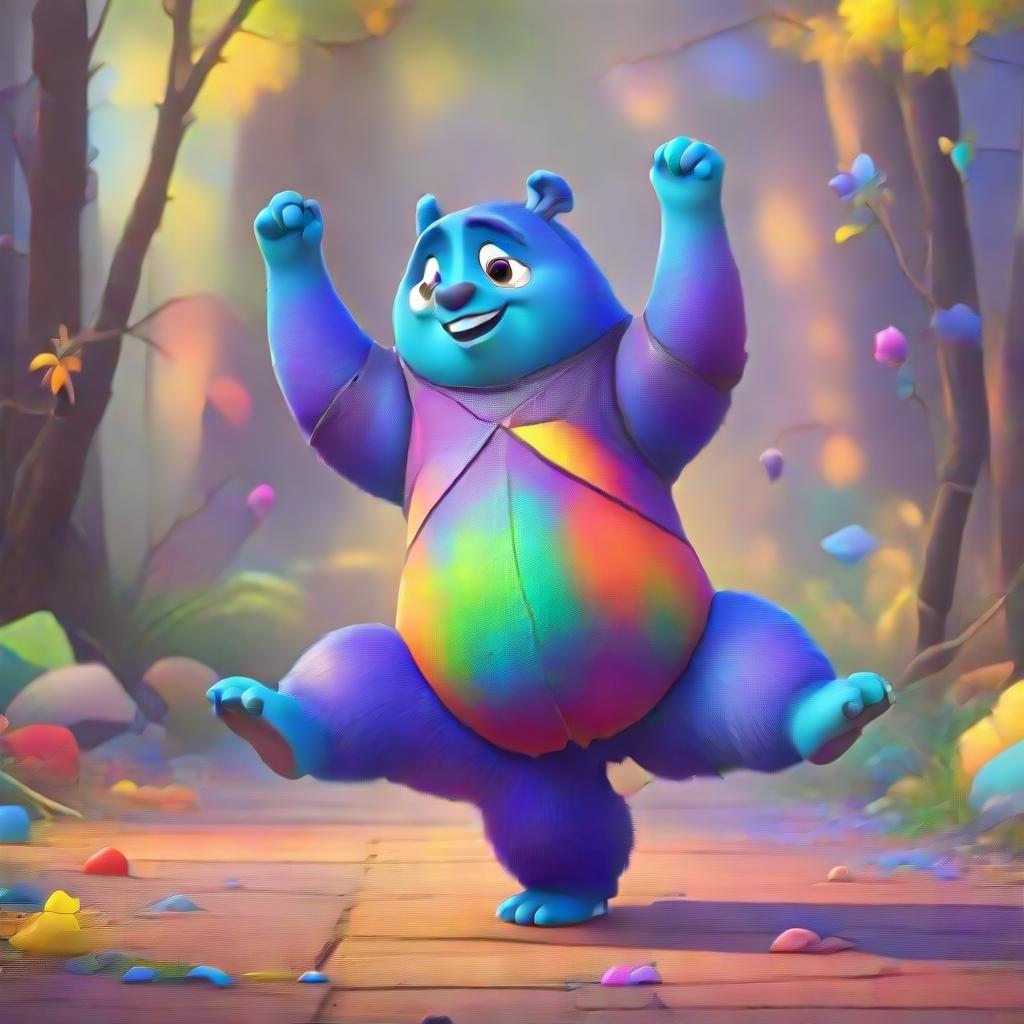}
    \includegraphics[width=0.18\linewidth]{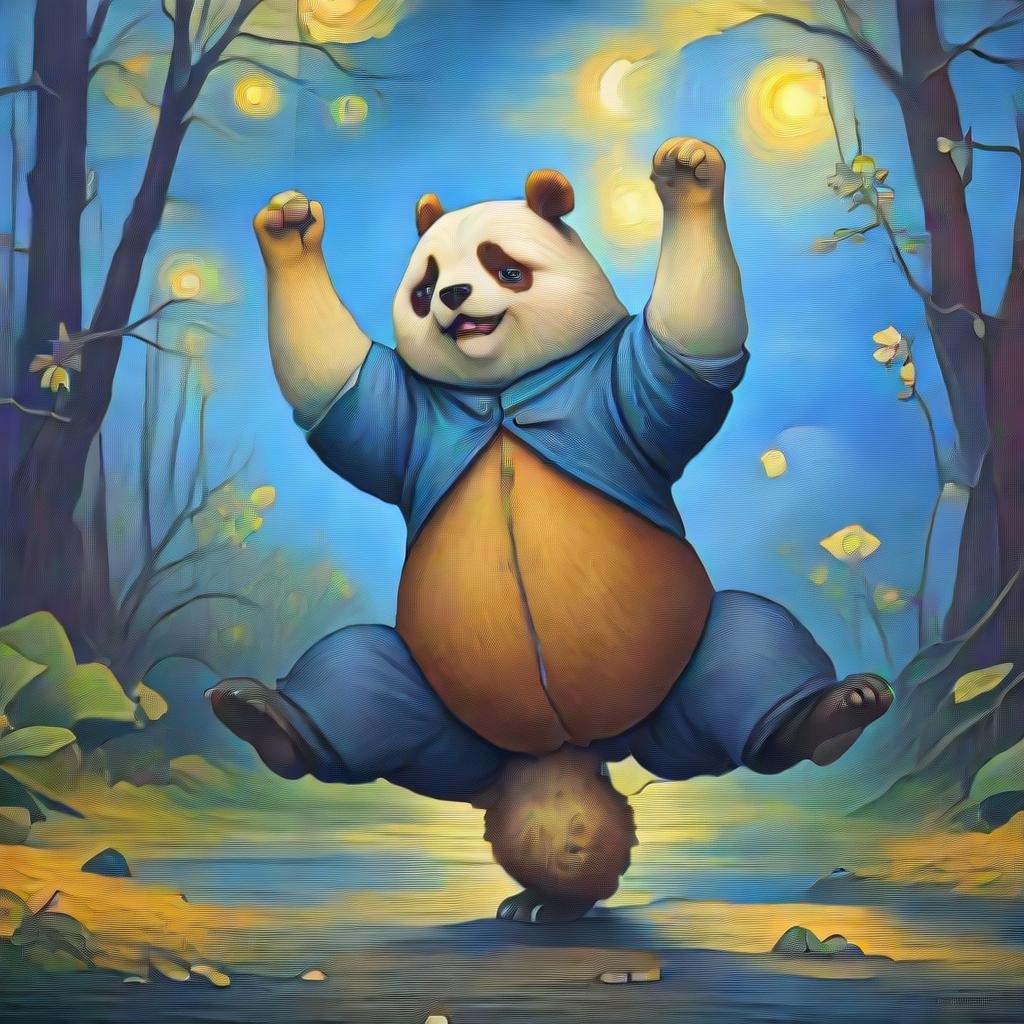}

    \includegraphics[width=0.18\linewidth]{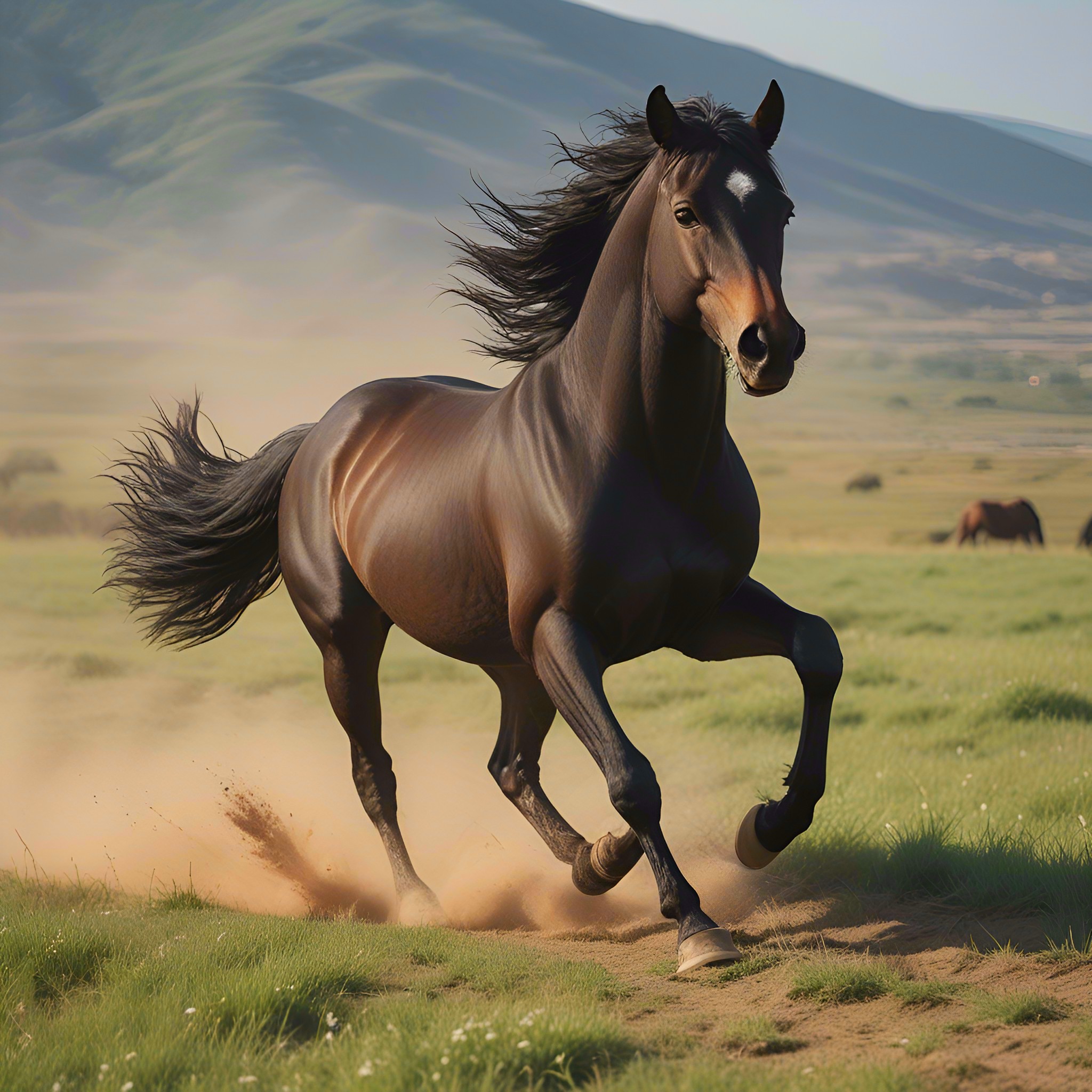}
    \includegraphics[width=0.18\linewidth]{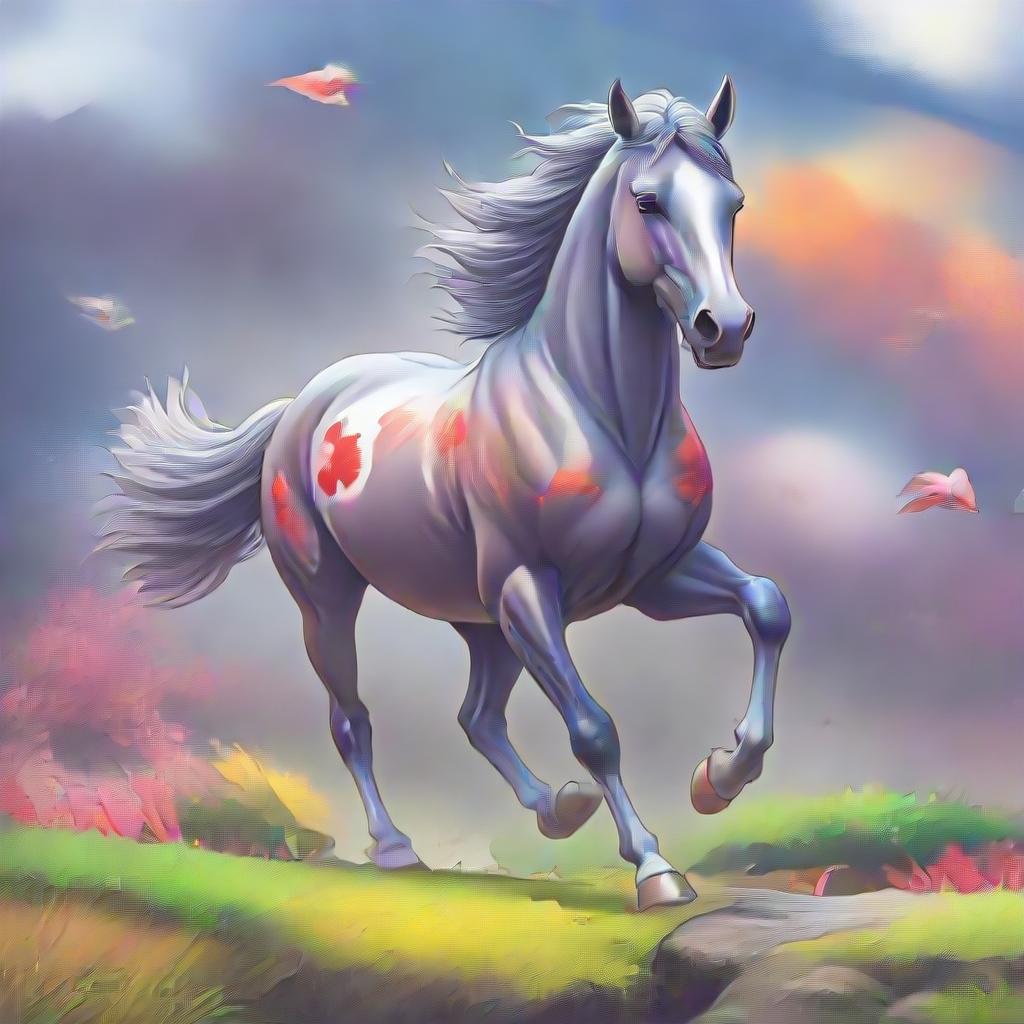}
    \includegraphics[width=0.18\linewidth]{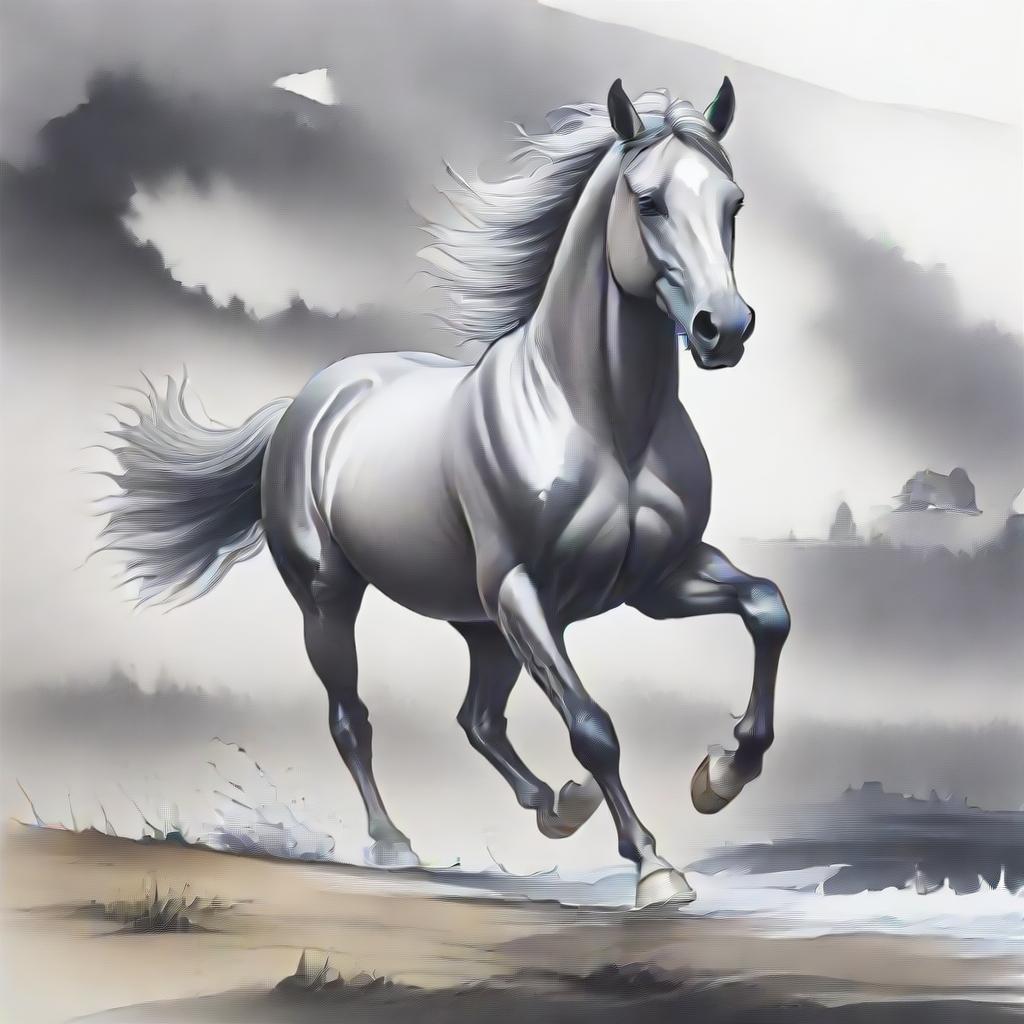}
    \includegraphics[width=0.18\linewidth]{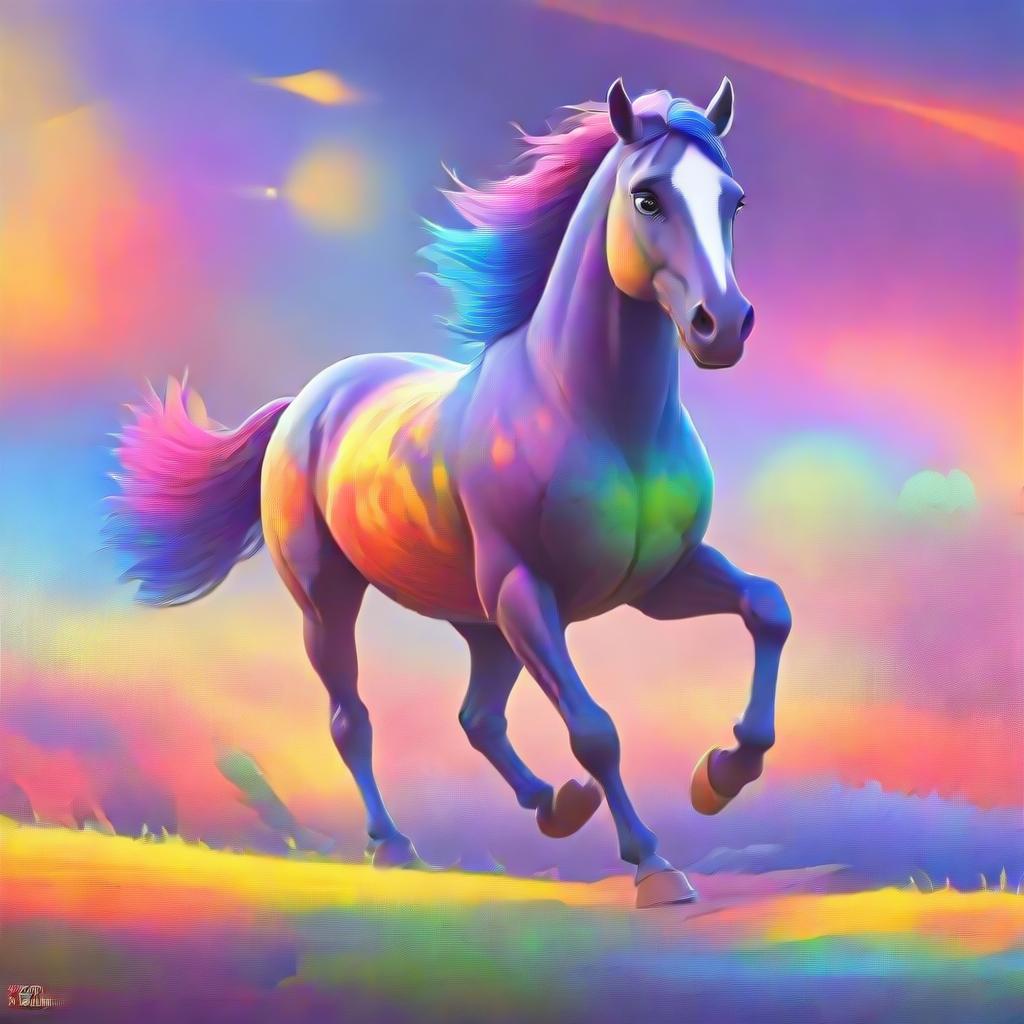}
    \includegraphics[width=0.18\linewidth]{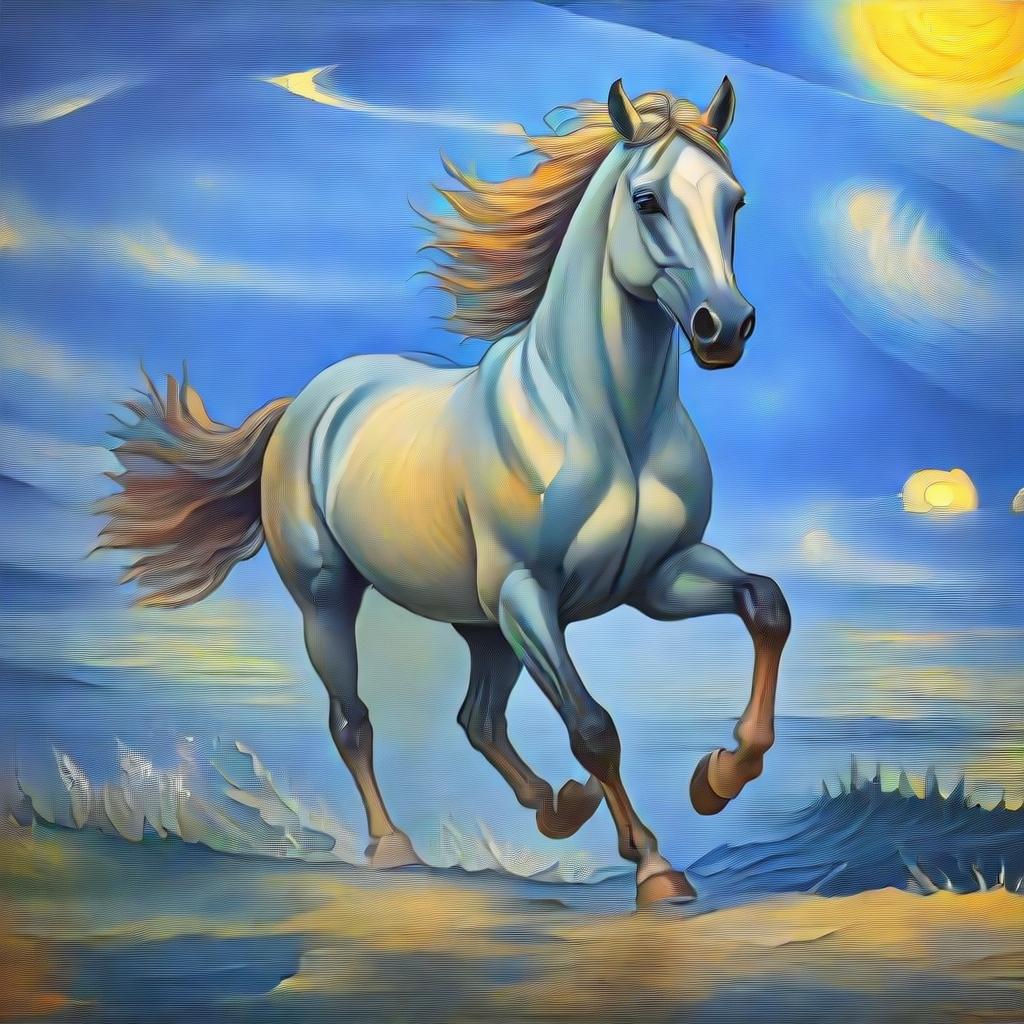}

    \includegraphics[width=0.18\linewidth]{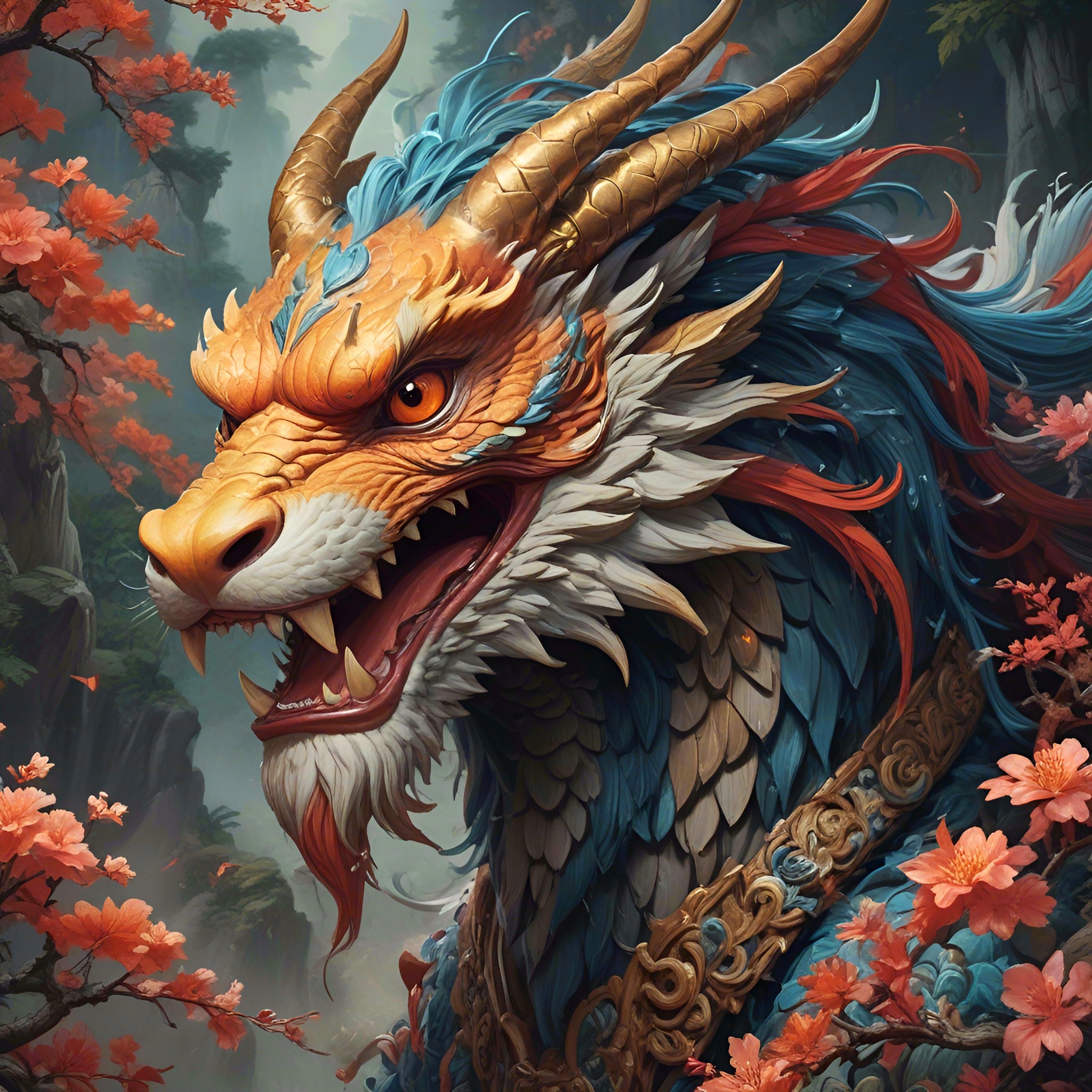}
    \includegraphics[width=0.18\linewidth]{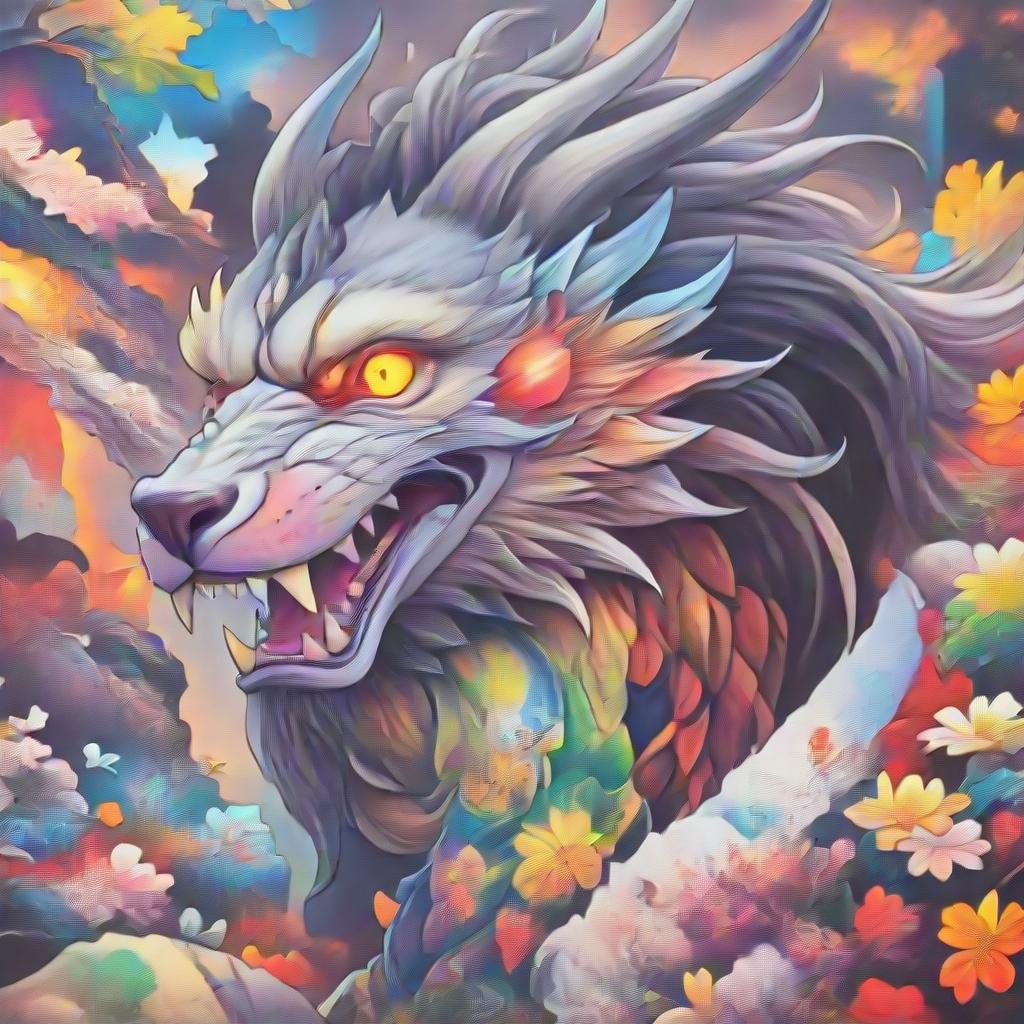}
    \includegraphics[width=0.18\linewidth]{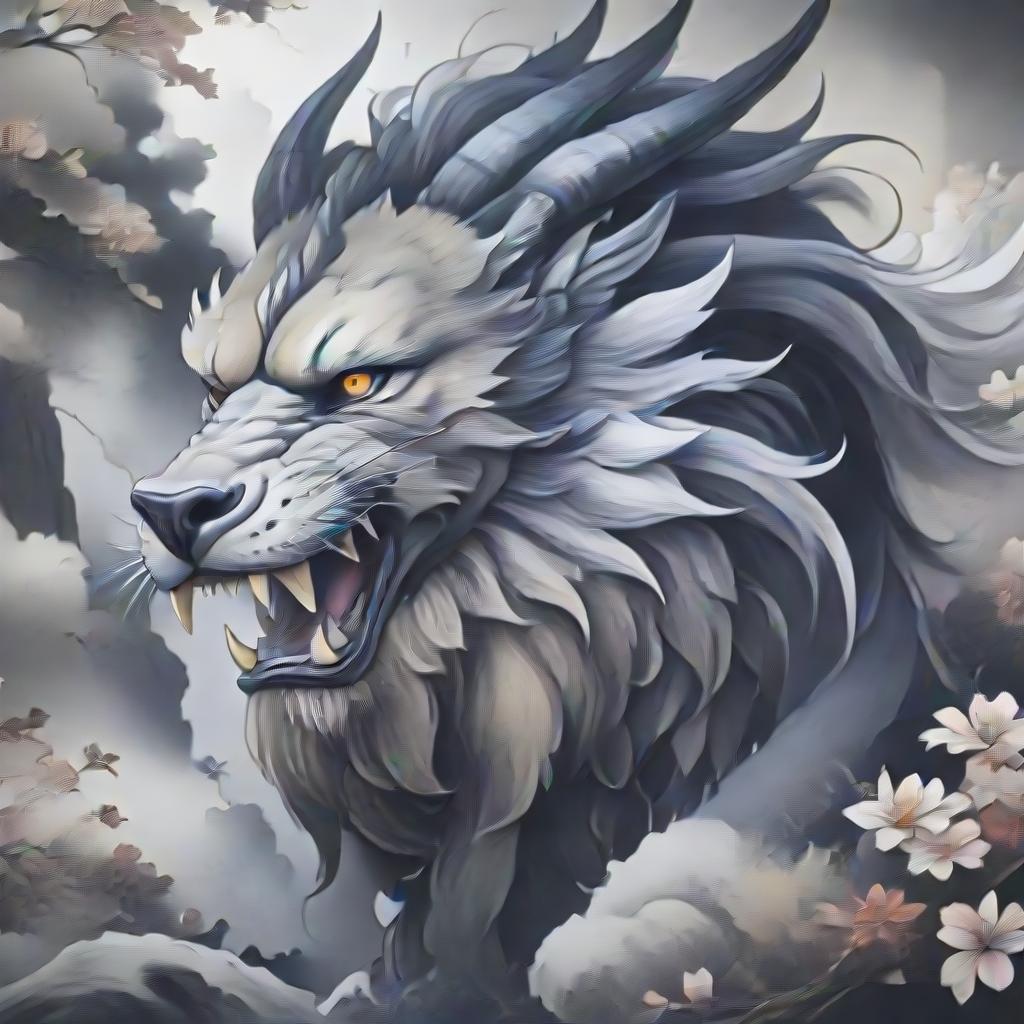}
    \includegraphics[width=0.18\linewidth]{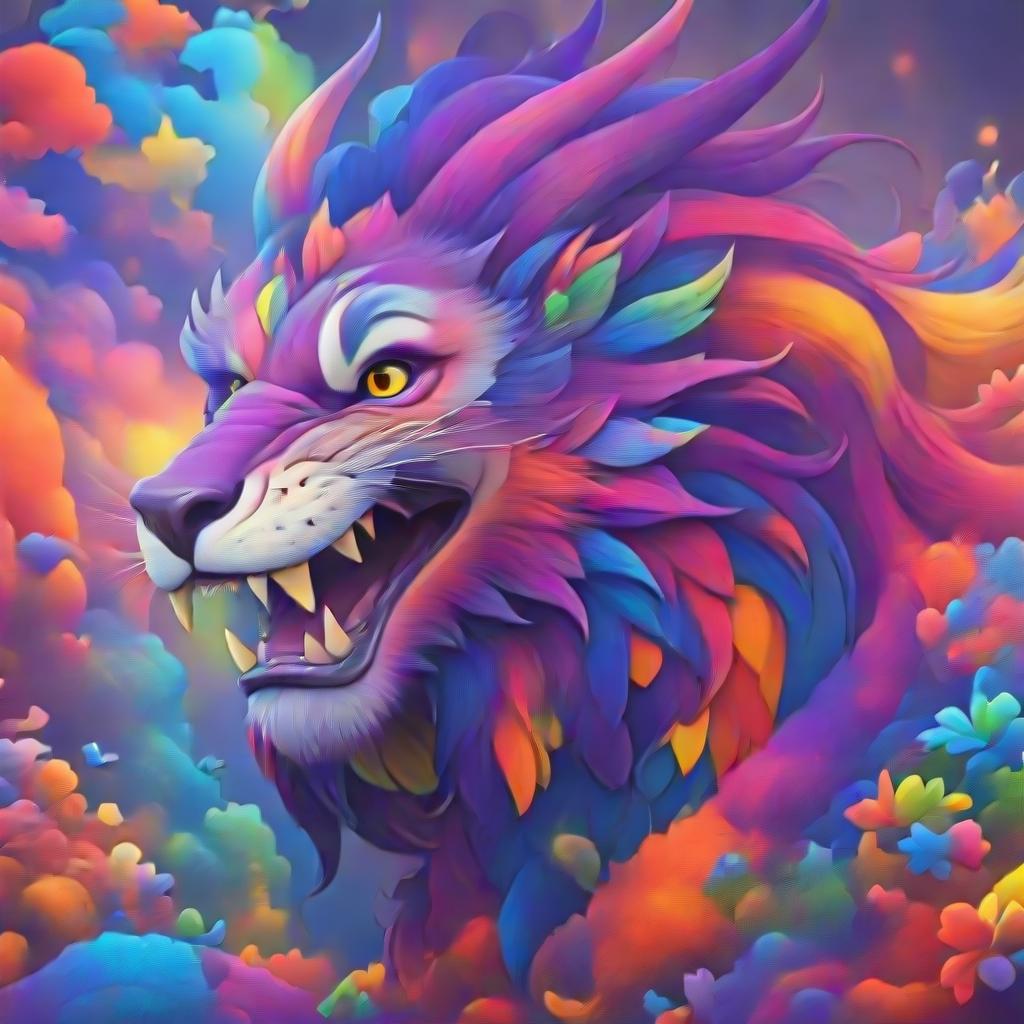}
    \includegraphics[width=0.18\linewidth]{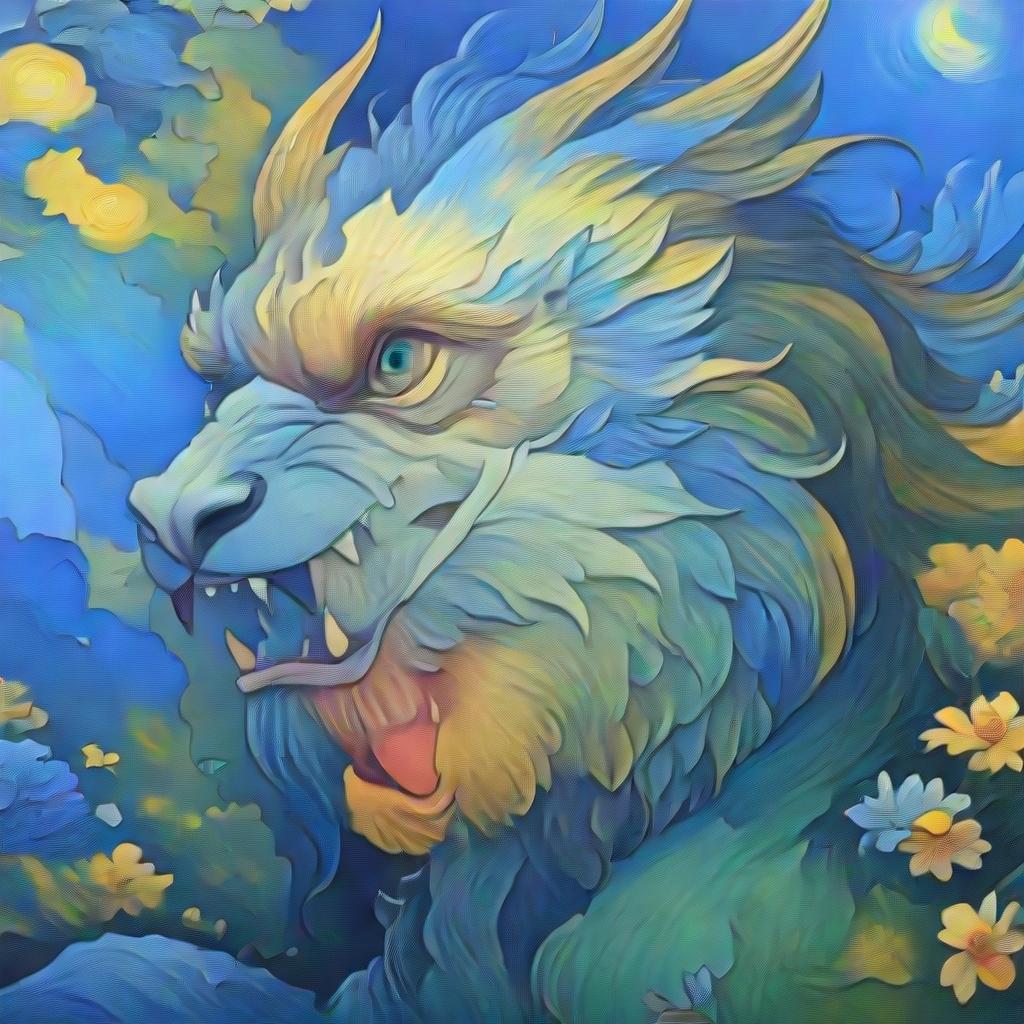}

    \includegraphics[width=0.18\linewidth]{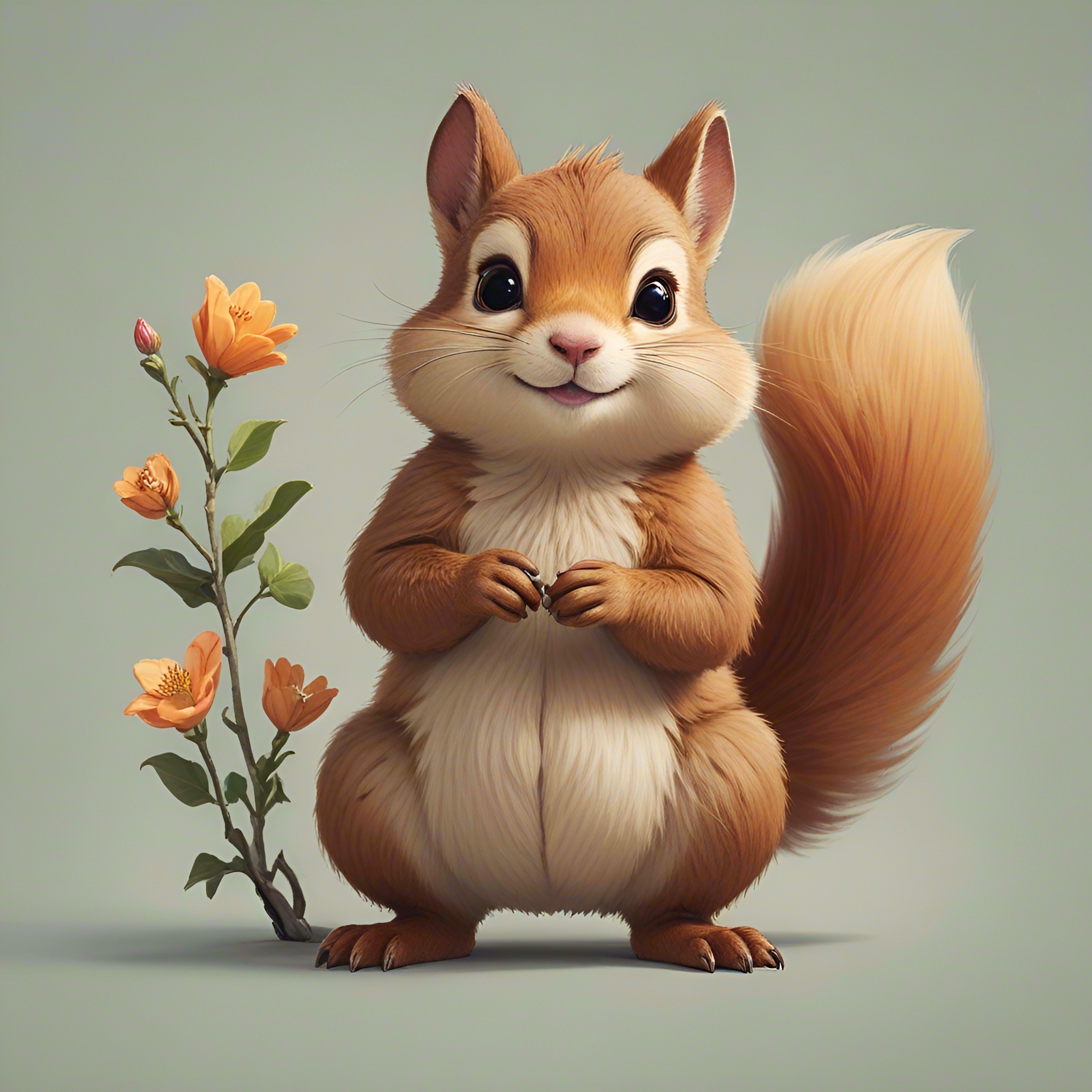}
    \includegraphics[width=0.18\linewidth]{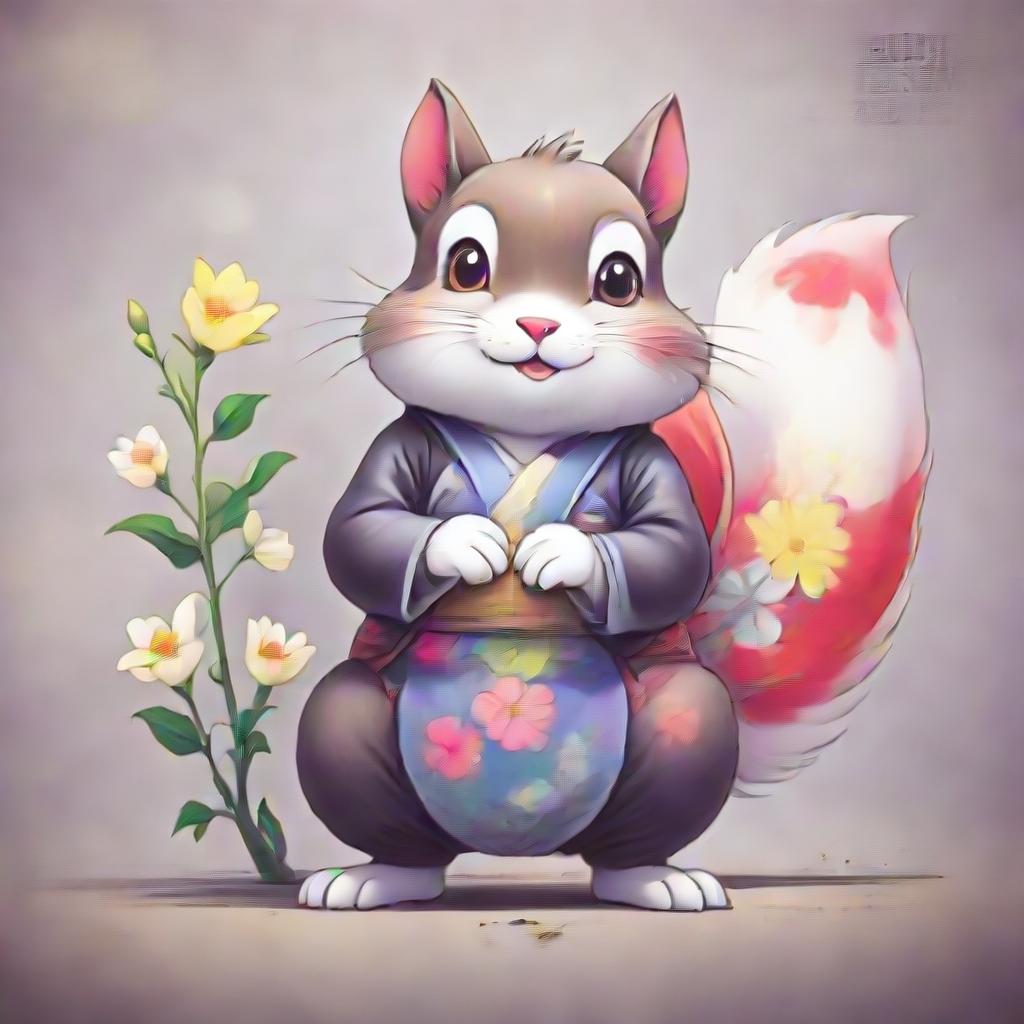}
    \includegraphics[width=0.18\linewidth]{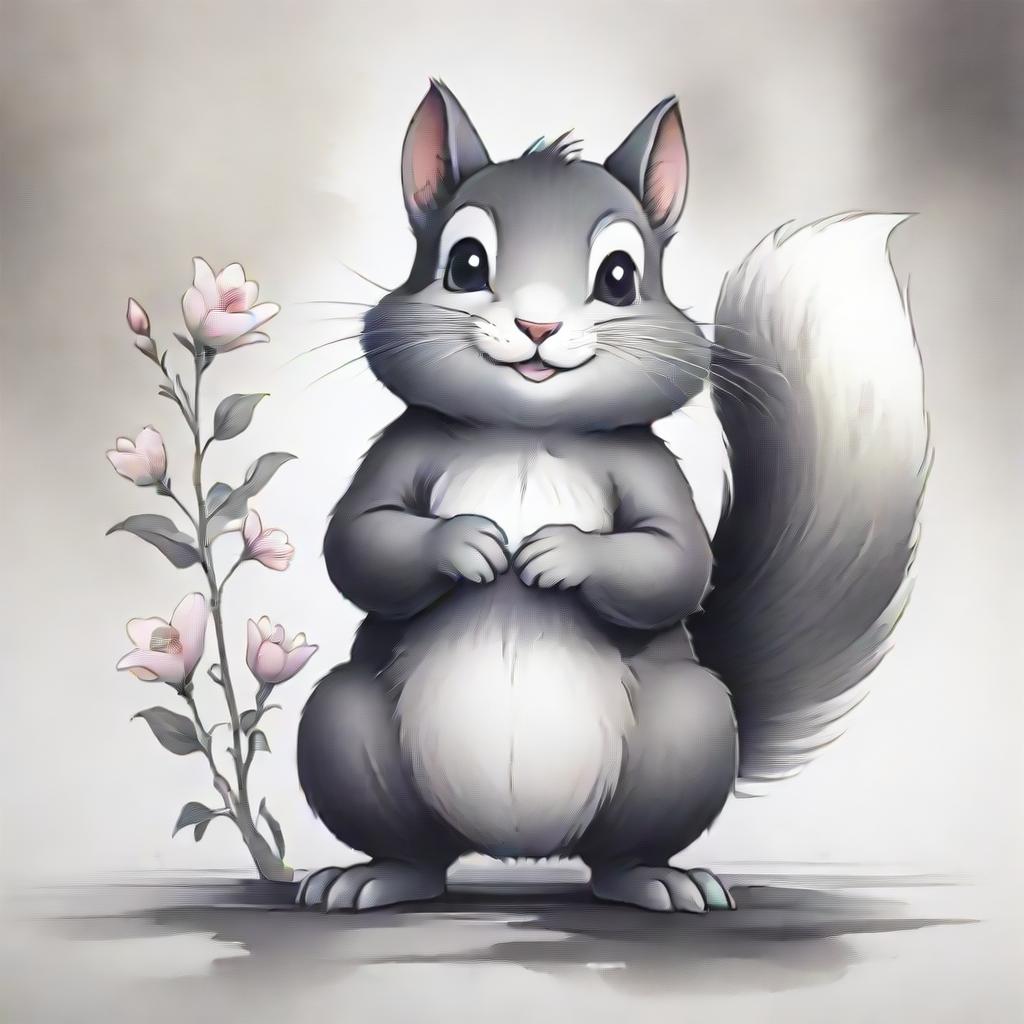}
    \includegraphics[width=0.18\linewidth]{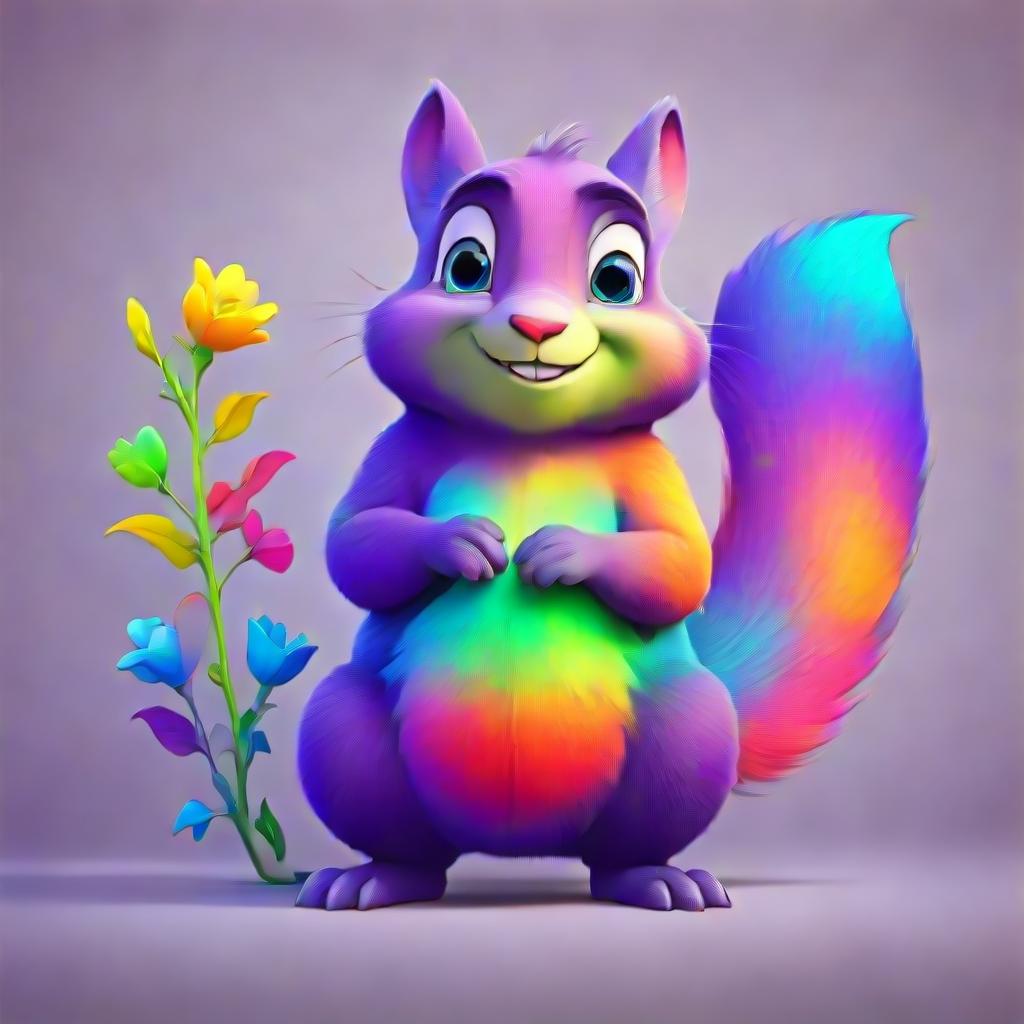}
    \includegraphics[width=0.18\linewidth]{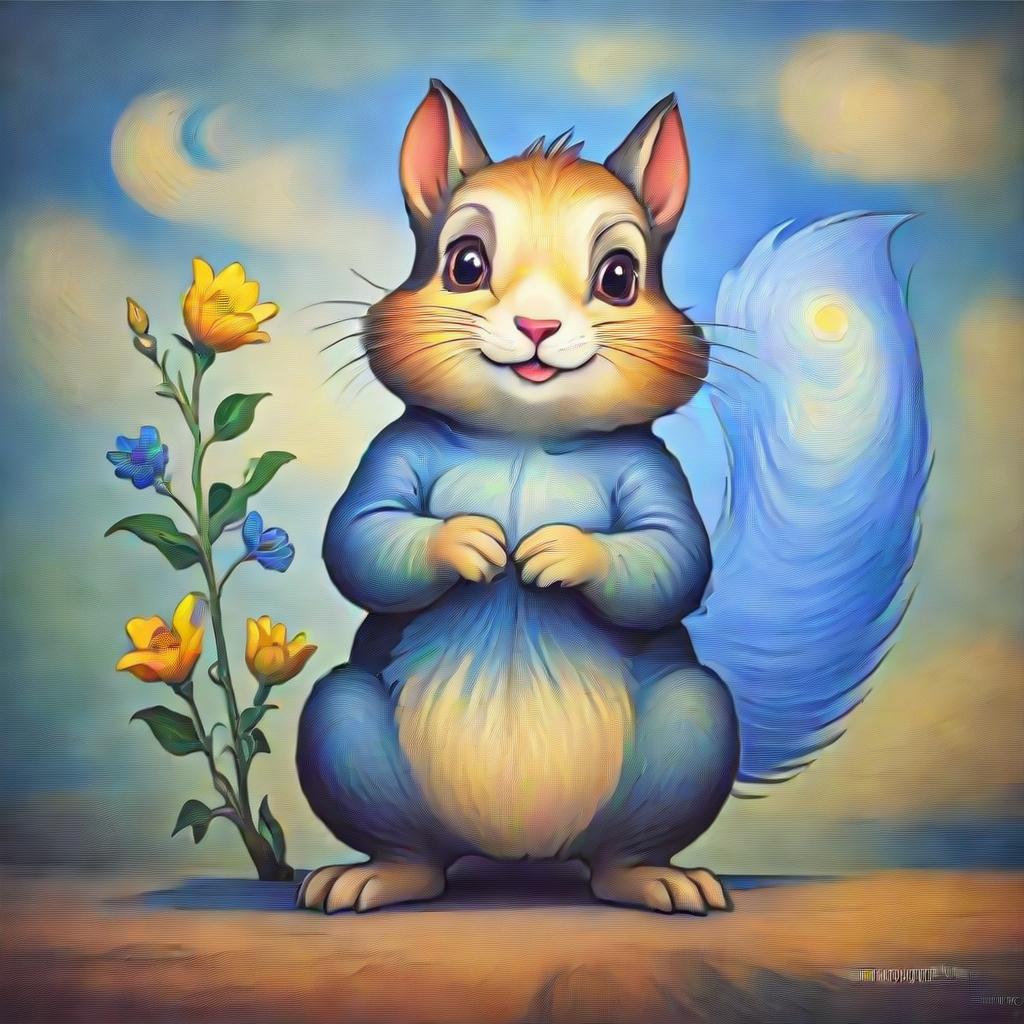}

    \caption{TLCM with image style transfer.
    The styles are presented at the top, and we apply image style transfer on the source image with our TLCM.
    Two-step sampling can produce highly stylized images with excellent results.}
    \label{fig:style}
\end{figure}
\begin{figure}[ht]
    \centering
    \includegraphics[width=0.19\linewidth]{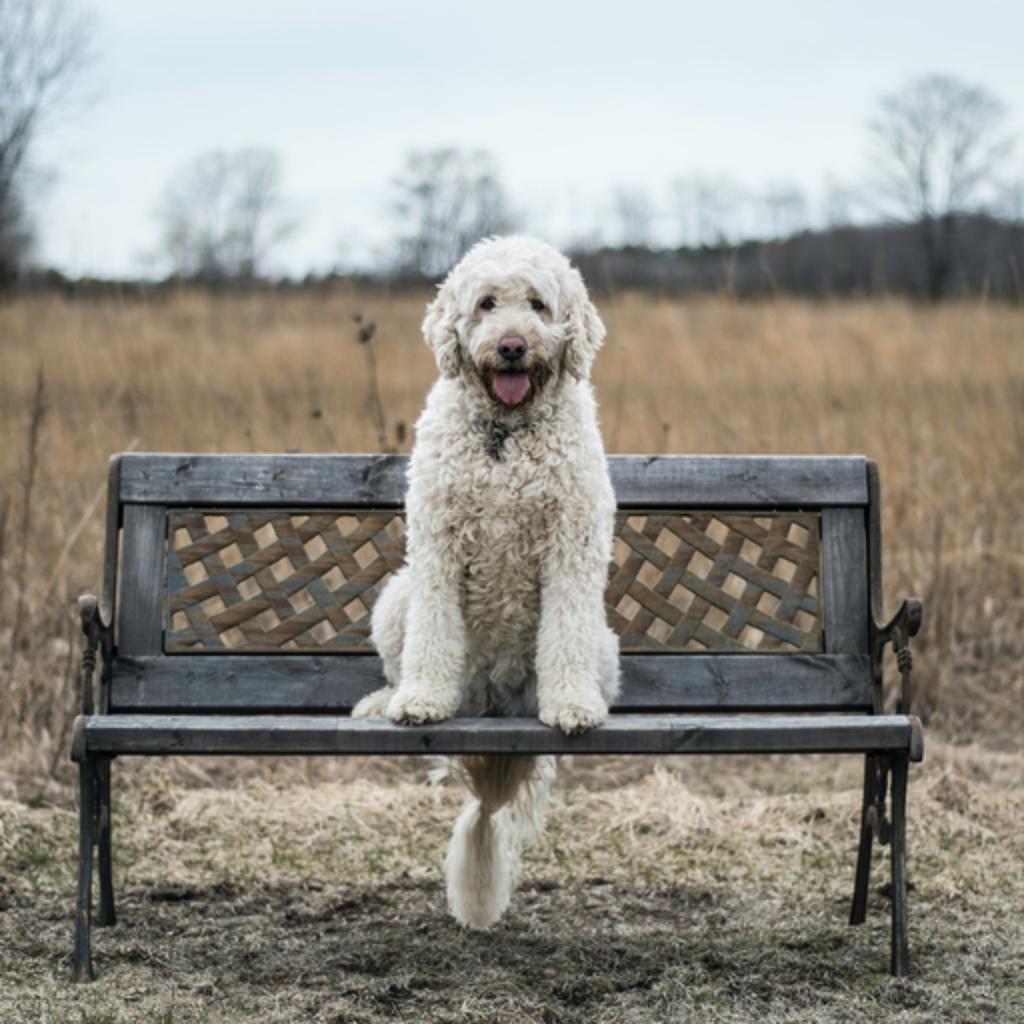}
    \includegraphics[width=0.19\linewidth]{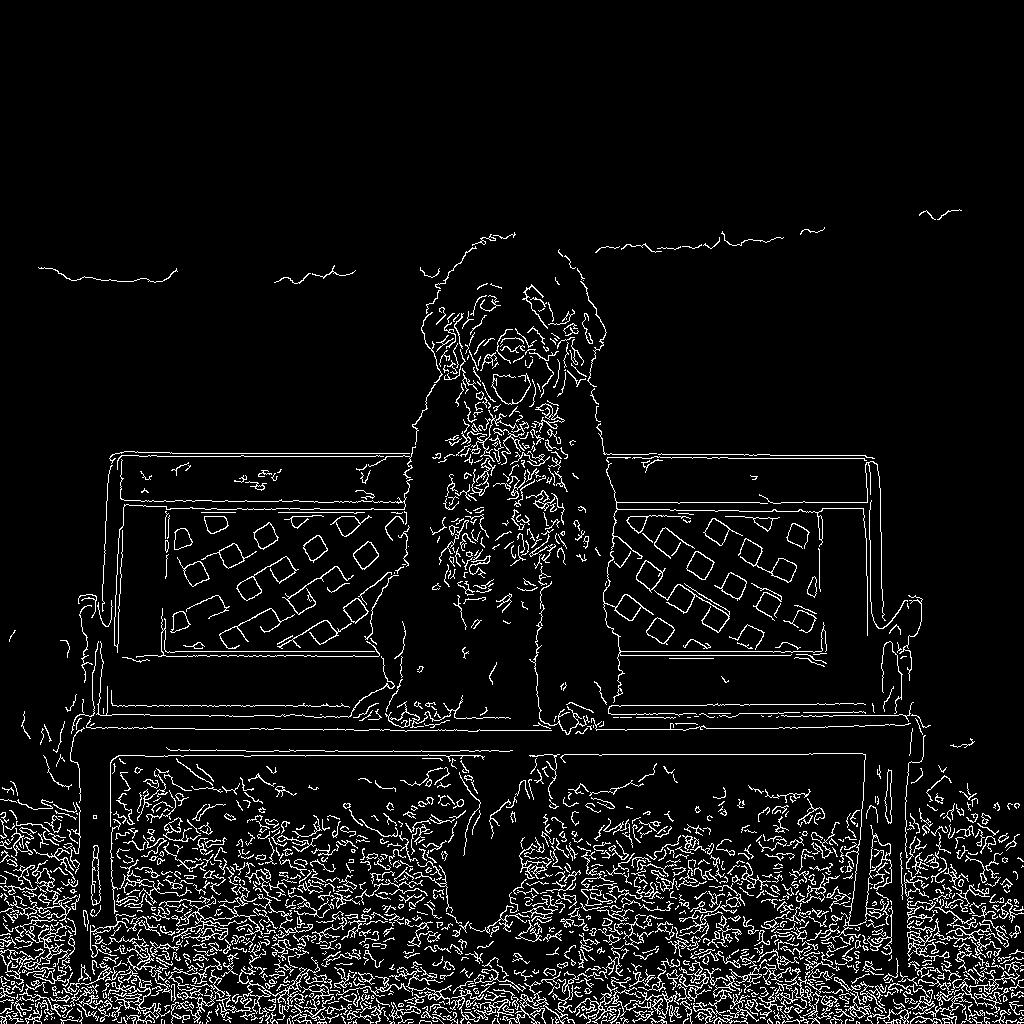}
    \includegraphics[width=0.19\linewidth]{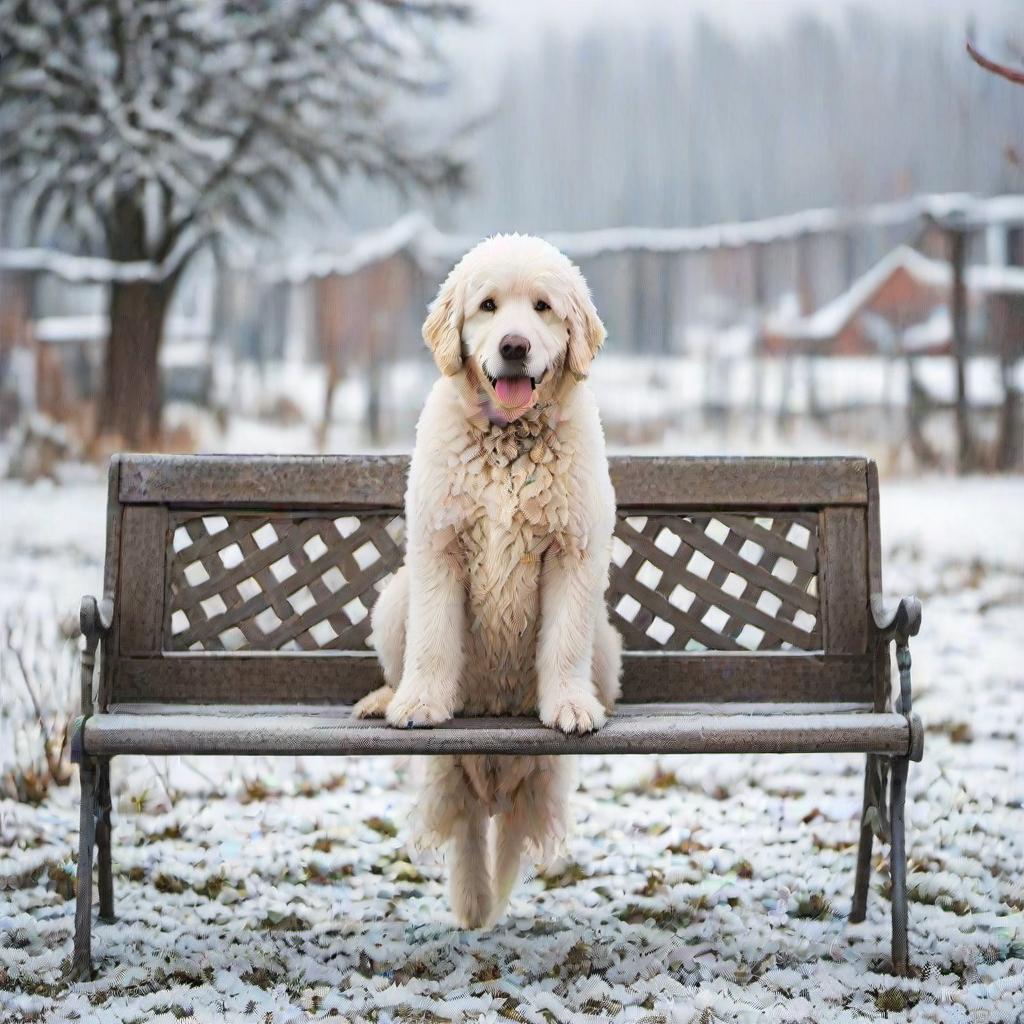}
    \includegraphics[width=0.19\linewidth]{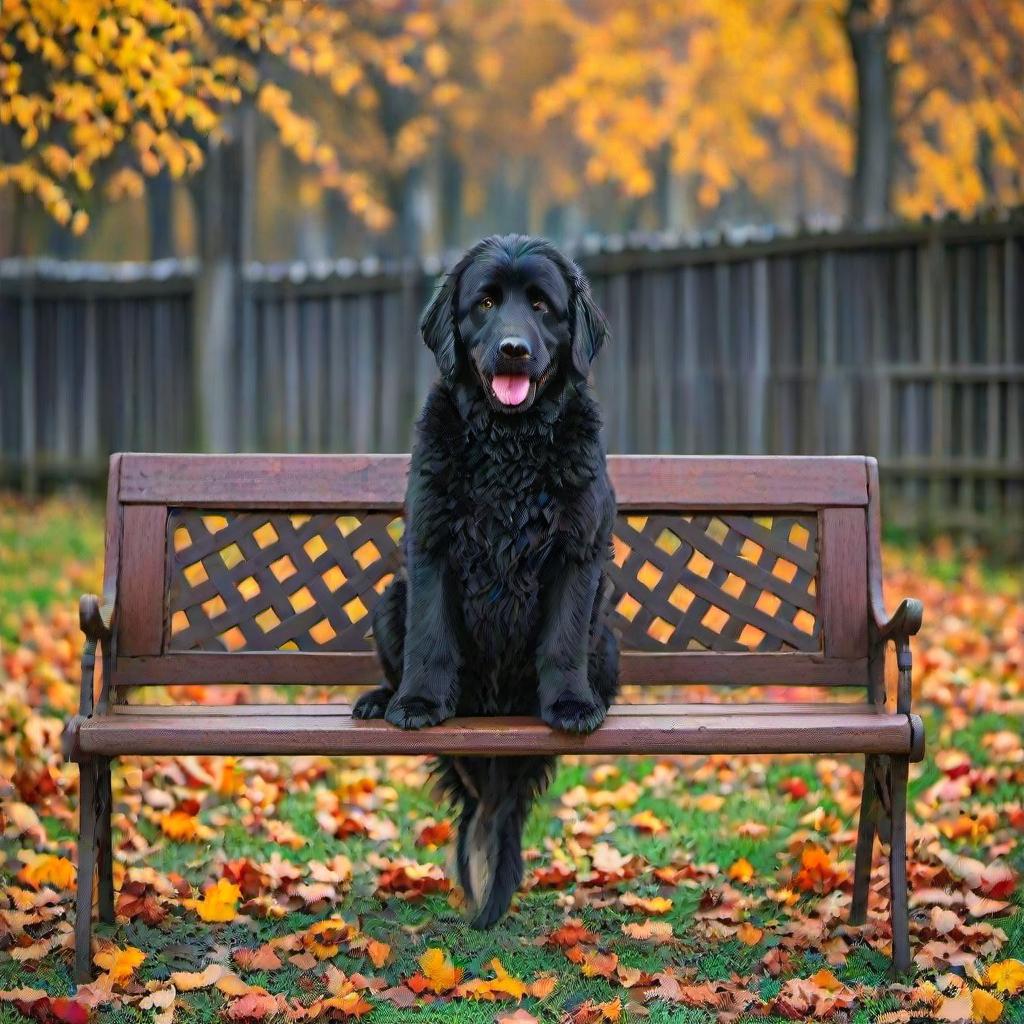}
    \includegraphics[width=0.19\linewidth]{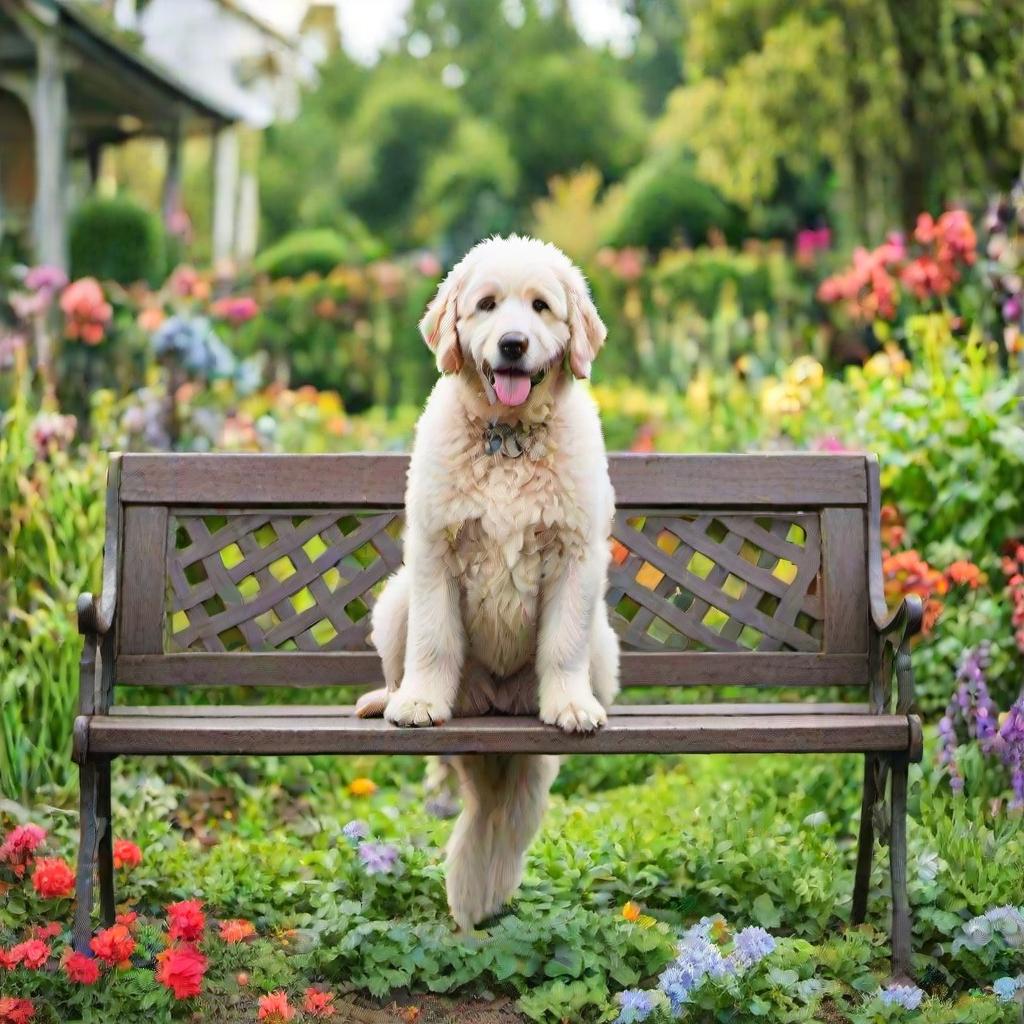}
    \parbox{0.19\linewidth}{\centering Source}
    \parbox{0.19\linewidth}{\centering Canny edge}
    \parbox{0.19\linewidth}{\centering A dog \\ \centering in the winter}
    \parbox{0.19\linewidth}{\centering A black dog \\ \centering in the autumn}
    \parbox{0.19\linewidth}{\centering A beautiful dog \\ \centering in the garden}

    \includegraphics[width=0.19\linewidth]{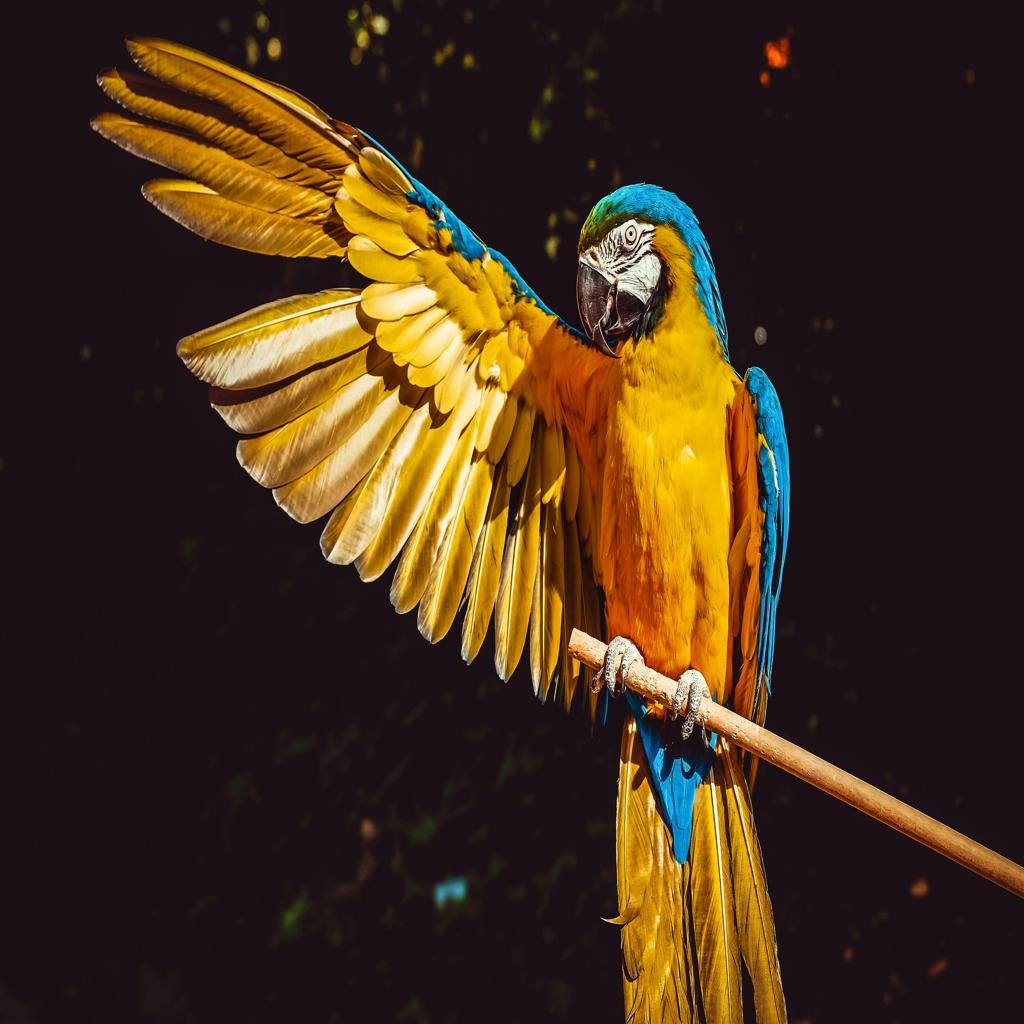}
    \includegraphics[width=0.19\linewidth]{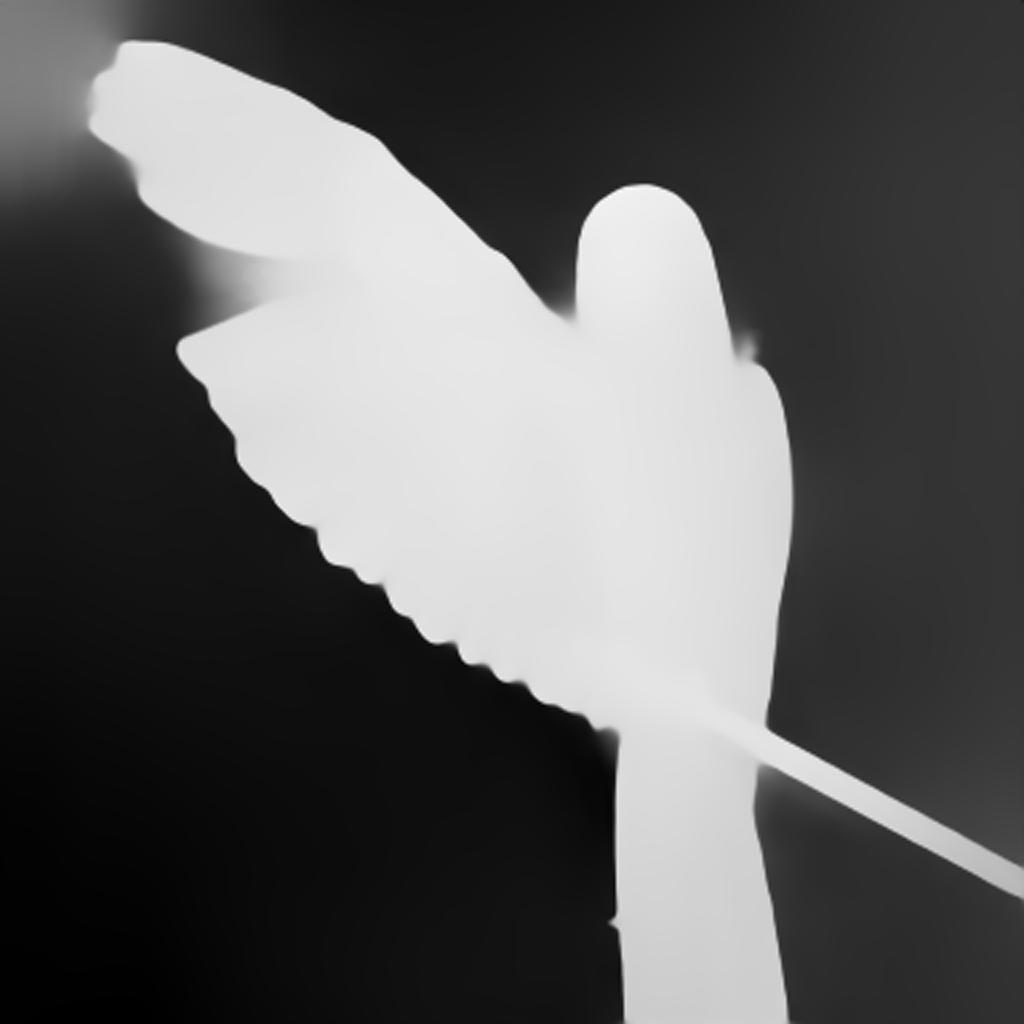}
    \includegraphics[width=0.19\linewidth]{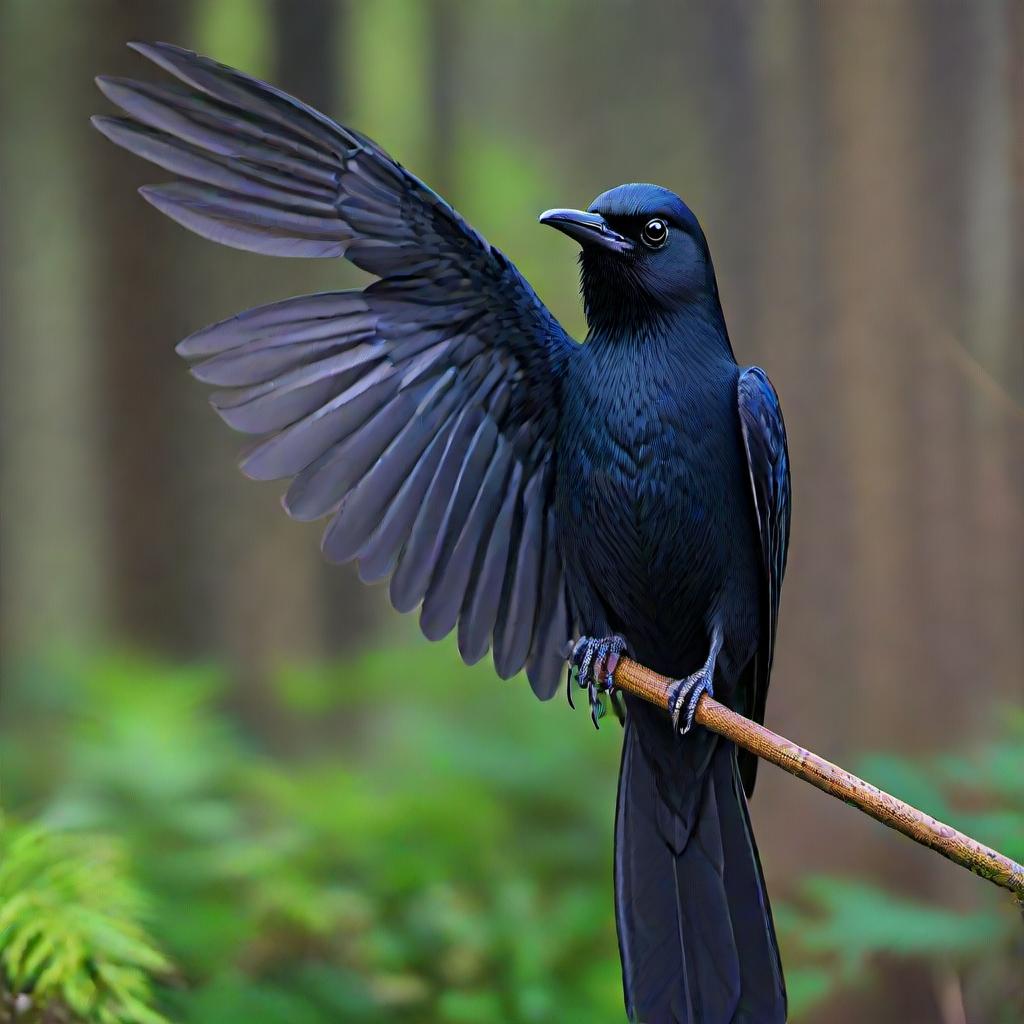}
    \includegraphics[width=0.19\linewidth]{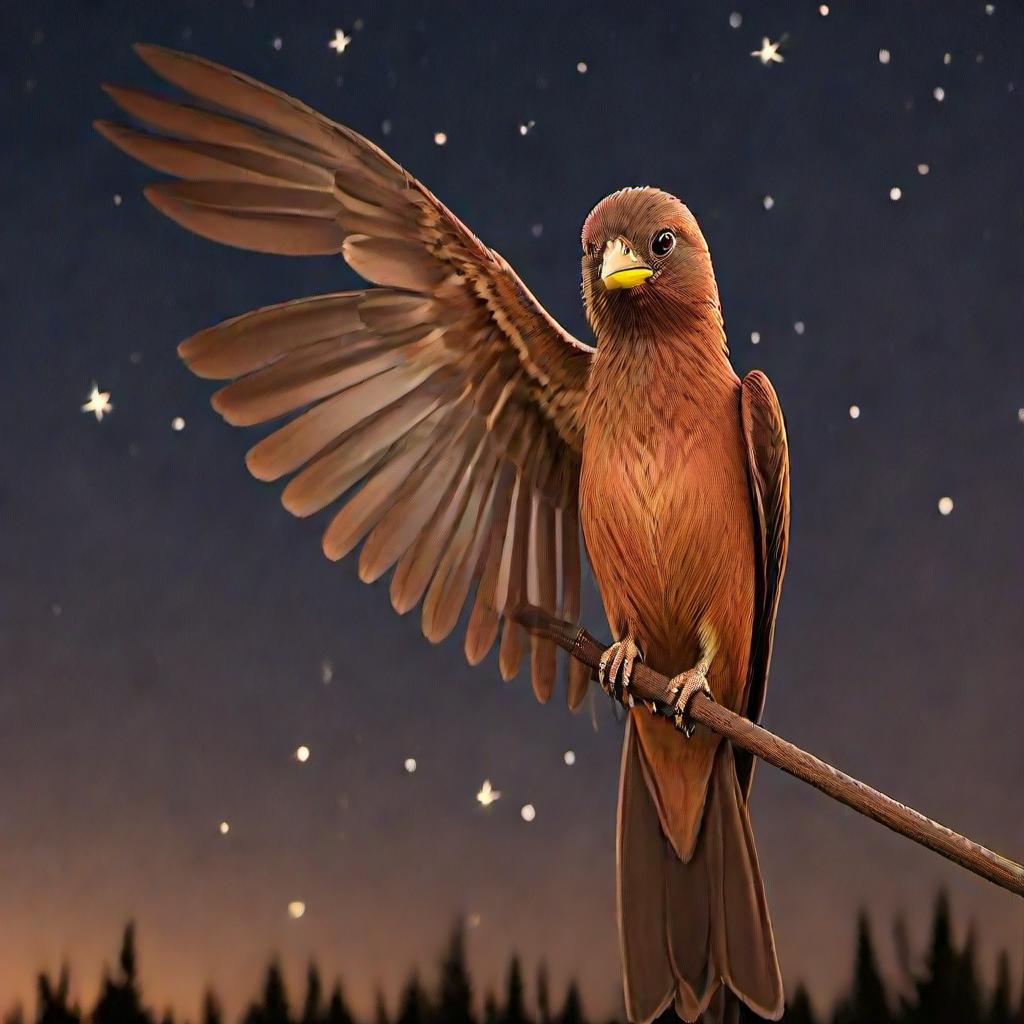}
    \includegraphics[width=0.19\linewidth]{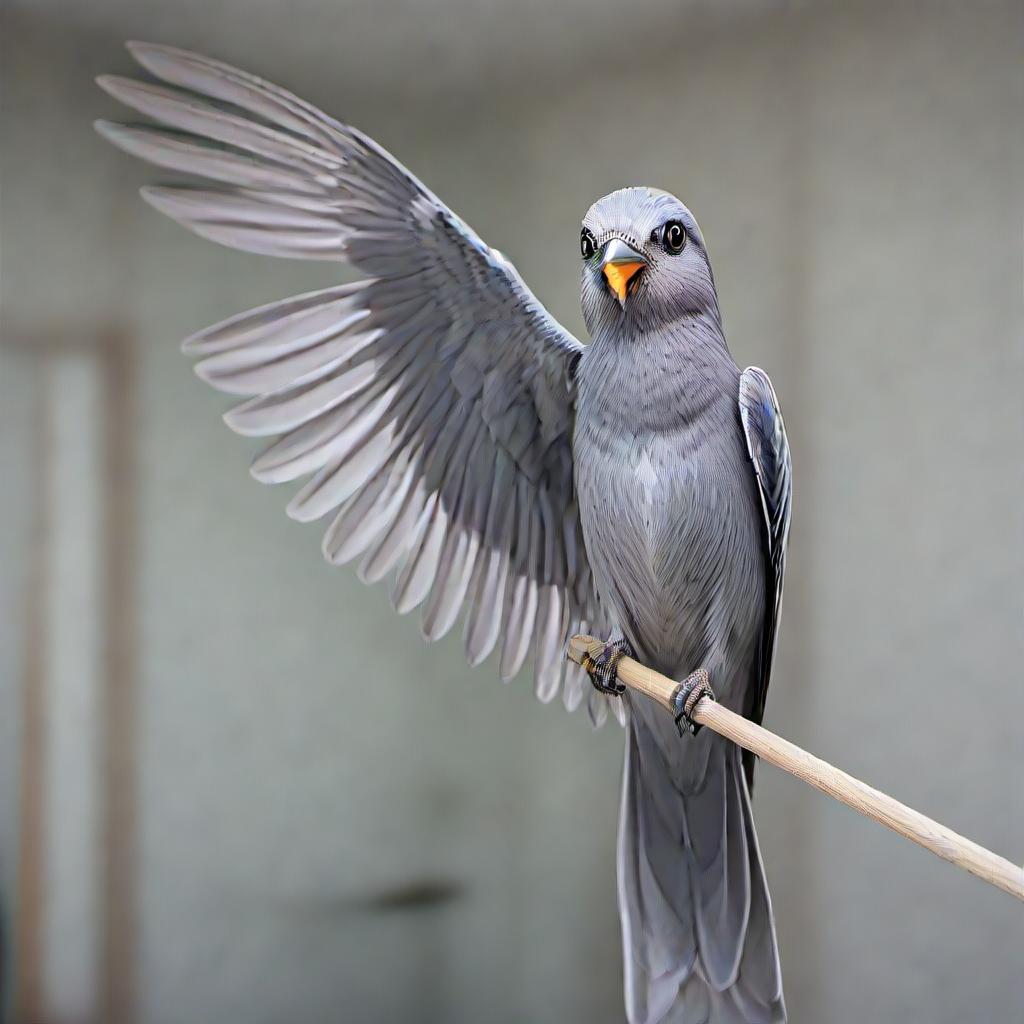}
    \parbox{0.19\linewidth}{\centering Source}
    \parbox{0.19\linewidth}{\centering Depth map}
    \parbox{0.19\linewidth}{\centering A black bird \\ \centering in the forest}
    \parbox{0.19\linewidth}{\centering A brown bird \\ \centering under the stars}
    \parbox{0.19\linewidth}{\centering A gray bird \\ \centering in the room}
    \caption{TLCM with ControlNet.
    Our TLCM can be incorporated into ControlNet pipeline and produce satisfactory results with 2 steps sampling.}
    \label{fig:controlnet}
\end{figure}
\subsection{Application}

\subsubsection{Acceleration of Image Style Transfer}\label{style}
Our TLCM LoRA is compatible with the pipeline of image style transfer~\citep{mou2024t2i}.
We present some examples in Figure~\ref{fig:style} with only 2-step sampling.

\subsubsection{Acceleration of Controllable generation}
Our TLCM LoRA is compatible with Controlnet, enabling accelerated controllable generation.
We utilize canny and depth ControlNet based on SDXL-base, together with TLCM LoRA in Figure~\ref{fig:controlnet}.
The results are sampled in 2 steps.
We observe our model achieves superior image quality and demonstrates compatibility with other models, e.g. ControlNet, while also providing enhanced acceleration capabilities.

\begin{figure}[ht]
    \centering
    \includegraphics[width=1\linewidth]{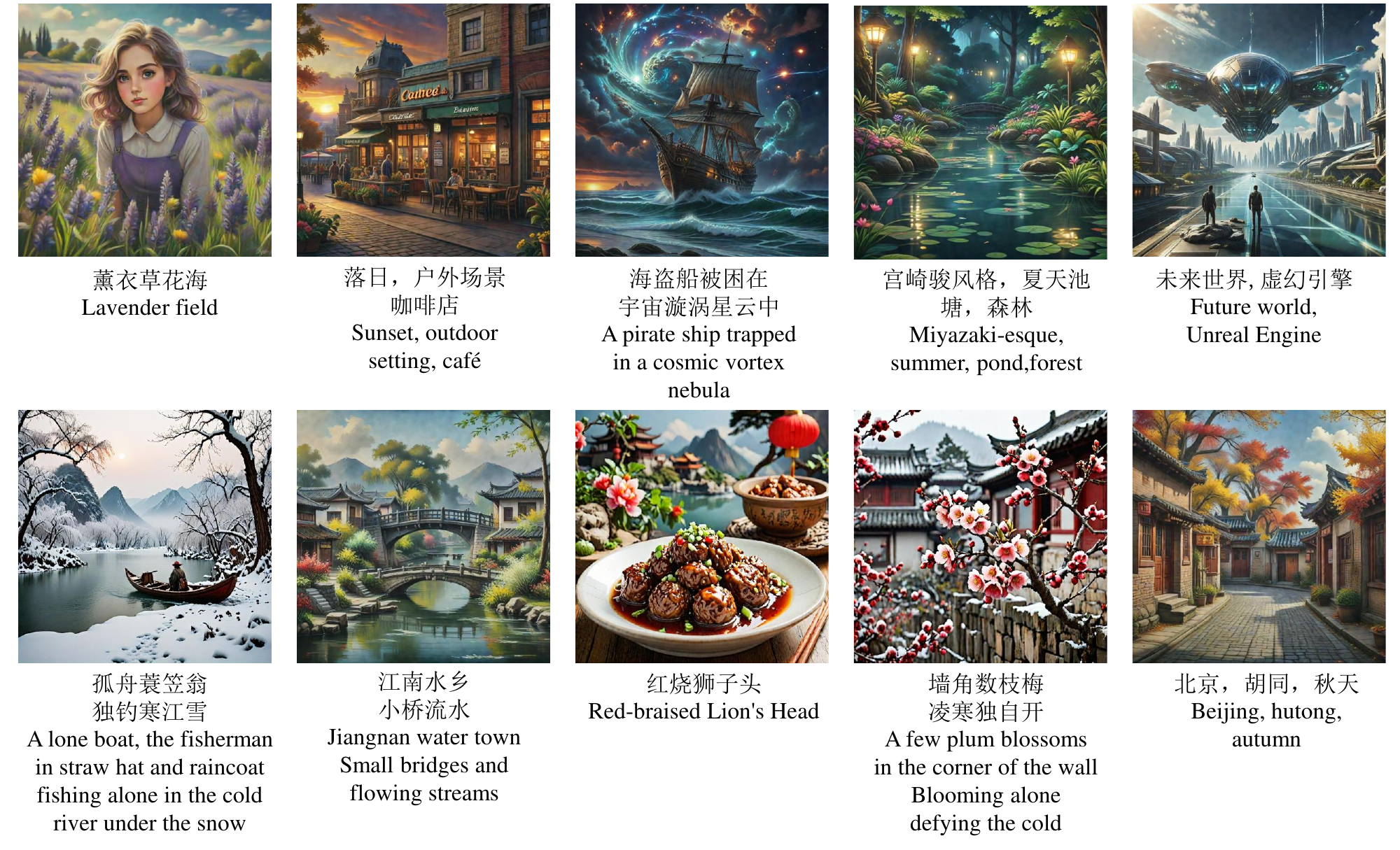}
    \caption{TLCM for Chinese-to-image generation.
    With 3 steps sampling, our TLCM model can produce images that align with Chinese semantic meaning.
    The first line presents images in general Chinese contexts, 
    while the second line showcases images in specific Chinese cultural settings.}
    \label{fig:chinese}
\end{figure}

\subsubsection{Acceleration of Chinese-to-image Generation}\label{chinese}
Our TLCM  can accelerate the generation speed of the Chinese-to-image diffusion model~\citep{ma2023pea}.
We present some examples in Figure~\ref{fig:chinese}.

\end{document}